\documentclass[journal,comsoc]{IEEEtran}

\usepackage{cite}
\usepackage[T1]{fontenc}
\usepackage{array}
\usepackage[dvipdfmx]{graphicx} 
\graphicspath{{images/}}
\DeclareGraphicsExtensions{.pdf,.jpg,.png}
\usepackage[fleqn]{amsmath}
\usepackage{algorithm}
\usepackage{algorithmicx}
\usepackage{algpseudocode}
\ifCLASSOPTIONcompsoc
 \usepackage[caption=false,font=normalsize,labelfont=sf,textfont=sf]{subfig}
\else
 \usepackage[caption=false,font=footnotesize]{subfig}
\fi
\usepackage{fixltx2e}

\usepackage{cite}
\usepackage[acronym]{glossaries}
\usepackage{mathtools}
\usepackage{amsfonts}
\usepackage{amssymb}
\usepackage{bbold} 
\usepackage{bm} 
\usepackage[dvipdfmx]{pdfcomment} 
\usepackage{multirow}
\usepackage{booktabs} 
\usepackage{threeparttable} 
\usepackage{color} 
\usepackage{grffile}
\usepackage{xspace}
\usepackage{tikz}
\usepackage{tikz-3dplot}
\usetikzlibrary{math,fit,shapes.geometric,arrows.meta}

\newenvironment{revhl}[1]{}{}
\newcommand{\revhlnon}[1]{#1}

\newacronym{IID}{IID}{independent and identically distributed}
\newacronym[plural={IoT}]{IoT}{IoT}{Internet of Things}
\newacronym[plural={IoE}]{IoE}{IoE}{Internet of Everything}
\newglossaryentry{Frechet}{name={Fr\'echet}, description={}}
\newacronym{FL}{FL}{federated learning}
\newacronym{CFL}{CFL}{centralized FL}
\newacronym{DFL}{DFL}{decentralized FL}
\newacronym{DL}{DL}{deep learning}
\newacronym{NN}{NN}{neural network}
\newacronym{DNN}{DNN}{deep neural network}
\newacronym{ML}{ML}{machine learning}
\newacronym{GBDT}{GBDT}{gradient boosting decision tree}
\newacronym{MSE}{MSE}{mean squared error}
\newacronym{KL}{KL}{Kullback-Leibler}
\newacronym{CDO}{CDO}{consensus-based distributed optimization}
\newacronym{CMFD}{CMFD}{consensus-based multi-hop federated distillation}
\newacronym{WSN}{WSN}{wireless sensor networks}
\newacronym{P2P}{P2P}{peer-to-peer}
\newacronym{BA}{BA}{Barab\'{a}si--Albert}
\newacronym{SGD}{SGD}{stochastic gradient descent}
\newacronym{CDSGD}{CDSGD}{consensus-based distributed SGD}
\newacronym{DPSGD}{D-PSGD}{decentralized parallel SGD}
\newacronym{FMNIST}{F-MNIST}{fashion MNIST}
\newacronym{GAN}{GAN}{generative adversarial network}
\newacronym{FD}{FD}{federated distillation}
\newacronym{FAug}{FAug}{federated augmentation}
\newacronym{RHS}{RHS}{right-hand side}
\newacronym{RKHS}{RKHS}{reproducing kernel {Hilbert} space}
\newacronym{DGD}{DGD}{distributed gradient descent}
\newacronym{ADMM}{ADMM}{alternating direction method of multipliers}
\newacronym{GADMM}{GADMM}{group ADMM}
\newacronym{D-ADMM}{D-ADMM}{distributed ADMM}
\newacronym{PDMM}{PDMM}{primal-dual method of multipliers}
\newacronym{KD}{KD}{knowledge distillation}
\newacronym{PI}{PI}{proportional-integral}
\newacronym{PID}{PID}{proportional-integral-derivative}
\newacronym{PropAlg}{FedF-ADMM}{federated function-space alternating direction method of multipliers}
\newacronym{CNN}{CNN}{convolutional neural network}
\newacronym{CAV}{CAV}{connected and automated vehicle}
\newacronym{DP}{DP}{differential privacy}
\newacronym{UAV}{UAV}{unmanned aerial vehicle}

\newcommand{\etal}{\textit{et al.\@\xspace}}
\newcommand{\wrt}{with respect to\@\xspace}

\renewcommand{\[}{\left[}
\renewcommand{\]}{\right]}
\renewcommand{\(}{\left(}
\renewcommand{\)}{\right)}

\newcommand{\diag}{\mathop{\mathrm{diag}}}
\newcommand{\inner}[3]{\langle #1, #2 \rangle_{#3}}
\newcommand{\defeq}{\triangleq}
\newcommand{\Real}{\mathbb{R}}
\newcommand{\Neigh}[1]{\mathcal{N}_{#1}}
\newcommand{\ZeroVec}{\bm{0}}

\newcommand{\Laplacian}{L}
\newcommand{\NumNeigh}[1]{n_{#1}}
\newcommand{\sharedD}{\mathcal{D}_\mathrm{s}}
\newcommand{\Devices}{\mathcal{U}}

\newcommand{\DsLocal}[1]{\mathcal{D}_{#1}}
\newcommand{\Pdf}{P}
\newcommand{\PdfLocal}[1]{P_{#1}}

\DeclareMathOperator*{\minimize}{minimize}
\DeclareMathOperator*{\argmin}{arg\,min}

\DeclareMathOperator*{\Expct}{\mathbb{E}}
\newcommand{\CostFn}{C}
\newcommand{\fn}{f}
\newcommand{\fngt}{\fn^\star}
\newcommand{\MeanFn}{\bar{\fn}}
\newcommand{\StackedFn}{\bm{\fn}}
\newcommand{\StackedMeanFn}{\bm{\MeanFn}}
\newcommand{\elemloss}{l}
\newcommand{\NetworkGraph}{\mathcal{G}}
\newcommand{\Edges}{\mathcal{E}}
\newcommand{\inval}{x}
\newcommand{\outval}{y}
\newcommand{\updOutval}{\tilde{\outval}}
\newcommand{\target}[2]{\hat{\outval}_{#1}^{#2}}
\newcommand{\StackedOutVal}{\bm{\outval}}
\newcommand{\StackedUpdOutVal}{\tilde{\bm{\outval}}}

\newcommand{\NumClasses}{K}
\newcommand{\NumDev}{N}
\newcommand{\LagrangeFn}{\mathcal{L}}
\newcommand{\LagMltpFn}{g}
\newcommand{\LagMltpFnWithCoef}{\hat{\LagMltpFn}}
\newcommand{\StackedMltpFn}{\bm{\LagMltpFn}}
\newcommand{\lr}{\eta}
\newcommand{\weight}{\bm{w}}
\newcommand{\intmWeight}{\tilde{\bm{w}}}
\newcommand{\Trans}{\top}
\newcommand{\Mean}[2]{M_{#1}\[#2\]}
\newcommand{\Card}[1]{\lvert{#1}\rvert}
\newcommand{\step}{t}
\newcommand{\tmpStep}{\tau}
\newcommand{\noiseSgd}{\bm{\delta}_1}
\newcommand{\noiseKd}{\bm{\delta}_2}

\newcommand{\StackedDiffY}{\Delta \bm{\outval}}
\newcommand{\StackedExpDiffY}{\overline{\Delta \bm{\outval}}}
\newcommand{\multiplier}{\bm{\lambda}}

\newcommand{\decayPrm}{\nu}
\newcommand{\kdCoef}{\rho}
\newcommand{\lrKdCoef}{\hat{\rho}}
\newcommand{\updDist}{\tilde{\bm{d}}}
\newcommand{\Dist}{\bm{d}}
\newcommand{\Incidence}{B}
\newcommand{\IncidenceT}{\Incidence^\Trans}
\newcommand{\DegMat}{D}
\newcommand{\InvDeg}{\DegMat^{-1}}
\newcommand{\AdjMat}{A}
\newcommand{\MatCoef}{L_\mathrm{e}}
\newcommand{\updateTerm}{\bm{u}}
\newcommand{\updateTermDiff}{\tilde{\bm{u}}}
\newcommand{\tmpVariable}{\bm{\gamma}}

\newcommand{\NumPrm}{N_\mathrm{p}}
\newcommand{\QuantPrm}{Q_\mathrm{p}}
\newcommand{\QuantOutput}{Q_\mathrm{o}}

\newcommand{\avgRate}{\beta}
\newcommand{\FedProxRegRate}{\alpha}
\newcommand{\momentumRate}{\varepsilon}
\newcommand{\admmPenalty}{\theta}

\begin{document}
\title{Function-Space ADMM for Decentralized Federated Learning: A Control Theoretic Perspective}
%

\author{Akihito~Taya,
        and~Yuuki~Nishiyama,
        and~Kaoru~Sezaki,~\IEEEmembership{Member,~IEEE}
\thanks{A. Taya
is with Institute of Industrial Science, The University of Tokyo, Meguro-ku, Tokyo 153-8505, Japan.\protect\\
E-mail: taya-a@iis.u-tokyo.ac.jp}
\thanks{Y. Nishiyama and K. Sezaki
are with Center for Spatial Information Science, The University of Tokyo, Kashiwa-shi, Chiba 277-8568, Japan.}
\thanks{Manuscript received April 19, 2005; revised August 26, 2015.}
}

\markboth{Journal of Sig..,~Vol.~XX, No.~XX, January~2020}%
{Shell \MakeLowercase{\textit{et al.}}: Bare Demo of IEEEtran.cls for Computer Society Journals}

\IEEEtitleabstractindextext{%
\glsresetall
\begin{abstract}
Decentralized \gls{FL} is a promising approach for training machine learning models on sensor networks, \gls{IoT} devices, and other edge systems where no central server exists.
While federated learning offers advantages such as preserving data privacy, it often suffers from non-\gls{IID} data distributions across devices, which cause significant performance degradation.
This issue is particularly severe when directly optimizing model parameters, because neural network training is inherently non-convex and standard convergence guarantees for convex optimization do not apply.
Unlike existing decentralized \gls{FL} methods that primarily operate in parameter space, we propose \gls{PropAlg}.
\Gls{PropAlg} exploits the convexity of loss functionals within function space to derive \gls{ADMM}-based update directions, which are subsequently projected onto the parameter space via knowledge distillation.
We further introduce a stabilization coefficient to enhance robustness under severe non-\gls{IID} settings and analyze its behavior from a control-theoretic perspective by interpreting it as a \gls{PI} term.
Experiments under challenging non-\gls{IID} scenarios, including settings where each device has data from only a single label, demonstrate that \gls{PropAlg} achieves faster and more stable convergence than existing decentralized \gls{FL} methods, while attaining higher accuracy and better consensus among devices.
\end{abstract}

}

\maketitle

\IEEEdisplaynontitleabstractindextext

%
\IEEEpeerreviewmaketitle

\glsresetall

\section{Introduction} \label{sec:intro}
\glsunset{FL}
\IEEEPARstart{F}{ederated} learning (\gls{FL}) has established itself as a leading paradigm for distributed training, enabling collaborative model learning across multiple devices while preserving data privacy \cite{mcmahan2016communication}.
It is particularly attractive for applying deep learning in privacy- and confidentiality-sensitive domains, where collecting data in a single location is infeasible.
By keeping data local and exchanging only model updates, \gls{FL} allows learning from large-scale data while mitigating privacy risks and corporate confidentiality concerns.
Since its introduction by McMahan \etal\cite{mcmahan2016communication}, \gls{FL} has been the subject of extensive research, leading to various advancements and applications, e.g., connected cars, healthcare, and \gls{IoT}.

Although \gls{FL} is a distributed approach to training models, most practical deployments still rely on a central server that maintains a global model.
In such a \gls{CFL} setting, client devices train locally and upload model updates to a central server, which aggregates the updates to maintain a global model.
However, this centralized approach introduces potential bottlenecks, such as a single point of failure and communication overhead.
To overcome these limitations, \gls{DFL} has emerged as an alternative \cite{beltran2022decentralized,pappas2021ipls,onoszko2021decentralized, chen2023enhancing}.
\Gls{DFL} eliminates the need for a central server, addressing single-point-of-failure risks and improving scalability.
Additionally, it enables independence from platform providers, making it a more flexible approach for distributed learning.
Consequently, \gls{DFL} has been applied to various applications, including sensor networks, \gls{IoT} systems, and cross-silo learning between enterprises.

\glsunset{PropAlg}
\glsunset{ADMM}
\glsunset{KD}
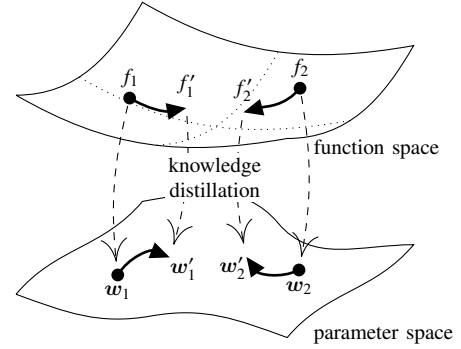
\begin{figure}[t]
    \centering
    \tdplotsetmaincoords{70}{-20} 
\begin{tikzpicture}[
    scale=0.8,
    transform shape, 
    tdplot_main_coords,
    projectarrow/.style={-{Computer Modern Rightarrow[angle=60:10pt]}, dashed, shorten >=5pt, shorten <=3pt},
    updatearrow/.style={-{Stealth[inset=0pt, angle=45:8pt]}, line width=1pt},
]

\coordinate (ua) at (-2,0,3.5);
\coordinate (ub) at (1.5,3.5,3.5);
\coordinate (uc) at (5,0,3.5);
\coordinate (ud) at (1.5,-3.5,3.5);
\coordinate (um1) at (-0.25, 1.75, 3.0);
\coordinate (um2) at (3.25, -1.75, 3.0);
\coordinate (um3) at (3.25, 1.75, 3.0);
\coordinate (um4) at (-0.25, -1.75, 3.0);
\coordinate (f1) at (0,0,3.2);
\coordinate (f2) at (3,0,3.0);
\coordinate (f1_t) at (1,0,2.9);
\coordinate (f2_t) at (2,0,2.7);

\coordinate (la) at (-2,0,0);
\coordinate (lb) at (1.5,3.5,0);
\coordinate (lc) at (5,0,0);
\coordinate (ld) at (1.5,-3.5,0);
\coordinate (w1) at (-0.1,0.3,0);
\coordinate (w2) at (2.8,-0.5,0);
\coordinate (w1_t) at (1,0.8,0);
\coordinate (w2_t) at (2.2,0.4,0);

\draw[thin] (ua)
 to [out=0, in=-120] (ub)
 to [out=-30, in=-170] (uc)
 to [out=-120, in=0] (ud)
 to [out=-170, in=-30] (ua);
\draw[thin] (la)
  to [out=50, in=-130] (lb)
  to [out=10, in=130] ($(lb)!0.5!(lc)$) to [out=-50,in=180] (lc)
  to [out=-130, in=50] (ld)
  to [out=160, in=10] ($(ld)!0.5!(la)$) to [out=-170,in=-20] (la);
\draw[thin, dotted] (um1) to[out=-30, in=-170] (um2);
\draw[thin, dotted] (um3) to[out=-120, in=0] (um4);

\node[anchor=west, xshift=9pt, yshift=-5pt] at (ud) {function space};
\node[anchor=west, xshift=9pt, yshift=1pt] at (ld) {parameter space};

\fill (f1) circle (3pt) node[above, inner sep=5pt] {$f_1$};
\fill (f2) circle (3pt) node[above, inner sep=5pt] {$f_2$};
\node[above, inner sep=5pt] at (f1_t) {$f'_1$};
\node[above, inner sep=5pt] at (f2_t) {$f'_2$};
\draw[updatearrow] (f1) to[out=-30, in=-170] (f1_t);
\draw[updatearrow] (f2) to[out=-130, in=15] (f2_t);

\fill (w1) circle (3pt) node[below, inner sep=5pt] {$\bm{w}_1$};
\fill (w2) circle (3pt) node[below, inner sep=5pt] {$\bm{w}_2$};
\node[below right, inner sep=0pt] at (w1_t) {$\bm{w}'_1$};
\node[below left, inner sep=0pt] at (w2_t) {$\bm{w}'_2$};
\draw[updatearrow] (w1) to[bend left=40] (w1_t);
\draw[updatearrow] (w2) to[bend left=40] (w2_t);

\draw[projectarrow] (f1) to[bend right=10] (w1);
\draw[projectarrow] (f1_t) to[bend left=10] (w1_t);
\draw[projectarrow] (f2) to[bend left=10] (w2);
\draw[projectarrow] (f2_t) to[bend right=10] (w2_t);
\node[fill=white, inner sep=2pt] at (1.5, 0, 1.8) {knowledge};
\node[fill=white, inner sep=2pt] at (1.5, 0, 1.4) {distillation};

\end{tikzpicture}
    \caption{Geometric illustration of \gls{PropAlg}.
    \Gls{PropAlg} determines virtual \gls{ADMM} update directions in function space and uses \gls{KD} to realize these directions as parameter updates.
    \begin{revhl}{1-6}Although a direct mapping from function space to parameter space is not available, $\bm{w}'_i$ is obtained by minimizing a distillation loss.\end{revhl}
    Solid arrows denote updates within each space, and dashed arrows represent the distillation-based mapping from function-space directions to parameter-space updates.
    }
    \label{fig:visual_abstract}
\end{figure}
\glsreset{PropAlg}
\glsreset{ADMM}
\glsreset{KD}

Despite its advantages, \gls{DFL} faces fundamental challenges.
A critical issue is that achieving consensus is difficult without a global coordinator, particularly because many widely used distributed optimization algorithms are designed under the assumption of convex optimization, whereas neural networks are inherently non-convex.
Furthermore, non-\gls{IID} data exacerbates these challenges because local update directions tend to diverge widely across devices.
Although research has extensively explored methods to improve the convergence properties of \gls{DFL} under non-\gls{IID} settings \cite{onoszko2021decentralized, chen2023enhancing}, many studies either focus on parameter convergence without providing theoretical guarantees or rely on strong assumptions such as convexity and smoothness that are unrealistic for deep models.

Instead of focusing on parameter convergence, our previous work \gls{CMFD} \cite{taya2022decentralized} leveraged consensus-based distributed optimization in function space, where the objective function becomes convex \wrt the prediction function.
By exchanging model outputs among devices through \gls{KD}, \gls{CMFD} aggregates prediction functions across heterogeneous models.
However, \gls{CMFD}, which relies on simple consensus-based distributed optimization, exhibits slow convergence, especially in sparsely connected networks, and struggles under severely non-\gls{IID} scenarios, motivating the need for a more robust and faster function-level \gls{DFL} algorithm.

To address these limitations, this paper proposes \gls{PropAlg}.
Unlike existing methods focusing on parameter-level consensus, \gls{PropAlg} exploits the convexity of loss functionals in function space.
As shown in Fig.~\ref{fig:visual_abstract}, \gls{PropAlg} determines \gls{ADMM}-inspired update directions in the function space and realizes them through \gls{KD}-based parameter updates, enabling each local model to move toward the target post-update functions prescribed by \gls{ADMM} iterations in the function space.
We also adopt a computationally efficient mechanism for handling Lagrange multipliers.
When \gls{ADMM} is formulated in function space, the Lagrange multipliers---normally vectors in standard parameter-space \gls{ADMM}---become functions whose outputs vary over the input domain.
Instead of explicitly training these multiplier functions, \gls{PropAlg} stores only their output values on a shared dataset, thereby reproducing the effect of functional multiplier updates while keeping computational overhead minimal.

Furthermore, we developed a technique to enhance the stability of \gls{PropAlg} under severe non-\gls{IID} settings and analyze its behavior from the perspective of control theory.
In single-label scenarios---representing the most extreme divergence from the global label distribution---local \gls{SGD} updates often produce unstable outputs for label-missing categories.
To mitigate this issue, we introduce a stabilization coefficient that smooths output fluctuations by incorporating information distilled from neighboring devices.
We show that the resulting update dynamics can be interpreted as a \gls{PI} control mechanism, in which the stabilization coefficient plays a role analogous to the integral term, suppressing long-term drift caused by non-\gls{IID}ness.
This perspective provides new insight into the dynamics of decentralized learning in function space and clarifies how the proposed mechanism improves empirical convergence stability.

\begin{revhl}{1-general}
Our focus is twofold: (i) to demonstrate that operating in function space can substantially improve convergence and stability in decentralized learning, and (ii) to use these empirical findings to clarify the limitations of heuristic parameter aggregation and motivate function-level formulations.
While practical deployment of \gls{KD}-based \gls{FL} still requires appropriate mechanisms for shared probing data and privacy protection, these aspects are complementary to the main algorithmic contribution of this work.
The present work shows that \gls{ADMM}-style optimization in function space offers a robust and effective design principle for decentralized learning under severe non-\gls{IID} conditions.
\end{revhl}

The contributions of this paper are summarized as follows:
\begin{enumerate}
    \item We propose \gls{PropAlg}, a function-level decentralized learning algorithm that integrates \gls{KD} with \gls{ADMM}-inspired updates. 
    By operating in function space, the proposed method leverages convexity of the prediction function even when neural network parameter optimization is non-convex.
    \item We introduce a stabilization coefficient to improve robustness under severe non-\gls{IID} settings. 
    This mechanism suppresses unstable output fluctuations for label-missing categories by incorporating distilled knowledge from neighboring devices.
    \item We provide a control-theoretic interpretation of the update dynamics of \gls{PropAlg}. 
    In particular, we show that the stabilization mechanism behaves analogously to the integral term of a \gls{PI} controller, offering conceptual insight into empirical stability in decentralized learning.
    \item We empirically demonstrate that \gls{PropAlg} achieves more stable and accelerated empirical convergence compared with existing decentralized learning methods, especially under severe statistical heterogeneity.
\end{enumerate}

The rest of this paper is organized as follows:
Sec.~\ref{sec:related} provides an overview of related work,
and Sec.~\ref{sec:problem} describes the system model and problem statement.
Sec.~\ref{sec:algorithm} explains the details of the proposed \gls{PropAlg},
and Sec.~\ref{sec:analysis} analyzes the effectiveness of the stabilization coefficient.
Evaluation results are then presented in Sec.~\ref{sec:evaluation}.
Finally, Sec.~\ref{sec:conclusion} concludes the paper.

\section{Related work} \label{sec:related}
\subsection{Decentralized learning algorithms in multi-hop networks} \label{sec:related_dfl}
\Gls{FL} was originally introduced as a privacy-preserving distributed machine learning framework \cite{mcmahan2016communication}.
Research in \gls{FL} spans numerous directions, including improving communication efficiency and enhancing performance under non-\gls{IID} settings.
\Gls{DFL} extends \gls{FL} to settings where clients collaborate without reliance on a central server \cite{beltran2022decentralized,pappas2021ipls}.
While some studies assume fully connected networks, \gls{DFL} in multi-hop topologies has also been actively investigated \cite{lian2017can,lalitha2019peer,savazzi2020federated}.

In multi-hop networks, devices cannot synchronize parameters globally, and thus learning algorithms must drive local parameters toward a common value.
Distributed consensus optimization---often referred to as gossip algorithms---is widely adopted for such settings.
These algorithms exchange parameters only with neighboring devices and iteratively align them through consensus operations \cite{lalitha2019peer,savazzi2020federated,lian2017can,sato2020network,sun2022decentralized}.

Jiang \etal \cite{jiang2017collaborative} introduced a basic \gls{DFL} algorithm, \gls{CDSGD}, where each device updates its parameters toward the average of its neighbors.
Lian \etal \cite{lian2017can} proposed \gls{DPSGD} and demonstrated that it can outperform centralized parallel \gls{SGD} by mitigating communication congestion.
Savazzi \etal \cite{savazzi2020federated} developed a \gls{DFL} framework leveraging the cooperation of \gls{IoT} devices, and Lalitha \etal \cite{lalitha2019peer} applied a Bayesian approach to peer-to-peer \gls{FL}.
Considering wireless channels, Sato \etal \cite{sato2020network} extended \gls{DPSGD} for decentralized training over wireless networks.
\begin{revhl}{1-4}
Sun \etal \cite{sun2022decentralized} proposed to use momentum in \gls{DFL} to improve training performance and integrated quantization techniques to reduce communication costs.
\end{revhl}

Beyond consensus-based distributed optimization approaches, several studies have applied \gls{ADMM} to \gls{DFL}.
\Gls{ADMM} offers faster convergence than gradient descent-based methods and supports parallel updates; therefore, \gls{D-ADMM} \cite{mota2013dadmm} has been used to address consensus optimization and is applicable to decentralized machine learning.
For example, Wang \etal \cite{wang2023communication} introduced an \gls{ADMM}-based distributed learning algorithm for sparse training, although its applicability to neural networks remains unexplored.
Niwa \etal \cite{niwa2020edge} proposed \gls{ADMM}- and \gls{PDMM}-based \gls{SGD} algorithms for neural network \gls{DFL} in edge-computing scenarios with non-\gls{IID} data.
\begin{revhl}{1-4}
Li \etal \cite{li2025dfedadmm} also developed an \gls{ADMM}-based \gls{DFL}, DFedADMM, which addresses the local inconsistency during the training process and avoid client drift caused by local over-fitting.
\end{revhl}

Although \gls{ADMM} has been extensively analyzed for convex optimization problems, research on applying \gls{ADMM} to neural-network \gls{FL}---where objectives are non-convex---remains limited.
Zhou \etal \cite{zhou2023federated} analyzed \gls{ADMM}-based \gls{CFL} without assuming convexity, but their evaluation was restricted to linear and logistic regression.
Other works in non-convex \gls{ADMM} include inexact \gls{ADMM} \cite{bai2022inexact,bai2025inexact} and linearized \gls{ADMM} \cite{liu2019linearized}.
These studies analyze convergence to stationary points, yet guaranteeing global optimality remains fundamentally difficult due to non-convexity.
Motivated by this, our work focuses on function-level convergence, which becomes a convex problem when formulated in function space as explained in Sec.~\ref{sec:problem}.

\subsection{Knowledge distillation-based federated learning}\label{sec:related_kdfl}
While the above approaches synchronize model parameters, \gls{KD}-based \gls{FL} provides an alternative path by aligning model outputs rather than parameters, mitigating challenges such as high communication costs and heterogeneous model architectures \cite{mora2024knowledge}.
Prior to its adoption in \gls{DFL}, co-distillation was proposed for data-center environments \cite{anil2018large,zhang2018deep}, where devices train local models and exchange shared output predictions, which are averaged to form soft labels.
This process can be viewed as a distributed form of \gls{KD} \cite{hinton2015distilling}.
While co-distillation was designed for data-center environments, and thus it assumes shared training data, the similar idea has been proposed for \gls{FL} scenarios.

Jeong \etal \cite{jeong2018communication} proposed \gls{KD}-based \gls{FL} known as \gls{FD} and \gls{FAug} to address non-\gls{IID} issues by generating missing samples using generative models, and later extended these schemes for multi-hop \gls{DFL} in \cite{jeong2019multi}.
Itahara \etal \cite{itahara2021distillation} proposed a \gls{KD}-based semi-supervised \gls{FL} algorithm to enhance robustness against malicious clients.
Model heterogeneity is another challenge in \gls{FL} that \gls{KD} can address.
Because devices in mobile and \gls{IoT} systems often employ different model architectures, \gls{KD}-based \gls{FL} is attractive for its model-agnostic nature.
Representative methods include Cronus \cite{chang2019cronus}, FedMD \cite{li2019fedmd}, and FedDF \cite{lin2020ensemble}.
Our prior work, \gls{CMFD} \cite{taya2022decentralized}, applied \gls{KD} for \gls{DFL} in multi-hop networks and showed that \gls{KD}-based \gls{DFL} can approximate consensus-based distributed optimization in function space, which remains convex even when parameter-space optimization is non-convex.

\begin{revhl}{1-9}
Recent studies continue to explore \gls{KD}-based \gls{FL} toward various directions.
Gad \etal \cite{gad2024communication} applied \gls{DP} to \gls{KD}-based \gls{FL} to mitigate privacy risks, and Sun \etal \cite{sun2024fkd} applied \gls{KD}-based \gls{FL} in medical image segmentation tasks.
Li \etal \cite{li2025pfedkd} proposed a personalized \gls{FL} using \gls{KD} in \gls{IoT} scenarios.
For decentralized ones, Soltani \etal \cite{soltani2024dflstar} proposed DFLStar, which utilizes self-\gls{KD} and client selection mechanism in \gls{DFL}.
Yang \etal \cite{yang2025dfun} applied \gls{KD}-based \gls{DFL} to \gls{UAV} and proposed DFUN-KDF, which utilized filtering techniques to ensure robustness against disconnection and Byzantine attacks.
\end{revhl}

\begin{revhl}{1-1,1-3}
Note that the practical adoption of \gls{KD}-based \gls{FL} entails inherent trade-offs, such as the requirement for a shared input dataset and additional computational overhead; our discussion serves as general motivation rather than a claim to resolve all such challenges.
One of the remaining challenges that we do not address in this work is the privacy risk, as shared input data for \gls{KD} can potentially leak information about local datasets.
To mitigate these risks, \gls{DP}, which is a common approach in \gls{FL} \cite{wei2020federated,he2024clustered,adnan2025framework}, can be integrated by adding noise to shared output values, offering a quantifiable privacy guarantee as proposed in \cite{gad2024communication}.

Alternatively, data-free \gls{KD} mechanisms can eliminate the need for sharing raw input data to mitigate privacy risks.
Zhu \etal \cite{zhu2021data} proposed data-free \gls{KD} for \gls{FL}, where the server trains a generator using noise resampling to learn the output layer from latent variables, and distributes it to clients, whose knowledge is distilled to user models.
Zhang \etal \cite{zhang2024fedgmkd} developed FedGMKD using Gaussian mixture models to compute prototype features and soft predictions in a data-free manner.
FedFTG \cite{zhang2022fine} is another data-free \gls{KD}-based \gls{FL} method that trains a generator to produce pseudo samples hard to classify rather than to learn the actual input distribution.
Using these data-free methods, \gls{KD}-based \gls{FL} can be implemented without sharing input data, meaning that the privacy risk of \gls{KD}-based \gls{FL} is not necessarily higher than that of parameter-sharing \gls{FL}, since logits can be derived when parameters are shared.
While these privacy-preserving techniques are orthogonal to our core contribution of realizing \gls{ADMM} in function space, we leave this integration as future work.
\end{revhl}

In contrast to prior \gls{KD}-based \gls{FL} methods, which mainly use distillation as a straightforward mechanism for model averaging, our approach employs a specialized distillation formulation designed to realize \gls{ADMM} in function space as explained in Sec.~\ref{sec:algorithm}.

\subsection{Control theory and federated learning}
Prior studies have explored connections between \gls{FL} and control theory.
Mashaal \etal \cite{mashaal2024extending} integrated \gls{PID} controller terms into SCAFFOLD \cite{karimireddy2020scaffold} to mitigate drift issues in \gls{FL}.
Li \etal \cite{li2025fed} proposed Fed-PID, which employs closed-loop feedback mechanisms between a server and clients to automatically adjust learning rates during training.
Gao \etal \cite{gao2023fedadt} proposed FedADT incorporating an adaptive learning rate and derivative term in the update of local models based on \gls{PID} control.
M{\"a}chler \etal \cite{machler2022fedpidavg} proposed FedPIDAvg, applying \gls{PID} control directly to model aggregation to achieve high performance for tumor segmentation tasks.
Separately, Zen \etal \cite{zeng2021federated} applied \gls{FL} to automotive control, where \glspl{CAV} use \gls{PID} controllers whose parameters are collaboratively optimized.

Unlike the above studies that employ control theory to tune optimization dynamics in parameter space, our work uses control theory to analyze model convergence in function space, focusing on the evolution of model outputs rather than parameters. 
To the best of our knowledge, this is the first study to reveal a control-theoretic interpretation of convergence behavior in decentralized learning from a function-space perspective.

\section{System model and problem statement} \label{sec:problem}
\begin{figure}[!t]
\centering
\includegraphics[width=0.4\textwidth]{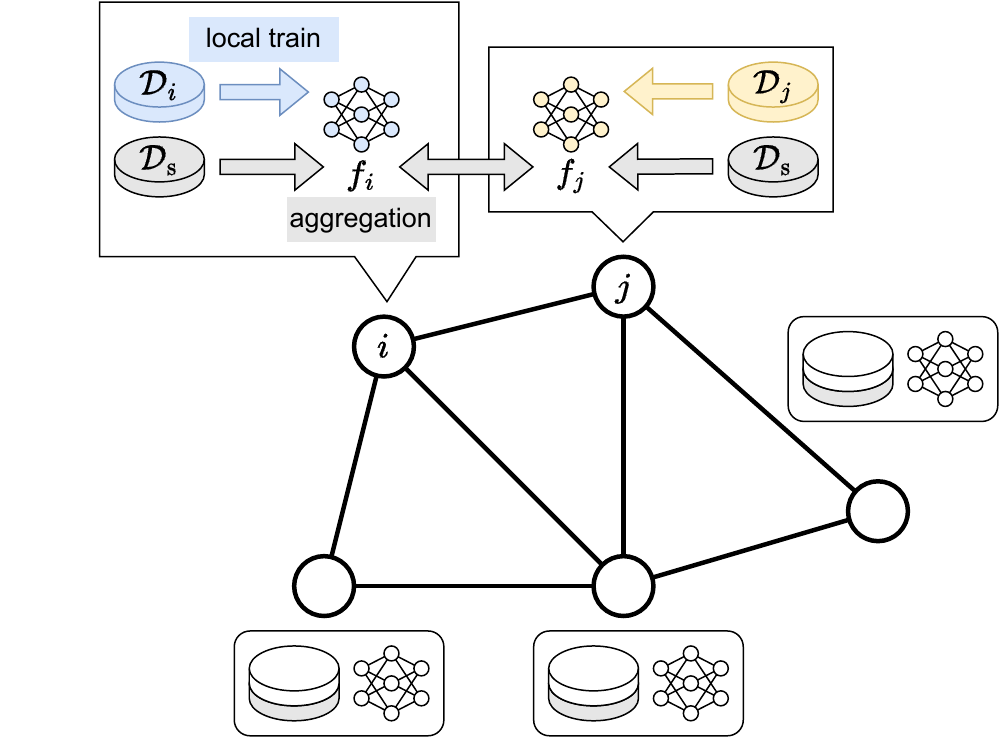}
\caption{System model of \gls{PropAlg}. Devices collaboratively train their local models on their own datasets by communicating with one another. They exchange distilled knowledge on a shared dataset to achieve consensus.}
\label{fig:system_model}
\end{figure}

We consider a \gls{DFL} system without central servers, where multiple devices collaboratively train their local models on their own datasets by communicating with one another, as shown in Fig.~\ref{fig:system_model}.
Let $\Devices$ denote the set of devices participating in the \gls{DFL} process.
There are $\NumDev$ devices in the system, where each device $i \in \Devices$ has its own local dataset $\DsLocal{i}$ sampled from a local distribution $\PdfLocal{i}$.
We adopt a non-\gls{IID} scenario with both label and feature distribution skew \cite{kairouz2019advances}, where the label distributions $\PdfLocal{i}(y)$ and feature distributions $\PdfLocal{i}(x)$ vary across devices, while the conditional distribution $\Pdf(y|x)$ is common across devices.
There is also a public unlabeled dataset $\sharedD$ shared among devices, sampled from a global distribution $\Pdf$ and containing data from all categories.
This setting follows \cite{itahara2021distillation,taya2022decentralized}; alternatively, one can generate an unlabeled shared dataset via generative augmentation methods as in \cite{jeong2018communication}.

Devices are connected in a multi-hop manner and can directly communicate only with their neighbors.
Let the communication graph be $\NetworkGraph = (\Devices, \Edges)$, where $\Edges$ is the set of edges, and $(i,j) \in \Edges$ indicates that devices $i$ and $j$ can exchange messages.
Let $\Neigh{i}$ denote the neighbor set of device $i$, and let $\NumNeigh{i}\defeq \Card{\Neigh{i}}$ be its size.
\begin{revhl}{2-6,2-11}
Let $\DegMat\defeq\diag\{\NumNeigh{i}\}$, $\AdjMat$, and $\Laplacian\defeq\DegMat-\AdjMat$ denote the degree matrix, adjacency matrix, and graph Laplacian matrix of the communication graph $\NetworkGraph$, respectively.
The row-stochastic matrix $\InvDeg\AdjMat$ is used to compute the average of neighboring values for each device.
We also define the incidence matrix $\Incidence \in \Real^{\Card{\Devices}\times 2\Card{\Edges}}$ by treating each undirected edge as two oppositely oriented directed edges.
This definition allows the differences between adjacent device values to be calculated as $\IncidenceT \bm{x}$, where $\bm{x}$ is a vector of device values.
Furthermore, the relationship $\Laplacian=\frac{1}{2}\Incidence \IncidenceT$ holds.
\end{revhl}

In the absence of central servers, the devices exchange model information to reach consensus on a global solution while continuing local training.
This problem can be formulated as follows:
\begin{revhl}{2-1}\end{revhl}
\begin{align}
  &\minimize_{\fn_i} \quad \CostFn \defeq \sum_{i=1}^{\NumDev} \CostFn_i (\fn_i), \label{eq:problem} \\
  &\quad \text{s.t.} \quad \forall \revhlnon{(i,j)} \in \Edges: \fn_i = \fn_j, \label{eq:problem_constraint} \\
  &\quad \mathrm{where} \quad \CostFn_i (\fn_i) \defeq \Expct_{\inval\sim\PdfLocal{i}} \[\elemloss(\fn_i(\inval), \fngt(\inval))\]. \label{eq:def_costfn}
\end{align}
Here, $\elemloss$, $\fn_i$, $\fngt$, and $\Edges$ are an element-wise cost function, the local model of device $i$, the ground truth model, and the set of device pairs that can communicate, respectively.
The local model $\fn_i$ can also be represented using a parameter vector $\weight_i$ as $\fn_{\weight_i}$ and $\fn(\cdot;\weight_i)$.

Conventional \gls{FL} algorithms such as FedAvg \cite{mcmahan2016communication} and FedProx \cite{li2020federated} are designed to minimize the above problem by optimizing the local parameters $\weight_i$.
Nevertheless, these approaches are inefficient because they use optimization procedures developed for convex problems on a non-convex objective: $\fn_i$ is non-convex in $\weight_i$, and thus so is $\CostFn$.
In contrast, our previous work \cite{taya2022decentralized} proposed to optimize the prediction model $\fn_i$ directly because $\CostFn$ is convex \wrt $\fn_i$ for typically used cost functions, e.g., \gls{MSE} and \gls{KL} divergence.
To improve the convergence performance of the previous work, which adopted a consensus-based optimization, we propose to use \gls{D-ADMM} in function space to directly optimize the prediction model $\fn_i$.
Although certain loss functions such as \gls{KL} divergence are convex but not strongly convex \wrt $\fn_i$, \gls{ADMM} is known to remain effective in practice even when strong convexity assumptions are not satisfied \cite{boyd2011distributed}.

\section{Function-level ADMM for decentralized federated learning} \label{sec:algorithm}
\subsection{Derivation and implementation} \label{sec:proposal_impl}
To solve the optimization problem \eqref{eq:problem} \wrt the prediction function, we propose \gls{PropAlg}, which applies the principles of \gls{D-ADMM} directly in function space.
Before introducing the algorithm, we reformulate the constraints \eqref{eq:problem_constraint} as follows:
\begin{align}
  \forall i \in \Devices: \fn_i = \Mean{i}{\fn_j}, \label{eq:problem_constraint_reformulated}
\end{align}
where $\Mean{i}{\inval_j} \defeq \frac{1}{\NumNeigh{i}}\sum_{j \in \Neigh{i}} \inval_j$ is the mean of the neighbors' values $\inval_j$.
The conditions \eqref{eq:problem_constraint} and \eqref{eq:problem_constraint_reformulated} are equivalent when the communication graph is connected, since enforcing consensus \wrt neighbor averages yields a global consensus in connected graphs.
This follows from the spectral property that the Laplacian matrix of a connected graph has a one-dimensional null space spanned solely by the all-ones vector.

Although several \gls{D-ADMM} variants exist, we adopt the following scheme due to its implementation simplicity and communication efficiency.
The augmented Lagrangian $\LagrangeFn$ is defined as a functional over prediction functions as follows:
\begin{align}
  &\LagrangeFn(\StackedFn, \StackedMeanFn, \StackedMltpFn)
     \defeq \sum_{i=1}^{\NumDev} \LagrangeFn_i(\fn_i, \MeanFn_i, \LagMltpFn_i),
\end{align}
where the local augmented Lagrangian $\LagrangeFn_i$ is given by:
\begin{align}
  &\LagrangeFn_i(\fn_i, \MeanFn_i, \LagMltpFn_i) \nonumber \\
  &\quad \defeq \CostFn_i(\fn_i) + \inner{\LagMltpFn_i}{\fn_i - \MeanFn_i}{\Pdf} + \frac{\kdCoef}{2} \left\| \fn_i - \MeanFn_i \right\|^2. \label{eq:def_lagrangian}
\end{align}
Here, $\kdCoef$, $\MeanFn_i\defeq\Mean{i}{\fn_j}$, and $\LagMltpFn_i$ denote the penalty parameter, mean of the neighbors' functions, Lagrange multipliers, respectively.
The inner product is defined as $\inner{f}{g}{\Pdf}\defeq\int_xf(x)g(x)\mathrm{d}\Pdf(x)$.
The bold symbols represent stacked vectors of functions, e.g., $\StackedFn \defeq [\fn_1, \fn_2, \ldots, \fn_\NumDev]^\Trans$.
Note that $\LagMltpFn_i$, which corresponds to the Lagrange multiplier, is itself a function, since we are optimizing functions.

At each communication round $\step$, each device $i$ performs the following two steps:
\begin{align}
  \text{Step 1.} \quad \fn_i^{\step+1}
    \leftarrow& \argmin_{\fn_i} \LagrangeFn_i(\fn_i, \MeanFn_i^\step, \LagMltpFn_i^\step), \label{eq:step1} \\
  \text{Step 2.} \quad \LagMltpFn_i^{\step+1}
    \leftarrow& \LagMltpFn_i^\step + \kdCoef (\fn_i^{\step+1} - \MeanFn_i^{\step+1}). \label{eq:step2}
\end{align}
This formulation allows \gls{ADMM} to operate directly on prediction functions, with the quadratic penalty encouraging local consensus among neighboring devices.
Note that, although a basic \gls{D-ADMM} calculates Step~1 sequentially for each device, we compute it in parallel for efficiency.

Since solving \eqref{eq:step1} is a functional optimization problem and it is difficult to obtain a closed-form solution, we propose a practical implementation that approximates the solution of the subproblem using \gls{KD}.
This approach allows the parameterized model to track the target function prescribed by the \gls{ADMM} update.
Before applying \gls{KD}, we rewrite \eqref{eq:step1}.
\begin{revhl}{2-2}
By completing the square with respect to $\fn_i$ and discarding the constant terms relative to $\fn_i$, we can rewrite \eqref{eq:step1} as follows:
\end{revhl}
\begin{align}
  \fn_i^{\step+1}
    &= \argmin_{\fn_i} \left\{ \CostFn_i(\fn_i) + \frac{\kdCoef}{2} \left\| \fn_i - \MeanFn_i^\step + \frac{1}{\kdCoef} \LagMltpFn_i^\step\right\|^2 \right\}. \label{eq:update_fn}
\end{align}
Instead of solving this subproblem exactly, we adopt a gradient-descent-based approach to update the prediction function $\fn_i^\step$.
Similar approaches have been used in the literature, such as inexact \gls{ADMM}~\cite{bai2025inexact} and linearized \gls{ADMM}~\cite{liu2019linearized}, where the subproblem is solved approximately to reduce computational complexity.
Assuming that each prediction function is parameterized by $\weight_i$, the parameter update induced by the functional objective becomes:
\begin{align}
  \weight_i^{\step+1}
    =& \weight_i^{\step} - \lr  \nabla_{\weight_i} \CostFn_i(\fn_{\weight_i^\step}) \nonumber \\
    &- \lr \nabla_{\weight_i} \frac{\kdCoef}{2} \left\| \fn_{\weight_i^\step} - \MeanFn_i^\step + \frac{1}{\kdCoef} \LagMltpFn_i^\step\right\|^2. \label{eq:update_fn_param}
\end{align}

\begin{revhl}{2-3}
To implement this update, we decompose it into two steps: a local update using the local dataset and an aggregation phase leveraging distilled knowledge from neighbors.
Since $\CostFn_i(\fn_{\weight_i^\step})$ is defined as the local cost function in \eqref{eq:def_costfn}, the term $\nabla_{\weight_i} \CostFn_i(\fn_{\weight_i^\step})$ represents the standard local update in \gls{FL} frameworks; thus, the first step is implemented using \gls{SGD} on the local dataset $\DsLocal{i}$ as follows:
\begin{align}
  \intmWeight_i^{\step} = \weight_i^{\step} - \lr \nabla_{\weight_i} \sum_{\inval\sim\DsLocal{i}} \elemloss\(\fn(\inval;\weight_i^\step), \fngt(\inval)\), \label{eq:localsgd}
\end{align}
where $\intmWeight_i^{\step}$ denotes the intermediate parameter after the local update.

Now, we derive the second step for the aggregation phase.
To calculate the third term of \eqref{eq:update_fn_param}, we can employ the \gls{KD} technique while fixing the neighbors' prediction functions $\fn_j^\step$ and the Lagrange multiplier $\LagMltpFn_i^\step$.
\end{revhl}
By defining the virtual target $\target{i,\inval}{\step}\defeq \MeanFn_i^\step(\inval) - \frac{1}{\kdCoef} \LagMltpFn_i^\step(\inval)$, the aggregation phase can be implemented as follows:
\begin{align}
  \weight_i^{\step+1} &\leftarrow \revhlnon{\intmWeight_i^{\step}}
  - \frac{\lrKdCoef}{2} \nabla_{\weight_i} \sum_{\inval\sim\sharedD} \left\| \fn(\inval;\revhlnon{\intmWeight_i^\step}) - \target{,\inval}{\step} \right\|^2, \label{eq:aggregation}
\end{align}
\begin{revhl}{2-4}
where $\lrKdCoef\defeq\lr\kdCoef$.
\end{revhl}
The virtual target $\target{i,\inval}{\step}$ represents the \gls{ADMM}-prescribed update direction in function space, and \gls{KD} provides a mechanism to move the parameterized model toward this target.
Fortunately, the multiplier functions $\LagMltpFn_j^\step$ do not need to be explicitly parameterized; it suffices to store their values $\LagMltpFn_{j}^\step(\inval)$ on the shared dataset $\sharedD$.

\begin{figure}[!t]
  \begin{algorithm}[H]
    \caption{Overview of the proposed \gls{PropAlg}}
    \label{alg:fl_dadmm}
    \begin{algorithmic}[1]
      \State Synchronize $\weight_i^0$ and initialize $\LagMltpFnWithCoef_{i,\inval}^{0} \leftarrow 0$ for all device $i$ \label{stp:init}
      \While {not converged}
        \ForAll{device $i\in\Devices$}
          \State update \revhlnon{$\intmWeight_i^\step$} using local dataset $\DsLocal{i}$ by \eqref{eq:localsgd} \label{stp:local_update}
          \State share the updated \revhlnon{$\intmWeight_i^\step$} with neighbors \label{stp:info_sharing}
          \State update $\LagMltpFnWithCoef_{i,\inval}^{\step}$ by \eqref{eq:step2} or \eqref{eq:robust_multiplier_update} \label{stp:multiplier_update}
          \State $\target{i,\inval}{\step}\defeq \Mean{i}{\fn(\inval;\revhlnon{\intmWeight_j^{\step+1}})} - \LagMltpFnWithCoef_{i,\inval}^{\step+1}$ \label{stp:target_update}
          \State update $\weight_i^\step$ using \gls{KD} by \eqref{eq:aggregation} \label{stp:aggregation}
        \EndFor
      \EndWhile
    \end{algorithmic}
  \end{algorithm}
\end{figure}

Now, we can summarize \gls{PropAlg} in Alg.~\ref{alg:fl_dadmm}.
\begin{revhl}{2-4}
Here, $\LagMltpFnWithCoef_{i,\inval}^\step$ is defined as $\LagMltpFnWithCoef_{i,\inval}^\step\defeq\frac{1}{\kdCoef} \LagMltpFn_{i}^\step(\inval)$.
\end{revhl}
First, we assume that all devices synchronize their initial parameters $\weight_i^0$ as suggested in \cite{mcmahan2016communication} and the outputs of the multiplier functions $\LagMltpFnWithCoef_{i,\inval}^{0} \leftarrow 0$ for all device $i$ (Line \ref{stp:init}).
In the local update step (Line \ref{stp:local_update}), each device updates its local model using \gls{SGD} with the local dataset $\DsLocal{i}$ in the same manner as typical \gls{FL} frameworks.
Then, each device shares the updated parameters with its neighbors (Line \ref{stp:info_sharing}).
In this step, the output values of the local models corresponding to the shared dataset can be shared instead of the parameters to reduce communication costs.
The outputs of multiplier functions and the virtual target are updated in Lines \ref{stp:multiplier_update} and \ref{stp:target_update}, respectively.
Finally, the local model is updated using the \gls{KD} technique in Line \ref{stp:aggregation} to achieve the consensus on the prediction function.

Although \gls{PropAlg} is effective for some scenarios, we found that the performance of Alg.~\ref{alg:fl_dadmm} deteriorates under severe non-\gls{IID} settings as evaluated in Sec.~\ref{sec:evaluation}.
To overcome this issue, we introduce a stabilization coefficient $\decayPrm$ to stabilize the output values for missing categories in each local device.
This coefficient smooths the multiplier updates by attenuating abrupt changes, and the modified update rule is given as follows:
\begin{align}
  \LagMltpFnWithCoef_{i,\inval}^{\step+1} \leftarrow (1-\decayPrm) \LagMltpFnWithCoef_{i,\inval}^{\step} + \fn(\inval;\revhlnon{\intmWeight_i^{\step+1}}) - \Mean{i}{\fn(\inval;\revhlnon{\intmWeight_j^{\step+1}})}, \label{eq:robust_multiplier_update}
\end{align}
where $\decayPrm\ (0\!<\!\decayPrm\!\ll\!1)$ is a hyperparameter that controls the decay rate of the multiplier function.
The effect of the stabilization coefficient $\decayPrm$ is analyzed in Sec.~\ref{sec:analysis}.

\begin{revhl}{2-4}
Note that $\kdCoef$ and $\LagMltpFn_{i,\inval}^\step$ can be effectively omitted from the implementation by employing $\lrKdCoef$ and $\LagMltpFnWithCoef_{i,\inval}^\step$ instead.
In this framework, $\lr$ and $\lrKdCoef$ serve as the primary tunable hyperparameters.
For simplicity, we shall refer to $\lrKdCoef$ and $\LagMltpFnWithCoef_{i,\inval}^\step$ simply as $\kdCoef$ and $\LagMltpFn_{i,\inval}^\step$ in the rest of the paper, including the following analysis and evaluations.
\end{revhl}

\subsection{Communication and computational costs} \label{sec:cost_analysis}
\begin{revhl}{1-2,2-10}
Generally, output sharing is recognized as more communication-efficient than parameter sharing, as the output dimension is typically significantly smaller than the parameter dimension \cite{gad2024communication,sun2024fkd}.
To confirm this, we compare the communication costs of these two options.

Assuming a model with $\NumPrm$ parameters and an output dimension of $\NumClasses$, where parameters and outputs are quantized to $\QuantPrm$ and $\QuantOutput$ bytes, respectively, the communication costs for parameter sharing and output sharing are given by $\QuantPrm \NumPrm$ and $\QuantOutput \NumClasses \Card{\sharedD}$, respectively.
While both $\NumPrm$ and required $\Card{\sharedD}$ tend to increase as tasks become more complex, making a general conclusion difficult, we provide a concrete example to compare the costs of our evaluation settings in Sec.~\ref{sec:evaluation}.
For instance, parameter-sharing requires 6.7\,MB for Fashion-MNIST and 8.6\,MB for CIFAR-10, whereas output-sharing requires only 40\,KB and 200\,KB, respectively.
As the model used is a simple \gls{CNN} model, larger architectures would further widen this gap in favor of output sharing.

Regarding computational cost, \gls{PropAlg} avoids the explicit training of multiplier functions $\LagMltpFn_i$ as neural networks; instead, it only requires storing and updating their output values $\LagMltpFnWithCoef_{i,\inval}$ on the shared dataset $\sharedD$.
Consequently, the additional computation compared to standard \gls{KD}-based \gls{FL} is limited to updating these values using \eqref{eq:robust_multiplier_update}.
While additional memory is needed to store $\LagMltpFnWithCoef_{i,\inval}$ for each $\inval\in\sharedD$ per neighbor, this overhead is negligible relative to the memory capacity of modern \gls{IoT} devices.

Compared to typical parameter-sharing \gls{FL}, \gls{PropAlg} incurs extra computations for the \gls{KD}-based aggregation steps in \eqref{eq:aggregation}.
Assuming consistent batch sizes, this cost is approximately $\Card{\sharedD}/\Card{\DsLocal{i}}$ times that of a local \gls{SGD} step, which is roughly double per communication round in our evaluation settings.
However, this is a reasonable trade-off for the improved performance and stability under severe non-\gls{IID} settings as shown in Sec.~\ref{sec:evaluation}.
\end{revhl}

\section{Interpretation through the Lens of PI Control} \label{sec:analysis}
To elucidate the behavior of \gls{PropAlg} under non-\gls{IID} conditions, specifically the impact and efficacy of the stabilization coefficient $\decayPrm$, we analyze how the model output changes during both local updates and aggregation.
Our analysis begins by modeling the changes in prediction outputs resulting from local \gls{SGD} to examine how data heterogeneity induces variability in update directions.
Subsequently, by reformulating the aggregation step, we demonstrate that the stabilization coefficient $\decayPrm$ functions analogously to the integral term in \gls{PI} control.
This control-theoretic perspective provides fundamental insight into how the proposed method suppresses variance in model updates caused by statistical heterogeneity across devices.

\begin{table}[!t]
\caption{\begin{revhl}{2-6}Variable definitions\end{revhl}}
\label{tbl:notation}
\centering
\begin{tabular}{cc}
\toprule
Symbol & Definition \\
\midrule
$\outval_i^\step$                           & Output value $\fn(\inval;\weight_i^{\step})$ of device $i$ at round $\step$ \\
$\updOutval_i^\step$                        & Output value $\fn(\inval;\intmWeight_i^{\step})$ after local \gls{SGD} update \\
\multirow{2}{*}{$\Delta \elemloss_i^\step$} & Change in the element-wise loss \\
                                            & $\elemloss\(\updOutval_i^{\step}, \fngt(\inval)\) - \elemloss\(\outval_i^{\step}, \fngt(\inval)\)$ \\
$\StackedOutVal^\step$                      & Output vector $\[\outval_1^\step, \ldots, \outval_\NumDev^\step\]^\Trans$ \\
$\StackedUpdOutVal^\step$                   & Output vector after local \gls{SGD} update $\[\updOutval_1^\step, \ldots, \updOutval_\NumDev^\step\]^\Trans$ \\
$\StackedDiffY^\step$                       & Changes in output vector after local \gls{SGD} $\StackedUpdOutVal^{\step} - \StackedOutVal^{\step}$\\
$\StackedExpDiffY^\step$, $\noiseSgd^\step$ & Expected value and noise term of $\StackedDiffY^\step$ \\
$\noiseKd^\step$                            & Noise vector caused by the \gls{KD}-based aggregation step \\
$\Dist^\step$, $\updDist^\step$             & Edge-wise discrepancy vectors $\IncidenceT \StackedOutVal^\step$, $\IncidenceT \StackedUpdOutVal^\step$ \\
$\multiplier^\step$                         & Lagrange multiplier vector $\[\LagMltpFn_1(\inval), \ldots, \LagMltpFn_\NumDev(\inval)\]^\Trans$\\
$\updateTerm^\step$                         & Update term $\Dist^{\step+1} - \Dist^\step$ \\
$\MatCoef$                                  & Coefficient matrix $\IncidenceT \InvDeg \Incidence$ \\
\bottomrule
\end{tabular}
\end{table}

In this section, we focus on the model output for a specific input $\inval$.
The notations used in this section are summarized in Table~\ref{tbl:notation}.
\begin{revhl}{1-8}
We employ boldface to denote network-wide vectors, i.e., concatenations across all devices.
Although individual outputs $\outval_i^\step$ are generally multi-dimensional, they are represented here using non-boldface notation because our analysis prioritizes the consensus behavior between devices over local values' characteristics.
\end{revhl}
Note that the definitions of the graph-related matrices $\DegMat$, $\AdjMat$, $\Laplacian$, and $\Incidence$ are provided in Section~\ref{sec:problem}.

We first consider the change in the element-wise loss $\elemloss$ at device $i$ denoted as $\Delta \elemloss_i^\step$, which results from the local \gls{SGD}-based parameter update:
\begin{revhl}{2-5}\end{revhl}
\begin{align}
  \Delta \elemloss_i^\step
    &\defeq \elemloss\(\updOutval_i^{\step}, \fngt(\inval)\) - \elemloss\(\outval_i^{\step}, \fngt(\inval)\) \nonumber \\
    &\approx \begin{dcases*}
      - \lr \left\|\frac{\partial \elemloss}{\partial \weight_i^\step}\right\|^2
          - \lr \sum_{\revhlnon{\inval^{\prime}\in\DsLocal{i}{\setminus}\{\!\inval\!\}}}
                  \left.\frac{\partial \elemloss}{\partial \weight_i^\step}\right|_\inval
            \cdot \left.\frac{\partial \elemloss}{\partial \weight_i^\step}\right|_{\inval^\prime} \\
            \hspace{13em} \textrm{if } \inval \in \DsLocal{i} \\
              - \revhlnon{\lr} \sum_{\inval^{\prime}\in\DsLocal{i}}
                  \left.\frac{\partial \elemloss}{\partial \weight_i^\step}\right|_\inval
            \cdot \left.\frac{\partial \elemloss}{\partial \weight_i^\step}\right|_{\inval^\prime} \\
            \hspace{13em} \textrm{otherwise,} \label{eq:loss_update}
    \end{dcases*}
\end{align}
where $\DsLocal{i}{\setminus}\{\inval\}$ denotes the local dataset excluding $\inval$.
The derivation of the above equation is provided in Appendix~\ref{sec:derivation_loss_update}.

\begin{revhl}{1-5}
While the first term in \eqref{eq:loss_update} for the case $\inval \in \DsLocal{i}$ is strictly non-positive, the second term can be either positive or negative.
Since the second term represents the inner product of the gradients corresponding to $\inval$ and other samples $\inval^{\prime}$, it is expected to be positive when $\inval$ and $\inval^{\prime}$ are similar or belong to the same category.
Conversely, when $\inval$ and $\inval^{\prime}$ are dissimilar, the inner product does not exhibit a consistent tendency to be positive or negative.
Consequently, the element-wise loss is expected to decrease primarily within the region supported by the local distribution $\PdfLocal{i}(x)$.
\end{revhl}

\begin{figure}[!t]
\centering
\includegraphics[width=0.4\textwidth]{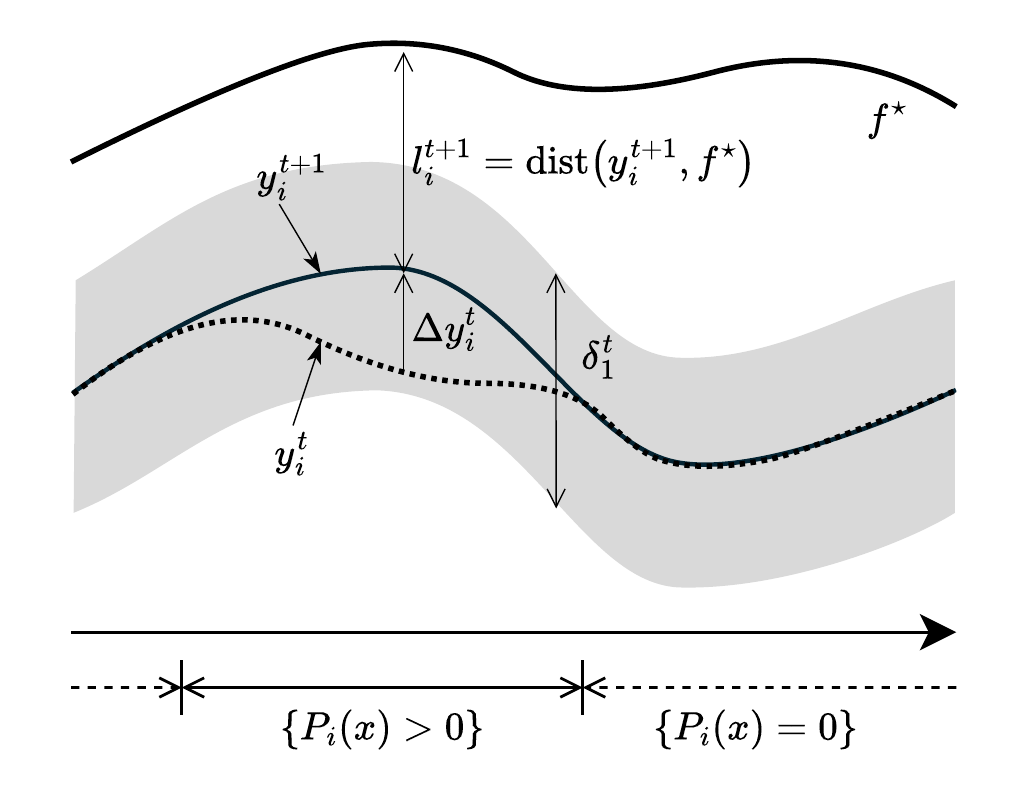}
\caption{Updates of the output values in non-\gls{IID} settings. The element-wise loss is expected to decrease only within the region supported by the local distribution $\PdfLocal{i}(x)$.}
\label{fig:grad_concept}
\end{figure}

From this observation, we model the update process of the network-wide output vector $\StackedOutVal^{\step}$.
Let $\StackedDiffY^\step$ denote the change in $\StackedOutVal^\step$ caused by the local \gls{SGD}-based update, i.e., $\StackedDiffY^\step \defeq \StackedUpdOutVal^\step - \StackedOutVal^\step$.
We decompose $\StackedDiffY^\step$ into its expected values $\StackedExpDiffY^\step$ and the noise term $\noiseSgd^\step$ as follows:
\begin{align}
  \StackedDiffY^\step &= \StackedExpDiffY^\step + \noiseSgd^\step. \label{eq:def_diff_y}
\end{align}
Considering the correspondence between \eqref{eq:loss_update} and \eqref{eq:def_diff_y}, $\StackedExpDiffY^\step$ is expected to point in the direction that decreases the element-wise loss for inputs $x$ included in the local distribution $\PdfLocal{i}(x)$.
In contrast, for inputs $x$ not present in $\PdfLocal{i}(x)$, $\StackedExpDiffY^\step$ are expected to be nearly zero.
This concept is illustrated in Fig.~\ref{fig:grad_concept}.

Next, we derive the change in the output vector $\StackedOutVal^\step$ caused by the aggregation step of \gls{PropAlg}.
Let $\multiplier^\step\defeq\[\LagMltpFn_1(\inval), \ldots, \LagMltpFn_\NumDev(\inval)\]^\Trans$ denote the Lagrange multiplier vector; consequently, from \eqref{eq:robust_multiplier_update}, $\multiplier^\step$ is updated as follows:
\begin{align}
  \multiplier^\step &= (1-\decayPrm) \multiplier^{\step-1} + \InvDeg \Laplacian \StackedUpdOutVal^\step \\
                    &= (1-\decayPrm) \multiplier^{\step-1} + \frac{1}{2} \InvDeg \Incidence \updDist^\step, \label{eq:multiplier_update}
\end{align}
where $\updDist^\step \defeq \IncidenceT \StackedUpdOutVal^\step$ denotes the differences in output values $\StackedUpdOutVal^\step$ across adjacent devices.

After the \gls{KD}-based aggregation step \eqref{eq:aggregation}, the output vector $\StackedOutVal^\step$ is updated as follows:
\begin{align}
  \StackedOutVal^{\step+1} &= \StackedUpdOutVal^\step - \kdCoef \(\InvDeg \Laplacian \StackedUpdOutVal^\step + \multiplier^\step + \noiseKd^\step\),
\end{align}
where $\noiseKd^\step$ is noise arising from the stochastic nature of \gls{KD}.
Using $\Dist^\step \defeq \IncidenceT \StackedOutVal^\step$, the above equation can be rewritten as follows:
\begin{align}
  \Dist^{\step+1} &= \updDist^\step - \kdCoef \(\frac{1}{2} \IncidenceT \InvDeg \Incidence \updDist^\step + \IncidenceT \multiplier^\step + \IncidenceT \noiseKd^\step\). \label{eq:dist_update_aggregation}
\end{align}
Finally, the update term $\updateTerm^\step\defeq \Dist^{\step+1} - \Dist^\step$ can be obtained as follows:
\begin{align}
  \updateTerm^\step
           &= \underbrace{\IncidenceT \StackedExpDiffY^\step}_{\text{SGD}}
            + \underbrace{
                \decayPrm \(\ZeroVec-\Dist^\step\)
              + \kdCoef \(\ZeroVec - \MatCoef \updDist^\step\)
            }_{\text{proportional control}} \nonumber \\
    \mbox{}&+ \underbrace{\decayPrm \sum_{\tmpStep=1}^{\step-1} \IncidenceT \StackedExpDiffY^\tmpStep}_{\text{momentum}}
            + \underbrace{
                \kdCoef \frac{1+\decayPrm}{2} \sum_{\tmpStep=1}^{\step-1} \(\ZeroVec - \MatCoef \updDist^{\tmpStep}\)
            }_{\text{integral control}} \nonumber \\
    \mbox{}&+ \underbrace{
                \IncidenceT \(\noiseSgd^\step - \kdCoef \noiseKd^\step
                    + \decayPrm \sum_{\tmpStep=1}^{\step-1} \(\noiseSgd^\tmpStep - \kdCoef\noiseKd^\tmpStep\)\)
            }_{\text{noise}}, \label{eq:dist_update}
\end{align}
where $\MatCoef$ is defined as $\MatCoef\defeq\IncidenceT \InvDeg \Incidence$.
Detailed derivations of $\updateTerm^\step$ are provided in Appendix~\ref{sec:derivation}.

Interestingly, the update dynamics derived in \eqref{eq:dist_update} exhibit structural equivalence to a \gls{PI} control scheme.
The second and third terms on the \gls{RHS} of \eqref{eq:dist_update} function as proportional control, acting to reduce the current discrepancy $\Dist^\step$ so that local models eventually reach consensus.
The fifth term, serves as integral control, accumulating past discrepancies and enforcing long-term consistency by eliminating residual steady-state errors caused by data heterogeneity.
While the last term represents stochastic fluctuations from \gls{SGD} and \gls{KD}.
In contrast, the first and fourth terms, which originated from the expected effect of \gls{SGD}-based local updates, are aligned with each device's local loss landscape rather than with the consensus objective.
From the perspective of \gls{PI}-based consensus control, these two terms act as disturbances that push model outputs in device-specific directions that reflect heterogeneous data distributions.
Notably the fourth term also provides a momentum-like effect by accumulating gradients over time.

In non-\gls{IID} scenarios where some devices lack training data for certain categories, $\StackedExpDiffY^\tmpStep$ becomes nearly zero, as previously discussed.
In such cases, learning for the missing categories proceeds primarily through the consensus mechanism, which propagates information from neighbors that possess the relevant data.
Consequently, the proportional and integral control terms that explicitly drive $\Dist^\tmpStep$ toward zero play a crucial role in propagating information across the network.
While the third and fifth terms also contribute to consistency indirectly by reducing $\MatCoef \updDist^{\tmpStep}$ (see Appendix~\ref{sec:reason_indirect_consensus}), the second term most directly promotes alignment between neighboring models.
Consequently, a positive stabilization coefficient $\decayPrm$ is particularly important in non-\gls{IID} scenarios, as it enforces output matching among neighboring devices, enabling them to effectively share knowledge about missing categories through prediction-level consensus.

\section{Performance evaluation} \label{sec:evaluation}

\subsection{Simulation setup} \label{sec:sim_setup}
We evaluated the performance of the proposed \gls{PropAlg} under non-\gls{IID} settings using the Fashion-MNIST and CIFAR-10 datasets.
The simulation parameters are summarized in Table~\ref{tbl:sim_param}.
\begin{revhl}{2-7}
In all experiments, each device employed the same model architecture consisting of two \gls{CNN} layers and two fully connected layers, with ReLU activation functions and layer normalization.
The detailed architecture of the model is presented in Table~\ref{tbl:cnn_arch}.
\end{revhl}
\begin{revhl}{1-7,2-9}
In \gls{PropAlg}, we set $\decayPrm = 0.01$ and optimized $\lr$ and $\kdCoef$ via Optuna \cite{Akiba2019-jz} within the ranges shown in Table~\ref{tbl:sim_param}.
The parameters $\lr$ and $\kdCoef$ were fixed during individual training runs.
For other algorithms, as the relevant hyperparameters differ, we specify their respective search spaces in Sec.~\ref{sec:comparison_algorithms}.
\end{revhl}

As label-imbalanced settings, we considered two scenarios, namely 1-class and 2-class distributions, where each device is assigned data sampled from only one or two classes, respectively.
The motivation for evaluating the 1-class distribution is that it represents the most severe case: the 1-class distribution exhibits the largest \gls{KL}-divergence from the uniform distribution.
From an entropy perspective, this corresponds to the lowest possible entropy of the label distribution, since all probability mass is concentrated on a single class.
In contrast, more balanced distributions, such as those generated by a Dirichlet prior, yield higher entropy and smaller divergence from the uniform distribution.
Therefore, the 1-class setting constitutes a more challenging benchmark than the Dirichlet-based distributions that are commonly adopted for evaluating \gls{FL} algorithms.

\begin{table}[!t]
\caption{\begin{revhl}{1-7,2-9}\end{revhl}Simulation parameters.}
\label{tbl:sim_param}
\centering
\begin{tabular}{cc}
\toprule
Parameters & Values / \revhlnon{Search spaces} \\
\midrule
Num. devices & 10 \\
Network topology & Ring \\
\multirow{2}{*}{Num. total training samples $\sum_{i\in\Devices} \Card{\DsLocal{i}}$} & Fashion-MNIST 10000 \\
                                                                                      & CIFAR-10 50000 \\
\multirow{2}{*}{Num. shared samples $\Card{\sharedD}$} & Fashion-MNIST 1000 \\
                                                       & CIFAR-10 5000 \\
Optimizer & SGD \\
Num. communication rounds & 2000 \\
\revhlnon{Stabilization coefficient $\decayPrm$} & \revhlnon{0.01} \\
\revhlnon{Learning rate $\lr$} & \revhlnon{$[10^{-4}, 5{\times}10^{-2}]$} \\
\revhlnon{Penalty parameter $\kdCoef$} & \revhlnon{$[10^{-4}, 10^{0}]$} \\
\bottomrule
\end{tabular}
\end{table}

\begin{table}[!t]
    \caption{\begin{revhl}{2-7}CNN architecture used in the experiments.\end{revhl}}
    \label{tbl:cnn_arch}
    \centering
    \begin{tabular}{llcc} 
        \toprule
        \multirow{2}{*}{Layers} & \multirow{2}{*}{Configuration} & \multicolumn{2}{c}{Num. parameters} \\
        \cmidrule(lr){3-4}
                                &                                & Fashion-MNIST & CIFAR-10 \\
        \midrule
        Conv2D    & Kernel: $5\!\times\!5$, Stride: 1 & 832       & 2,432     \\
        LayerNorm & ---                               & 64        & 64        \\
        Dropout   & Rate: 0.4                         & ---       & ---       \\
        Conv2D    & Kernel: $5\!\times\!5$, Stride: 1 & 51,264    & 51,264    \\
        LayerNorm & ---                               & 128       & 128       \\
        Dropout   & Rate: 0.4                         & ---       & ---       \\
        Dense     & Units: 512                        & 1,606,144 & 2,097,664 \\
        Dropout   & Rate: 0.2                         & ---       & ---       \\
        Dense     & Units: 10                         & 5,130     & 5,130     \\
        \midrule
        \textbf{Total} &                              & 1,663,432 & 2,156,682 \\
                       &                              & (6.7\,MB) & (8.6\,MB) \\
        \bottomrule
    \end{tabular}
\end{table}

\subsection{Comparison algorithms} \label{sec:comparison_algorithms}
We compared the proposed \gls{PropAlg} with conventional algorithms and their decentralized versions.
In the local update step, all algorithms update local models using \gls{SGD} with the local data, whereas the aggregation step differs.

\subsubsection{DecFedAvg}
We implemented a decentralized version of FedAvg \cite{mcmahan2016communication} as a typical algorithm for \gls{DFL}.
With DecFedAvg, each device first updates its local parameters by \gls{SGD} using its local dataset, and then updates its parameters for aggregation using the average of the neighboring devices' parameters, denoted by $\Mean{i}{\weight_j^{\step}}$, as follows:
\begin{align}
  \weight_i^{\step+1} = \(1 - \avgRate\) \weight_i^\step + \avgRate \Mean{i}{\weight_j^{\step}}, \label{eq:averaging_update}
\end{align}
where averaging rate $\avgRate$ is a hyperparameter that controls the rate of parameter averaging.
DecFedAvg is categorized in the same class as \gls{CDSGD} \cite{jiang2017collaborative} and \gls{DPSGD} \cite{lian2017can}.
\begin{revhl}{2-9}
Hyperparameters $\lr$ and $\avgRate$ were optimized within the ranges of $[10^{-4}, 10^{-1}]$ and $[10^{-1}, 8{\times}10^{-1}]$, respectively.
\end{revhl}


\subsubsection{DecFedProx}
We implemented a decentralized version of FedProx \cite{li2020federated} as a comparison algorithm.
FedProx introduces a proximal term to control the divergence of local models from the global model.
Since the global model does not exist in \gls{DFL} settings, we replace the global model with the average of the neighboring devices' parameters; thus, the update rule of the gradient descent step is given by:
\begin{align}
  \weight_i^{\step+1} = \weight_i^\step - \lr \nabla_{\weight_i} \(\CostFn_i(\weight_i^\step) + \FedProxRegRate \left\|\weight_i^\step - \Mean{i}{\weight_j^{\step}}\right\|^2\),
\end{align}
where $\FedProxRegRate$ is the proximal coefficient.
Aggregation is performed in the same manner as DecFedAvg.
\begin{revhl}{2-9}
Hyperparameters $\lr$, $\FedProxRegRate$, and $\avgRate$ were optimized within the ranges of $[10^{-4}, 10^{-2}]$, $[10^{-2}, 5{\times}10^{-1}]$, and $[10^{-1}, 8{\times}10^{-1}]$, respectively.
\end{revhl}

\begin{revhl}{1-4}
\subsubsection{DFedAvgM}
We evaluated DFedAvgM \cite{sun2022decentralized} as a decentralized \gls{FL} algorithm that incorporates a momentum term into the parameter averaging process to accelerate training.
\begin{revhl}{2-9}
We optimized the momentum parameter $\momentumRate$ in the ranges of $[0.8, 0.99]$, in addition to $\lr$ in $[10^{-4}, 10^{-1}]$ and $\avgRate$ in $[10^{-2}, 5{\times}10^{-1}]$.
\end{revhl}

\subsubsection{DFedADMM}
As a parameter-level \gls{ADMM}-based \gls{DFL}, we evaluated DFedADMM \cite{li2025dfedadmm}.
\begin{revhl}{2-9}
There are three hyperparameters in DFedADMM: learning rate $\lr$, penalty weight $\admmPenalty$, and averaging rate $\avgRate$, which were optimized within the ranges of $[10^{-4}, 10^{-1}]$, $[10^{-2}, 5{\times}10^{0}]$, and $[10^{-2}, 5{\times}10^{-1}]$, respectively.
\end{revhl}
\end{revhl}

\subsubsection{CMFD}
\Gls{CMFD} \cite{taya2022decentralized} is a \gls{KD}-based function-level consensus algorithm for \gls{DFL}.
This algorithm requires a shared dataset for model aggregation using \gls{KD}.
In each communication round, local models are first updated by \gls{SGD} using local datasets, and then each device performs \gls{KD}-based aggregation using the shared dataset.
The local parameters are updated in the aggregation step as follows:
\begin{align}
  \weight_i^{\step+1} = \weight_i^\step - \frac{\kdCoef}{2} \nabla_{\weight_i} \sum_{\inval\in\sharedD} \left\| \fn_{\weight_i^\step}(\inval) - \MeanFn_{\weight_j^\step}(\inval)\right\|^2,
\end{align} 
where $\kdCoef$ denotes the learning rate for the \gls{KD} step.
The difference between \gls{CMFD} and the proposed \gls{PropAlg} is the definition of the virtual targets: the former uses the average of neighboring models' outputs, whereas the latter employs the virtual target $\target{i,\inval}{\step}\defeq \MeanFn_{\weight_i^\step}(\inval) - \LagMltpFn_i^\step(\inval)$, which incorporates the Lagrange multiplier.
\begin{revhl}{2-9}
Hyperparameters $\lr$ and $\kdCoef$ were optimized within the ranges of $[10^{-3}, 10^{-1}]$ and $[10^{-3}, 10^{0}]$, respectively.
\end{revhl}

\def\twidth{0.16\textwidth}
\def\trimtop{60}
\def\trimright{25}

\begin{figure}[!t]
\centering
\subfloat[DecFedAvg]{\includegraphics[width=\twidth,clip,trim=0 0 {\trimright} {\trimtop}]{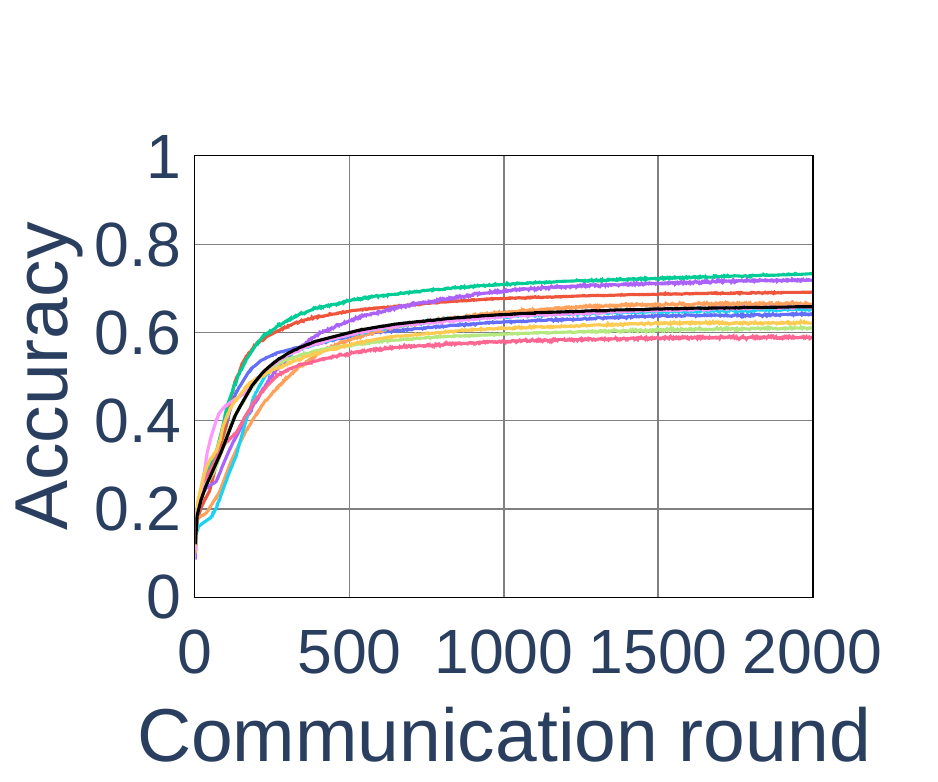}}
\subfloat[DecFedProx]{\includegraphics[width=\twidth,clip,trim=0 0 {\trimright} {\trimtop}]{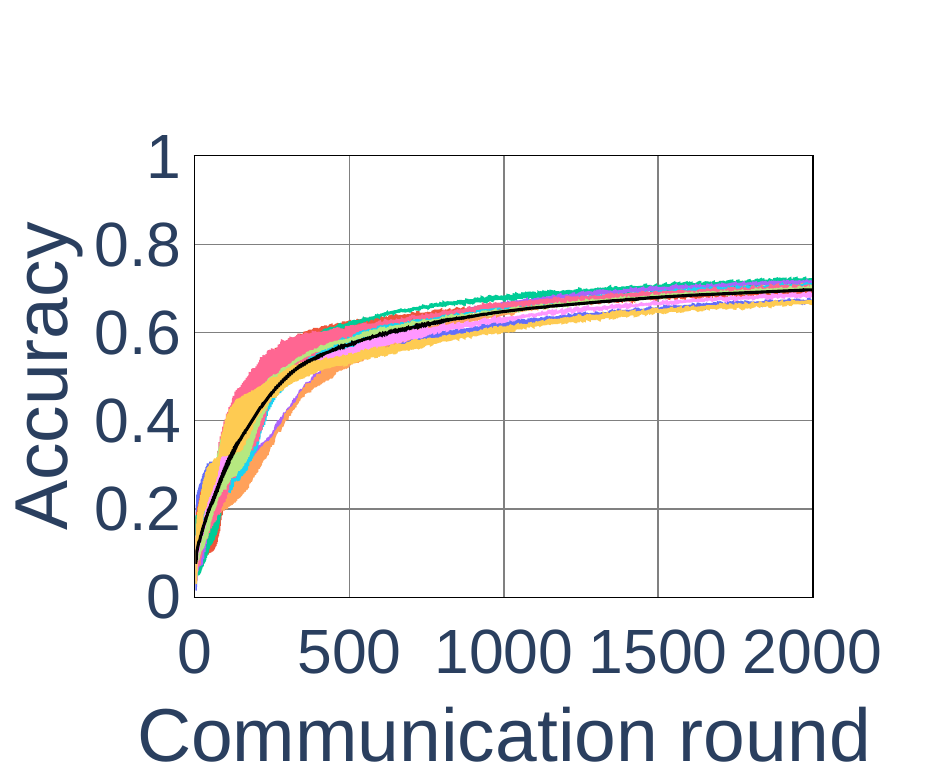}}
\subfloat[\revhlnon{DFedAvgM}]{\includegraphics[width=\twidth,clip,trim=0 0 {\trimright} {\trimtop}]{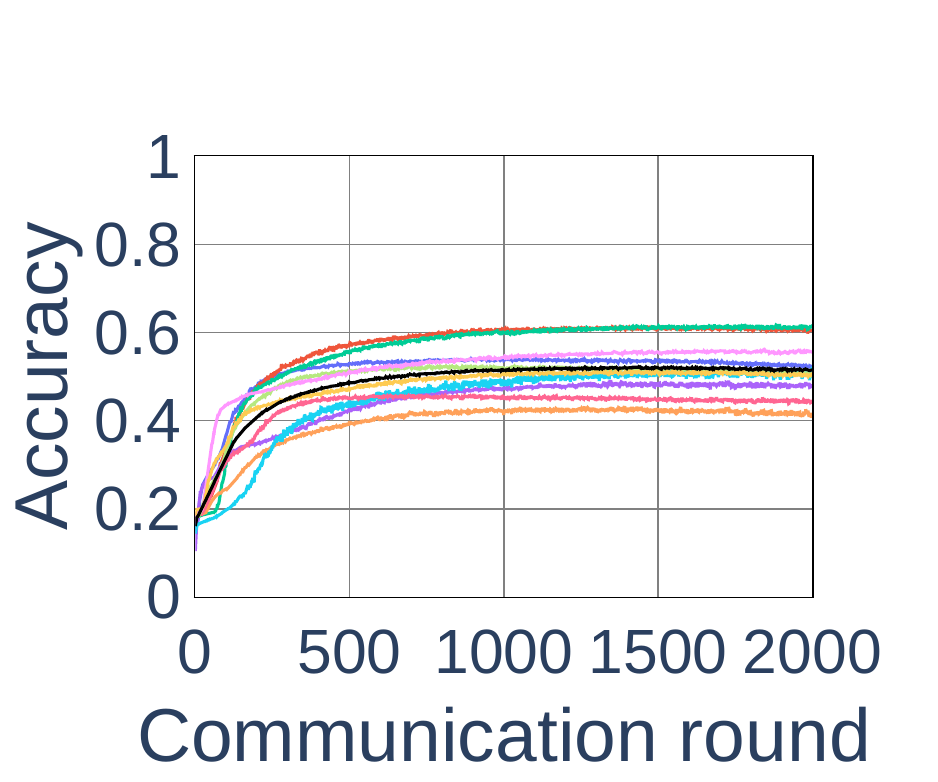}}\\
\subfloat[\revhlnon{DFedADMM}]{\includegraphics[width=\twidth,clip,trim=0 0 {\trimright} {\trimtop}]{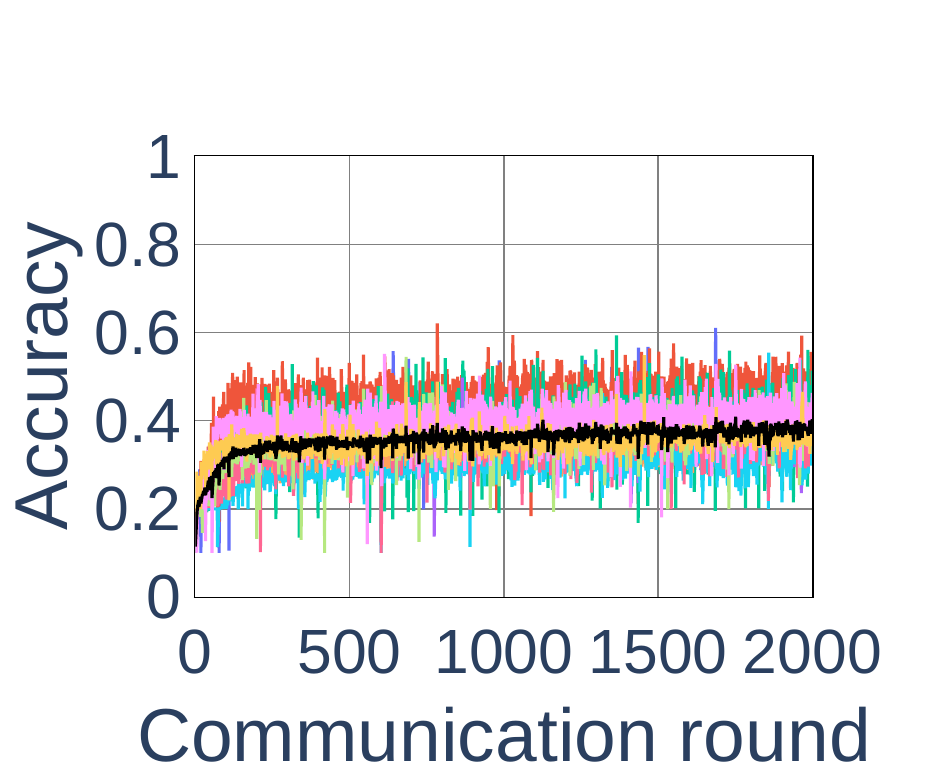}}
\subfloat[\gls{CMFD}]{\includegraphics[width=\twidth,clip,trim=0 0 {\trimright} {\trimtop}]{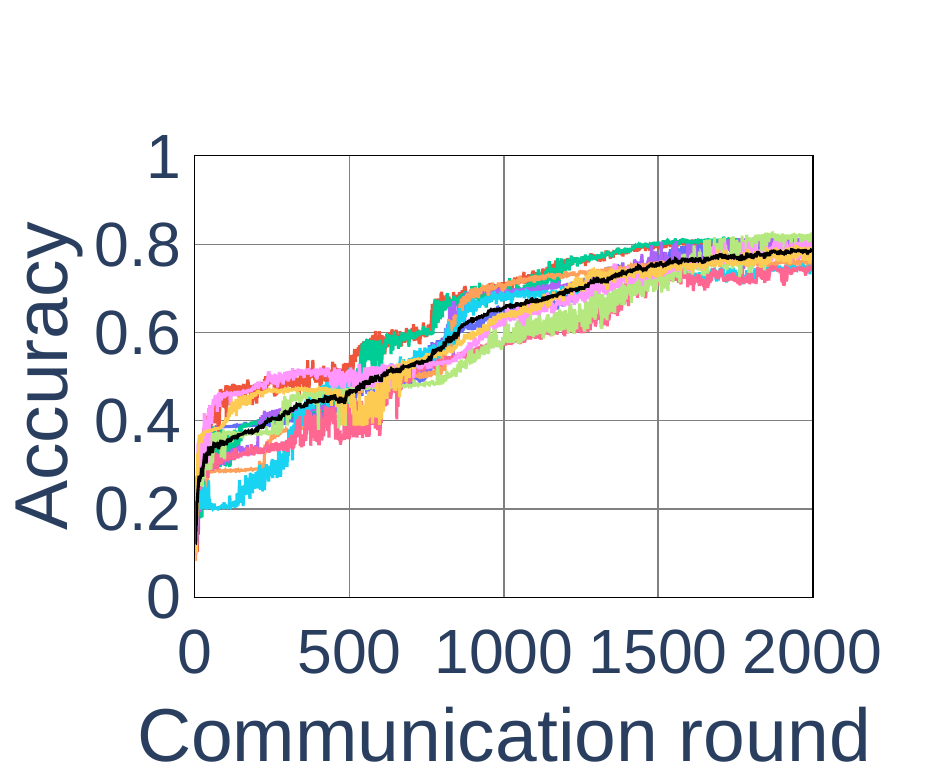}}
\subfloat[\gls{PropAlg} (Proposed)]{\includegraphics[width=\twidth,clip,trim=0 0 {\trimright} {\trimtop}]{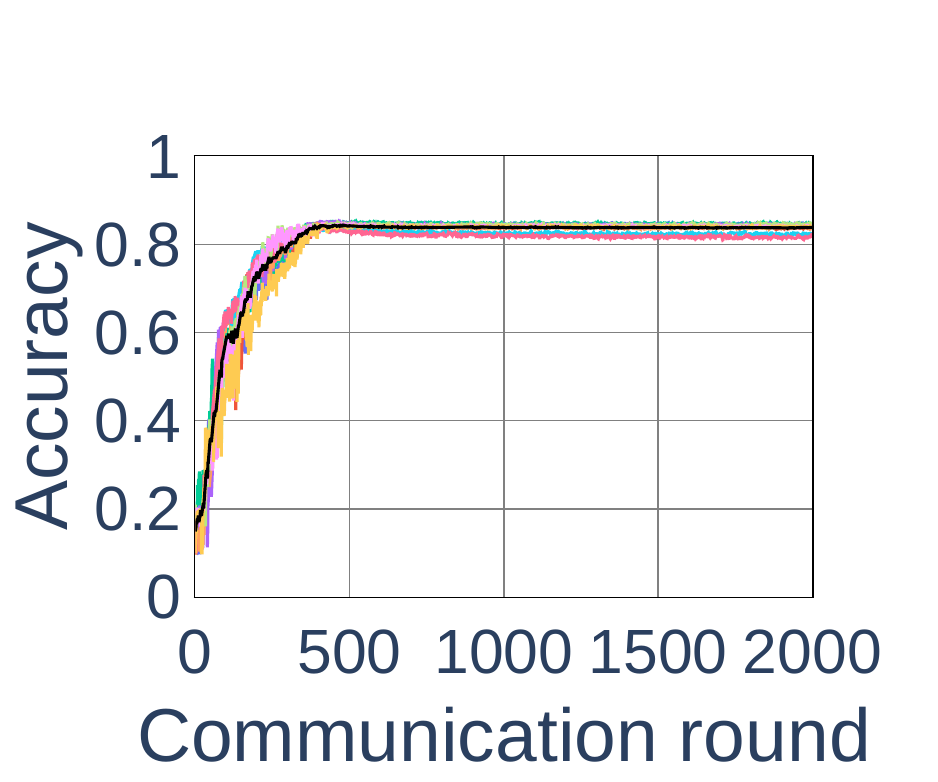}} 
\caption{Convergence performance on Fashion-MNIST. Each device has 500 samples from 2 labels and 1000 unlabeled shared samples. The black line indicates the average accuracy across devices, while colored lines represent individual device accuracies. The proposed \gls{PropAlg} achieves higher accuracy, better consensus, and faster convergence than other algorithms.}
\label{fig:conv_fmnist_500x2}
\end{figure}

\begin{figure}[!t]
\centering
\subfloat[DecFedAvg]{\includegraphics[width=\twidth,clip,trim=0 0 {\trimright} {\trimtop}]{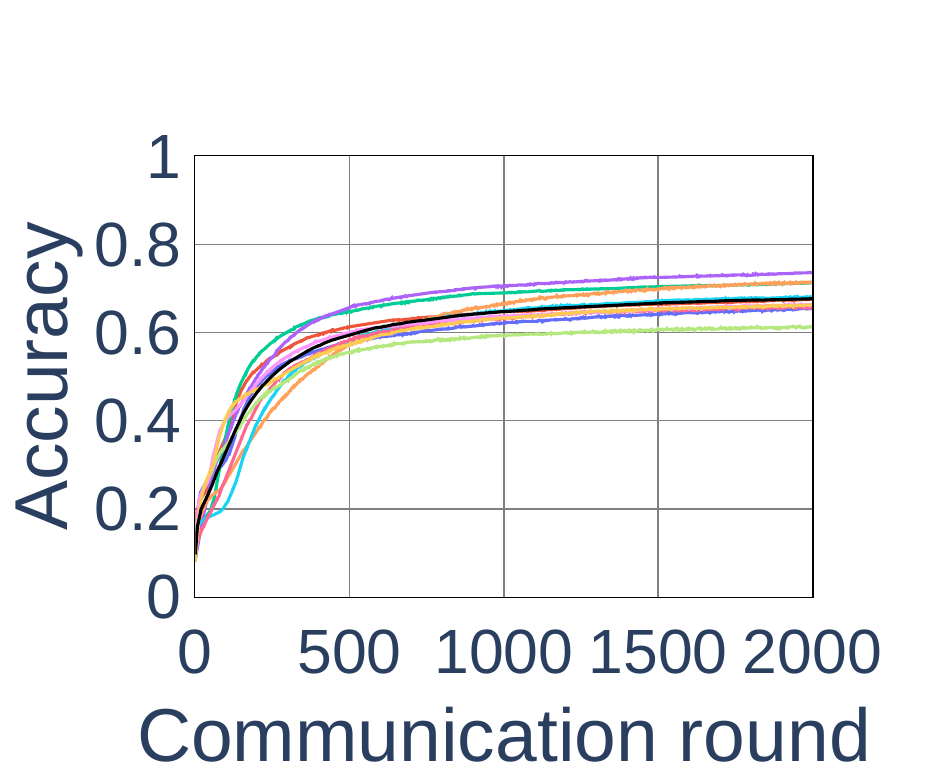}}
\subfloat[DecFedProx]{\includegraphics[width=\twidth,clip,trim=0 0 {\trimright} {\trimtop}]{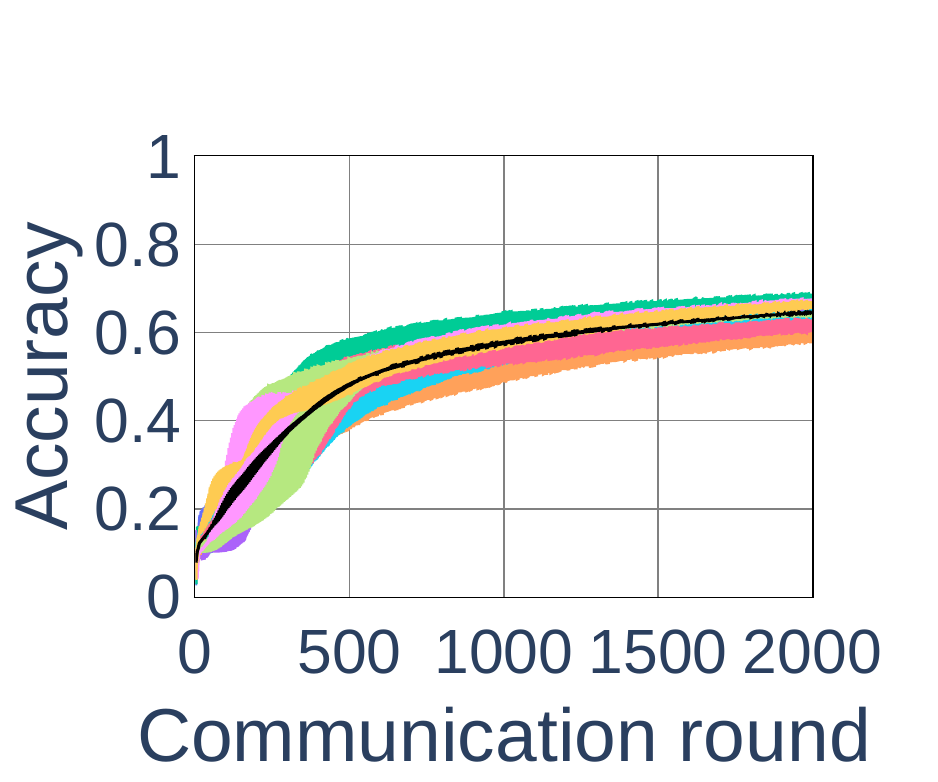}}
\subfloat[\revhlnon{DFedAvgM}]{\includegraphics[width=\twidth,clip,trim=0 0 {\trimright} {\trimtop}]{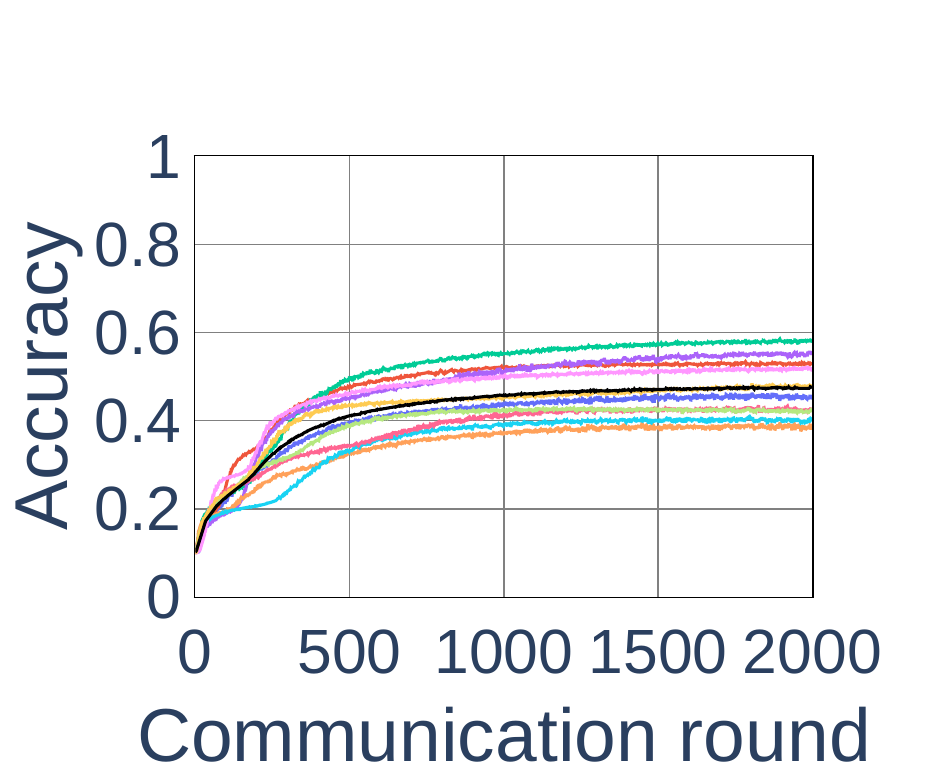}}\\
\subfloat[\revhlnon{DFedADMM}]{\includegraphics[width=\twidth,clip,trim=0 0 {\trimright} {\trimtop}]{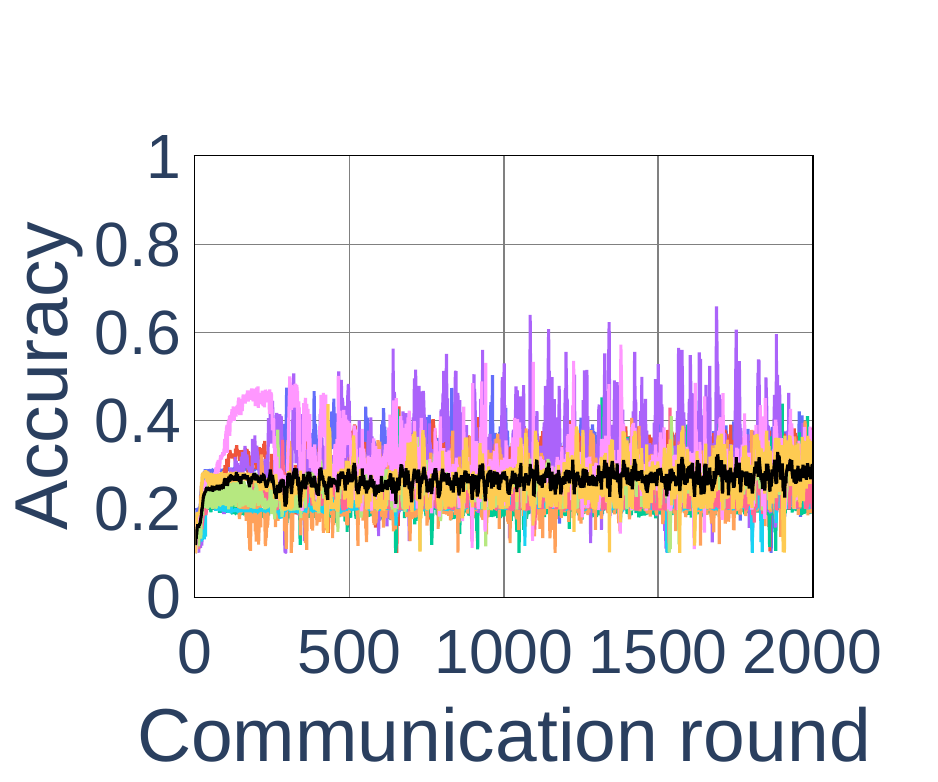}}
\subfloat[\gls{CMFD}]{\includegraphics[width=\twidth,clip,trim=0 0 {\trimright} {\trimtop}]{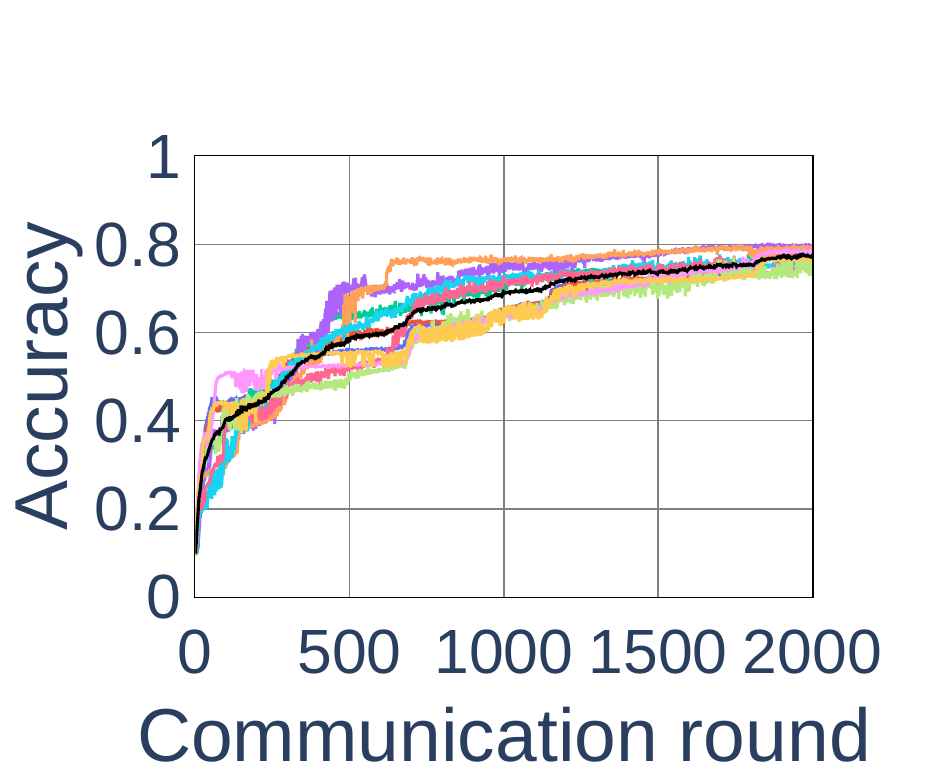}}
\subfloat[\gls{PropAlg} (Proposed)]{\includegraphics[width=\twidth,clip,trim=0 0 {\trimright} {\trimtop}]{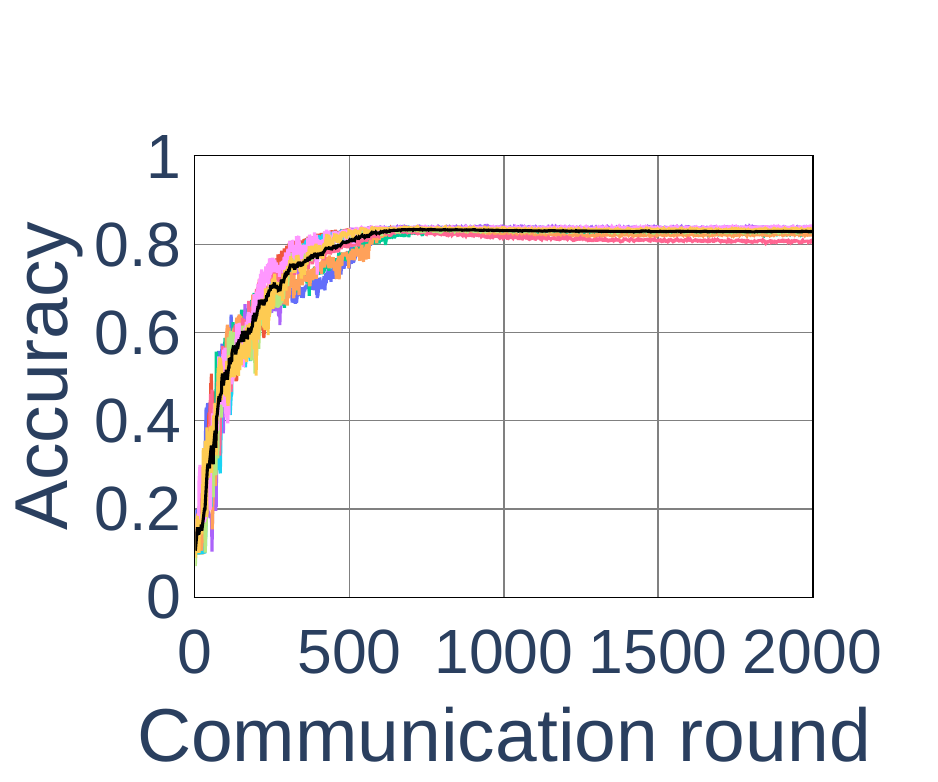}} 
\caption{Convergence performance on Fashion-MNIST. Each device has 1000 samples from 1 label and 1000 unlabeled shared samples. The black and colored lines represent average and individual device accuracies, respectively. Because of the severe data heterogeneity, individual devices exhibit large fluctuations in accuracy in DecFedProx and DFedADMM. In contrast, the proposed \gls{PropAlg} shows stable convergence and outperforms other algorithms.}
\label{fig:conv_fmnist_1000x1}
\end{figure}

\subsection{Convergence performance} \label{sec:conv_performance}
\begin{figure}[!t]
\centering
\subfloat[DecFedAvg]{\includegraphics[width=\twidth,clip,trim=0 0 {\trimright} {\trimtop}]{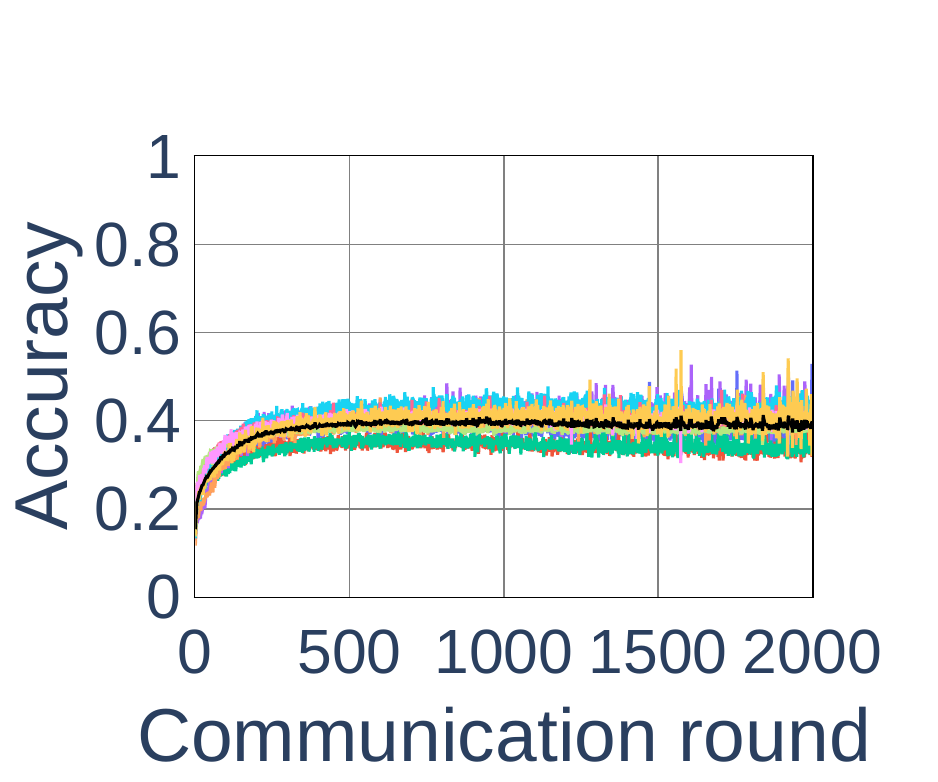}}
\subfloat[DecFedProx]{\includegraphics[width=\twidth,clip,trim=0 0 {\trimright} {\trimtop}]{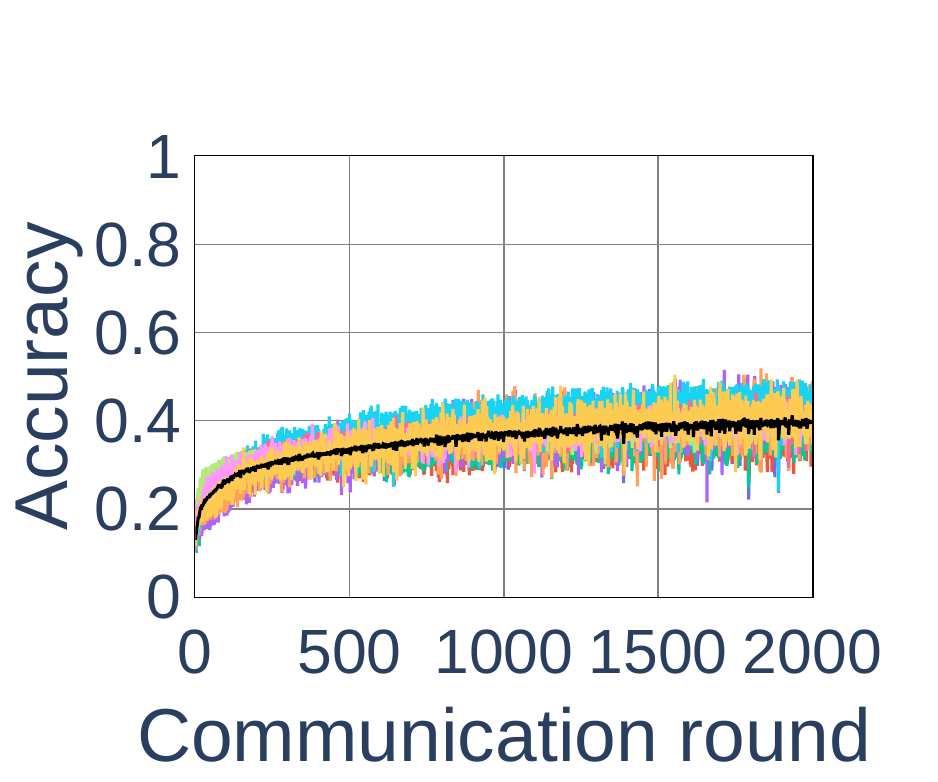}}
\subfloat[\revhlnon{DFedAvgM}]{\includegraphics[width=\twidth,clip,trim=0 0 {\trimright} {\trimtop}]{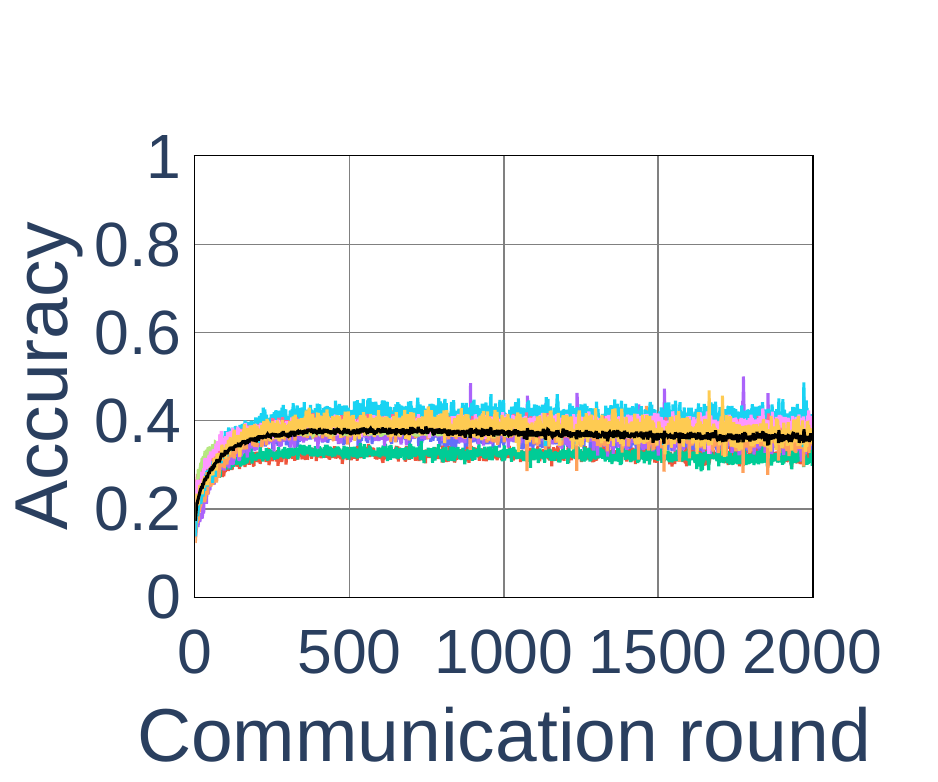}}\\
\subfloat[\revhlnon{DFedADMM}]{\includegraphics[width=\twidth,clip,trim=0 0 {\trimright} {\trimtop}]{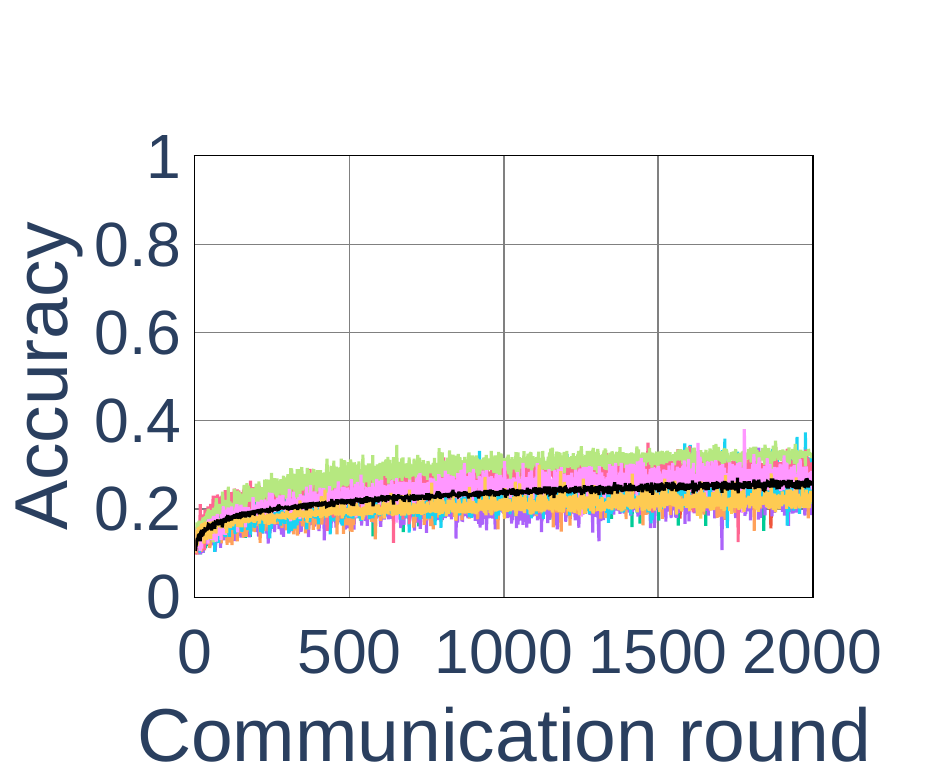}}
\subfloat[\gls{CMFD}]{\includegraphics[width=\twidth,clip,trim=0 0 {\trimright} {\trimtop}]{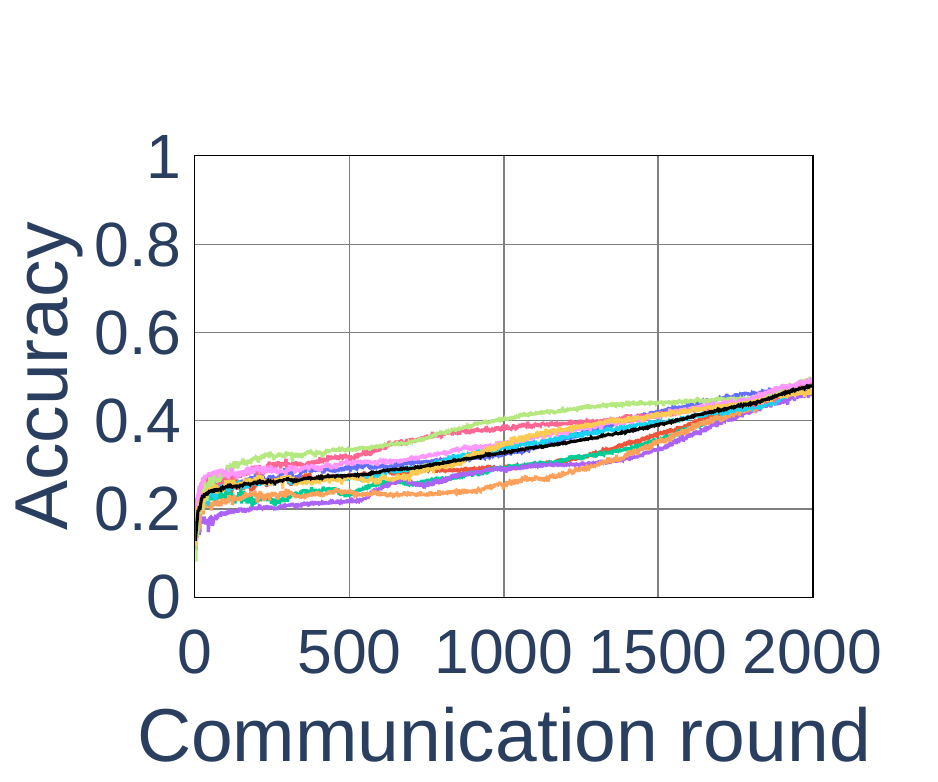}}
\subfloat[\gls{PropAlg} (Proposed)]{\includegraphics[width=\twidth,clip,trim=0 0 {\trimright} {\trimtop}]{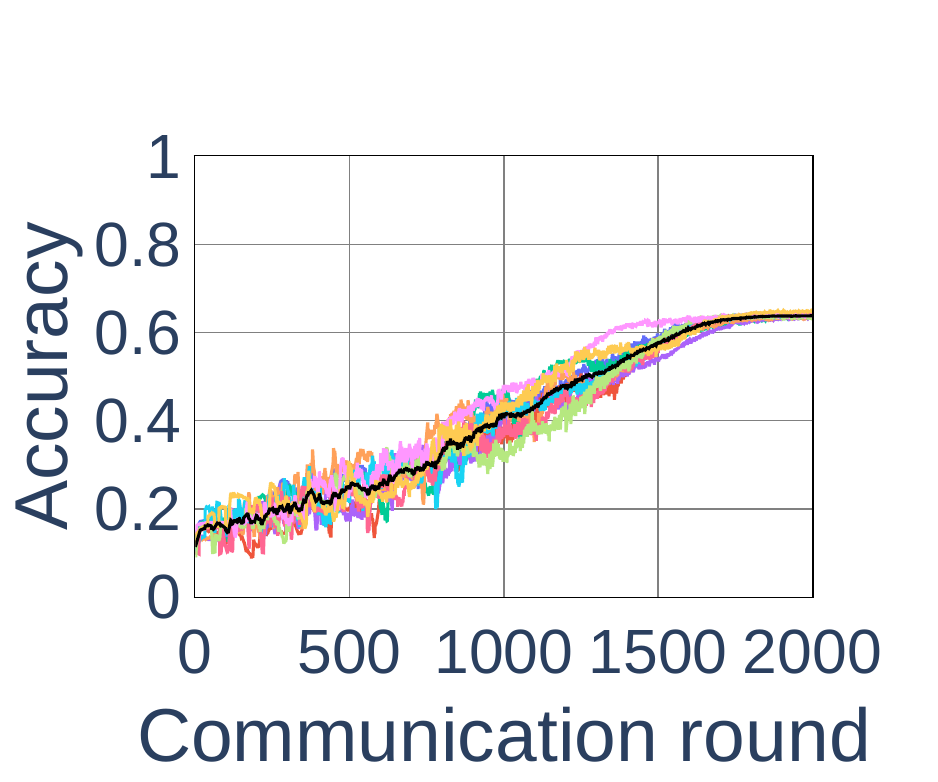}} 
\caption{Convergence performance on CIFAR-10. Each device has 2500 samples from 2 labels and 5000 unlabeled shared samples. The black and colored lines represent average and individual device accuracies, respectively. Compared with Fashion-MNIST results, when using parameter-level aggregation methods (DecFedAvg, DecFedProx, etc.), the accuracies of individual models fluctuate significantly. \Gls{CMFD} shows stable convergence, but the proposed \gls{PropAlg} achieves much higher accuracy and better consensus.}
\label{fig:conv_cifar10_2500x2}
\end{figure}

\begin{figure}[!t]
\centering
\subfloat[DecFedAvg]{\includegraphics[width=\twidth,clip,trim=0 0 {\trimright} {\trimtop}]{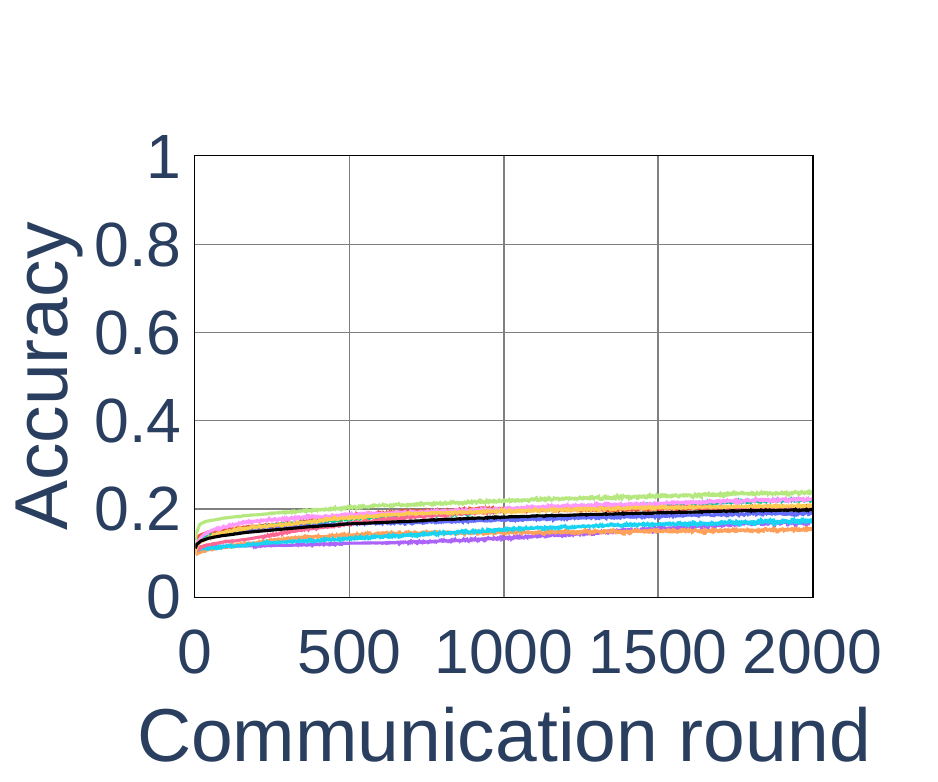}}
\subfloat[DecFedProx]{\includegraphics[width=\twidth,clip,trim=0 0 {\trimright} {\trimtop}]{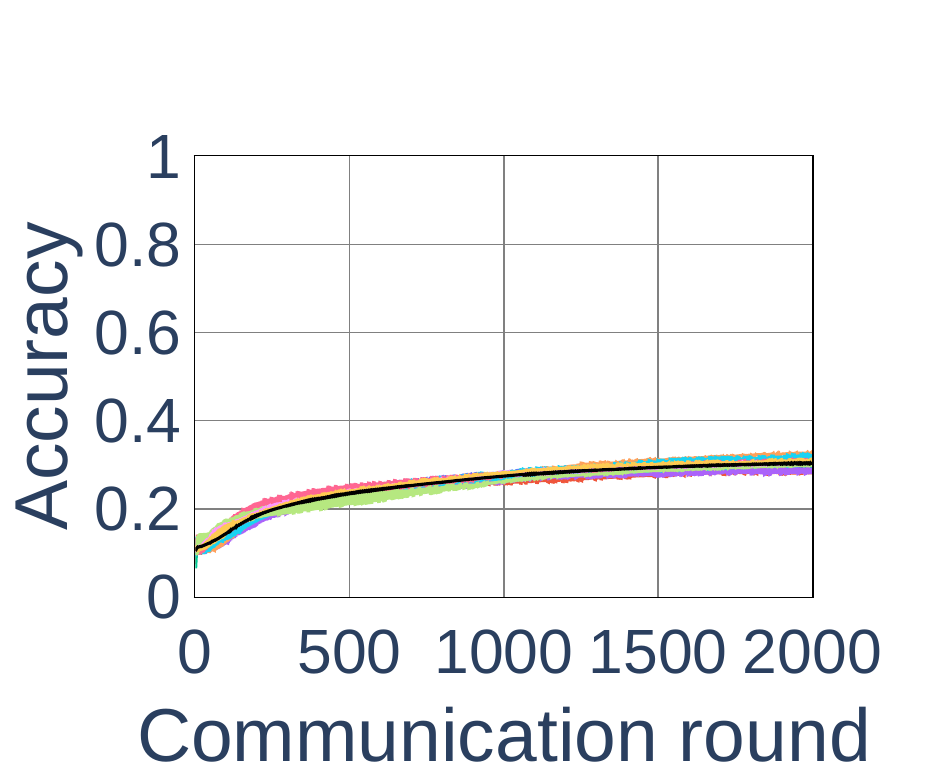}}
\subfloat[\revhlnon{DFedAvgM}]{\includegraphics[width=\twidth,clip,trim=0 0 {\trimright} {\trimtop}]{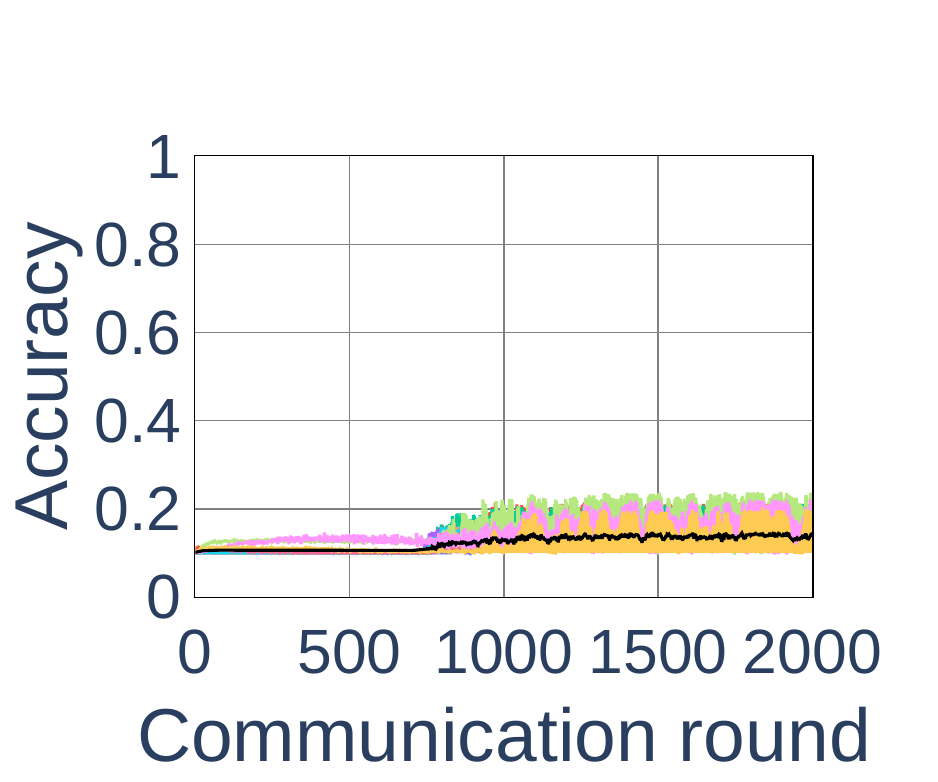}}\\
\subfloat[\revhlnon{DFedADMM}]{\includegraphics[width=\twidth,clip,trim=0 0 {\trimright} {\trimtop}]{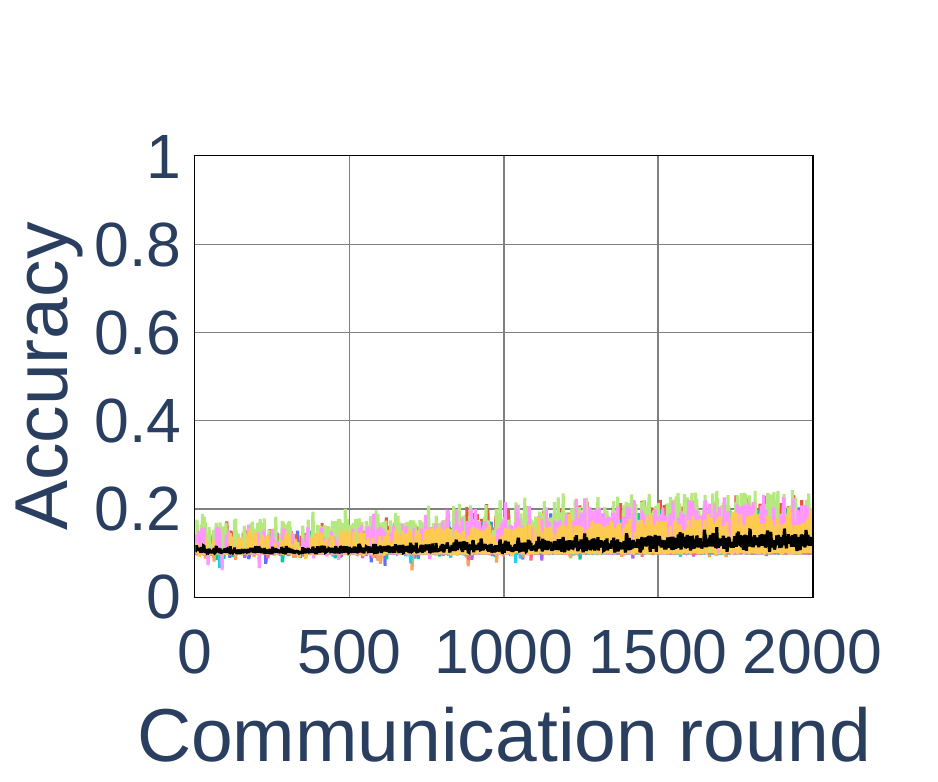}}
\subfloat[\gls{CMFD}]{\includegraphics[width=\twidth,clip,trim=0 0 {\trimright} {\trimtop}]{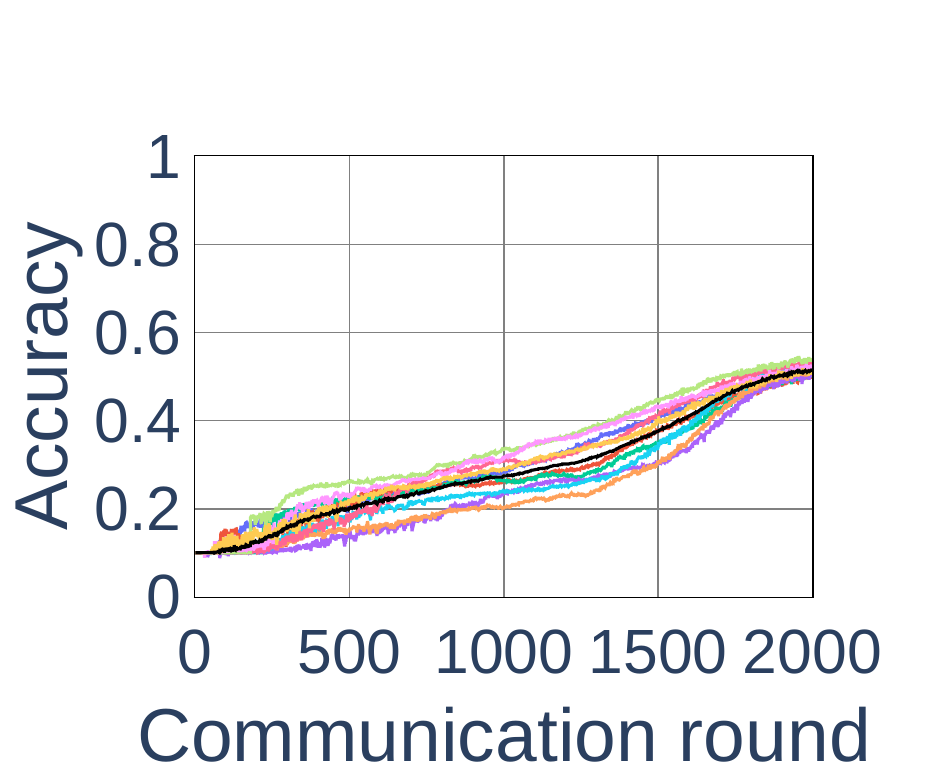}}
\subfloat[\gls{PropAlg} (Proposed)]{\includegraphics[width=\twidth,clip,trim=0 0 {\trimright} {\trimtop}]{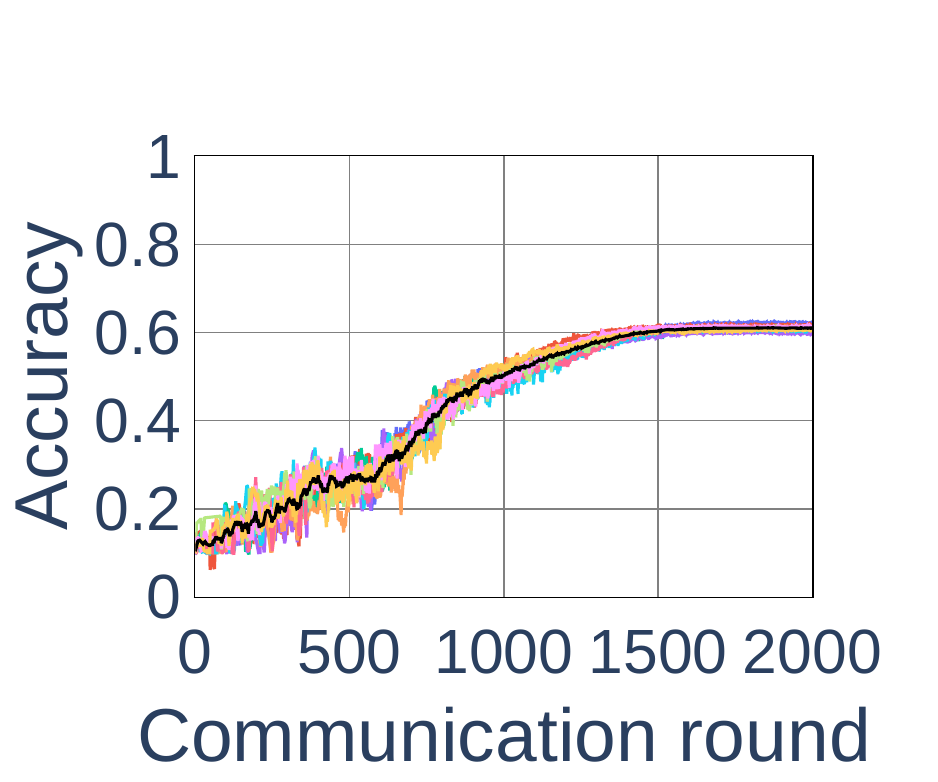}} 
\caption{Convergence performance on CIFAR-10. Each device has 5000 samples from 1 label and 5000 unlabeled shared samples. The black and colored lines represent average and individual device accuracies, respectively. In this severe non-\gls{IID} setting, Parameter-level aggregation methods achieve only low accuracy. Even in such a case, the proposed \gls{PropAlg} does not suffer from significant performance degradation.}
\label{fig:conv_cifar10_5000x1}
\end{figure}

Figs.~\ref{fig:conv_fmnist_500x2}~and~\ref{fig:conv_fmnist_1000x1} present the learning performance on Fashion-MNIST, while Figs.~\ref{fig:conv_cifar10_2500x2} and \ref{fig:conv_cifar10_5000x1} show the convergence performance on CIFAR-10.
In Figs.~\ref{fig:conv_fmnist_500x2} and \ref{fig:conv_cifar10_2500x2}, the dataset is distributed such that each of the 10 devices receives samples from two labels, whereas in Figs.~\ref{fig:conv_fmnist_1000x1} and \ref{fig:conv_cifar10_5000x1} each device is assigned samples from only a single label.
In these experiments, each device holds 1,000 samples for Fashion-MNIST and 5,000 samples for CIFAR-10.
For \gls{CMFD} and the proposed \gls{PropAlg}, we used 1,000 shared samples for Fashion-MNIST and 5,000 shared samples for CIFAR-10 in the \gls{KD} step.
In each figure, eleven lines are plotted: the black line represents the average accuracy across all devices, while the colored lines indicate the accuracies of individual local models.

\begin{revhl}{1-4}
In Figs.~\ref{fig:conv_fmnist_500x2}~and~\ref{fig:conv_fmnist_1000x1}, although the accuracies of individual models appear stable in DecFedAvg, there is a performance gap among devices, indicating poor consensus.
Similar phenomena can be observed in DFedAvgM.
For DecFedProx and DFedADMM, the individual model accuracies fluctuate significantly due to the severe data heterogeneity, as seen in Figs.~\ref{fig:conv_fmnist_1000x1}~and~\ref{fig:conv_cifar10_2500x2}.
Especially, parameter-level algorithms (DecFedAvg, DecFedProx, DFedAvgM, and DFedADMM) could not learn CIFAR-10 under 1-class settings, as shown in Fig.~\ref{fig:conv_cifar10_5000x1}.
In contrast, \gls{CMFD} achieves relatively higher accuracy than the parameter-level algorithms, but it requires a longer time to converge.
\end{revhl}

\begin{figure*}[!t]
\centering
\includegraphics[width=0.95\textwidth]{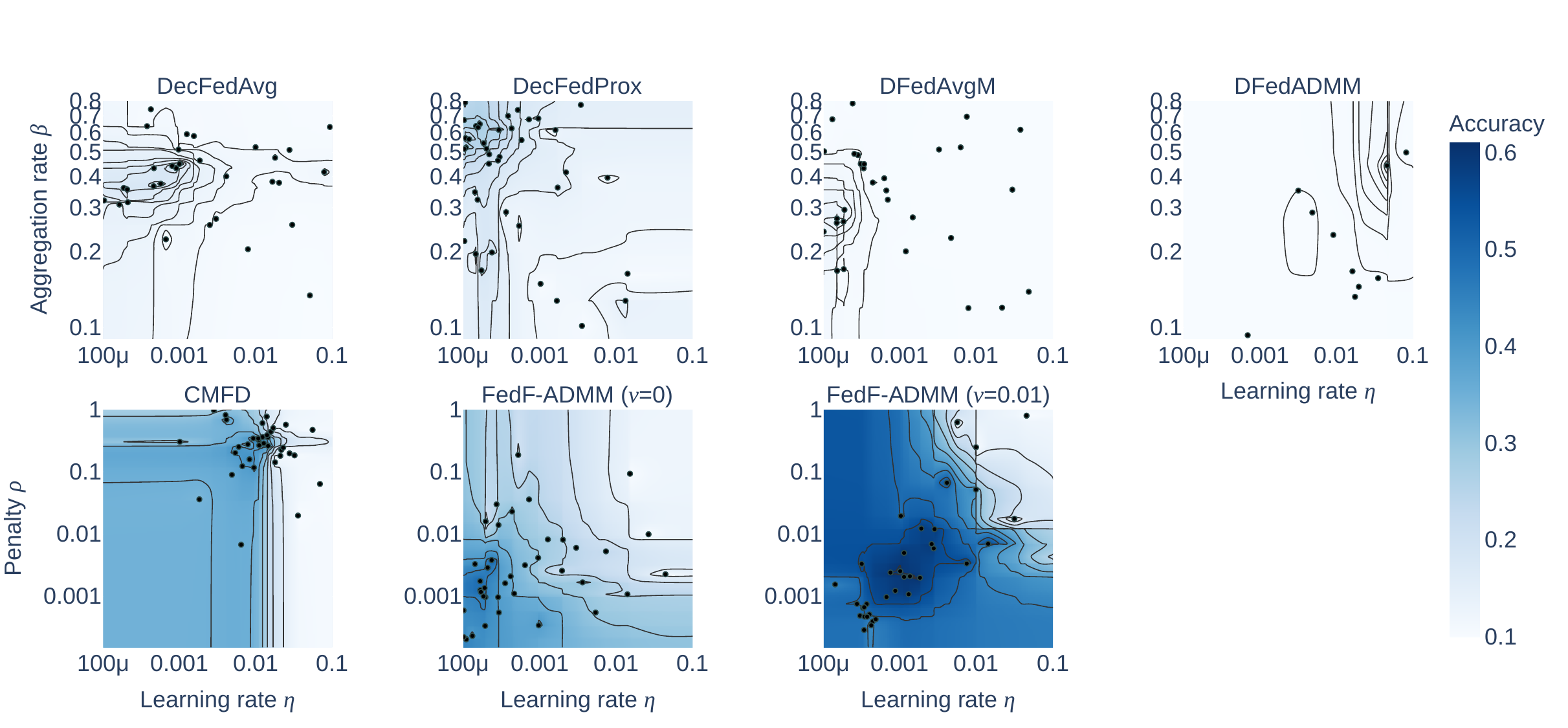}
\caption{Tuning results on CIFAR-10 with the 1-class distribution. \Gls{PropAlg} outperforms the other algorithms in terms of accuracy. In addition, the stabilization coefficient $\decayPrm$ contributes to improving training stability against variations in the other hyperparameters.}
\label{fig:tuning_results}
\end{figure*}

On the other hand, the proposed \gls{PropAlg} exhibits stable convergence across all devices: the accuracies of local models converge to nearly identical values, indicating a small variance among models, while also achieving higher accuracy than existing approaches.
Table~\ref{tbl:final_acc} lists the final accuracy and the accuracy gap, which is defined as the difference between the maximum and minimum accuracies across devices.
It is also notable that, because Fashion-MNIST represents a relatively simple classification task, the performance gap between the 1-class and 2-class settings is negligible.
In contrast, for the more complex CIFAR-10 dataset, algorithms based on parameter aggregation struggle to learn effectively under 1-class settings.
Therefore, the proposed method proves especially effective under severe data heterogeneity, where device data are strongly biased toward specific labels.
Such robustness makes the proposed method suitable for real-world federated learning scenarios in communication and \gls{IoT} systems, where balanced data availability cannot be guaranteed.

Fig.~\ref{fig:tuning_results} shows the heatmaps of the achieved accuracy for each algorithm under the 1-class distribution setting.
As shown in the figures, the proposed \gls{PropAlg} achieves higher accuracy and more stable convergence than the compared algorithms.
It is also notable that the stabilization coefficient $\decayPrm$ contributes to improving training stability against variations in the other hyperparameters.
\begin{revhl}{1-7}
Based on these results, we suggest that the parameters $\lr$ and $\kdCoef$ be set to values within the ranges $[10^{-4}, 10^{-2}]$ and $[10^{-3}, 10^{-2}]$, respectively.
\end{revhl}

\begin{table}[!t]
\caption{Comparison of average accuracy and accuracy gap.}
\label{tbl:final_acc}
\centering
\begin{tabular}{cccc}
\toprule
Settings & Algorithm & Final acc. (\%)  & Acc. gap (pp) \\
\midrule
F-MNIST      & DecFedAvg & 65.9 & 14.3 \\
(500x2)      & DecFedProx & 69.7 & 5.5 \\
             & \revhlnon{DFedAvgM} & \revhlnon{51.6} & \revhlnon{19.7} \\
             & \revhlnon{DFedADMM} & \revhlnon{40.3} & \revhlnon{20.5} \\
             & \gls{CMFD} & 78.4 & 7.0 \\
             & \gls{PropAlg} & \textbf{83.9} & \textbf{2.8} \\
\midrule
F-MNIST      & DecFedAvg & 67.7 & 12.3 \\
(1000x1)     & DecFedProx & 64.6 & 7.7 \\
             & \revhlnon{DFedAvgM} & \revhlnon{47.5} & \revhlnon{19.6} \\
             & \revhlnon{DFedADMM} & \revhlnon{30.2} & \revhlnon{16.1} \\
             & \gls{CMFD} & 77.2 & 6.6 \\
             & \gls{PropAlg} & \textbf{83.0} & \textbf{3.3} \\
\midrule
F-MNIST      & DecFedAvg & \textbf{86.4} & 8.3 \\
(Dirichlet)  & \revhlnon{DFedAvgM} & \revhlnon{83.5} & \revhlnon{7.8} \\
             & \revhlnon{DFedADMM} & \revhlnon{81.5} & \revhlnon{13.7} \\
             & \gls{CMFD} & 84.0 & 4.5 \\
             & \gls{PropAlg} & 84.8 & \textbf{3.0} \\
\midrule
CIFAR-10     & DecFedAvg & 39.8 & 9.5 \\
(2500x2)     & DecFedProx & 39.7 & 15.7 \\
             & \revhlnon{DFedAvgM} & \revhlnon{36.1} & \revhlnon{11.5} \\
             & \revhlnon{DFedADMM} & \revhlnon{26.3} & \revhlnon{9.6} \\
             & \gls{CMFD} & 45.6 & 6.1 \\
             & \gls{PropAlg} & \textbf{63.8} & \textbf{1.6} \\
\midrule
CIFAR-10     & DecFedAvg & 19.9 & 8.8 \\
(5000x1)     & DecFedProx & 30.4 & 4.9 \\
             & \revhlnon{DFedAvgM} & \revhlnon{14.0} & \revhlnon{10.9} \\
             & \revhlnon{DFedADMM} & \revhlnon{11.6} & \revhlnon{6.0} \\
             & \gls{CMFD} & 51.7 & 3.9 \\
             & \gls{PropAlg} & \textbf{61.0} & \textbf{2.4} \\
\midrule
CIFAR-10     & DecFedAvg & 68.5 & 19.7 \\
(Dirichlet)  & \revhlnon{DFedAvgM} & \textbf{\revhlnon{68.9}} & \revhlnon{23.3} \\
             & \revhlnon{DFedADMM} & \revhlnon{64.8} & \revhlnon{15.0} \\
             & \gls{CMFD} & 63.2 & 10.9 \\
             & \gls{PropAlg} & 65.0 & \textbf{5.5} \\
\bottomrule
\end{tabular}
\end{table}

\begin{figure}[!t]
\centering
\subfloat[DecFedAvg]{\includegraphics[width=\twidth,clip,trim=0 0 {\trimright} {\trimtop}]{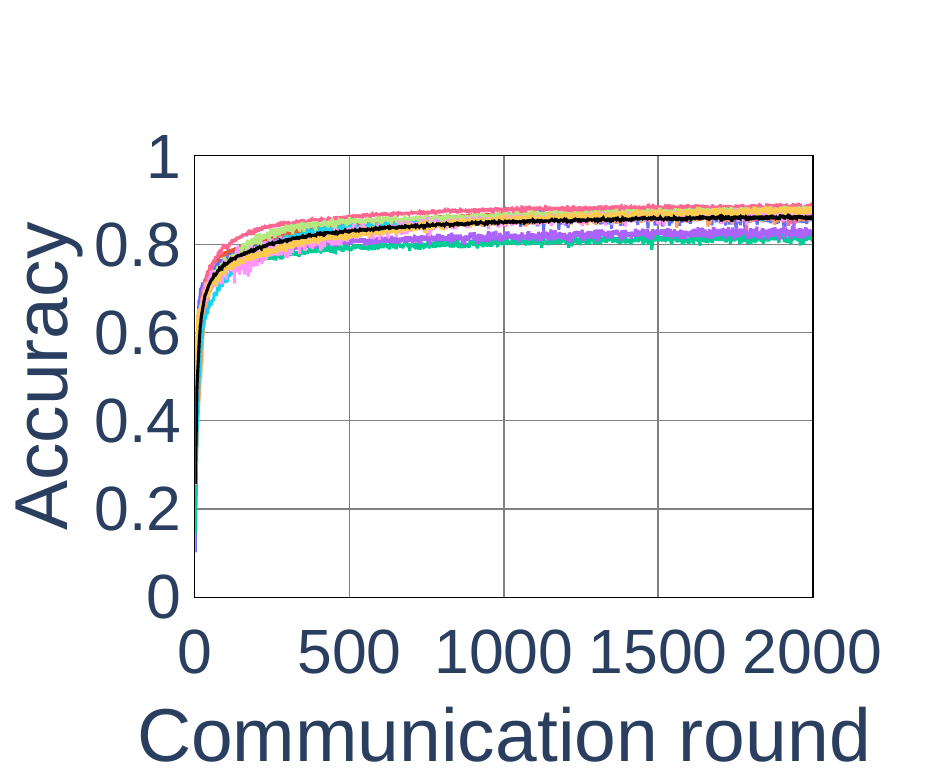}}
\subfloat[\revhlnon{DFedAvgM}]{\includegraphics[width=\twidth,clip,trim=0 0 {\trimright} {\trimtop}]{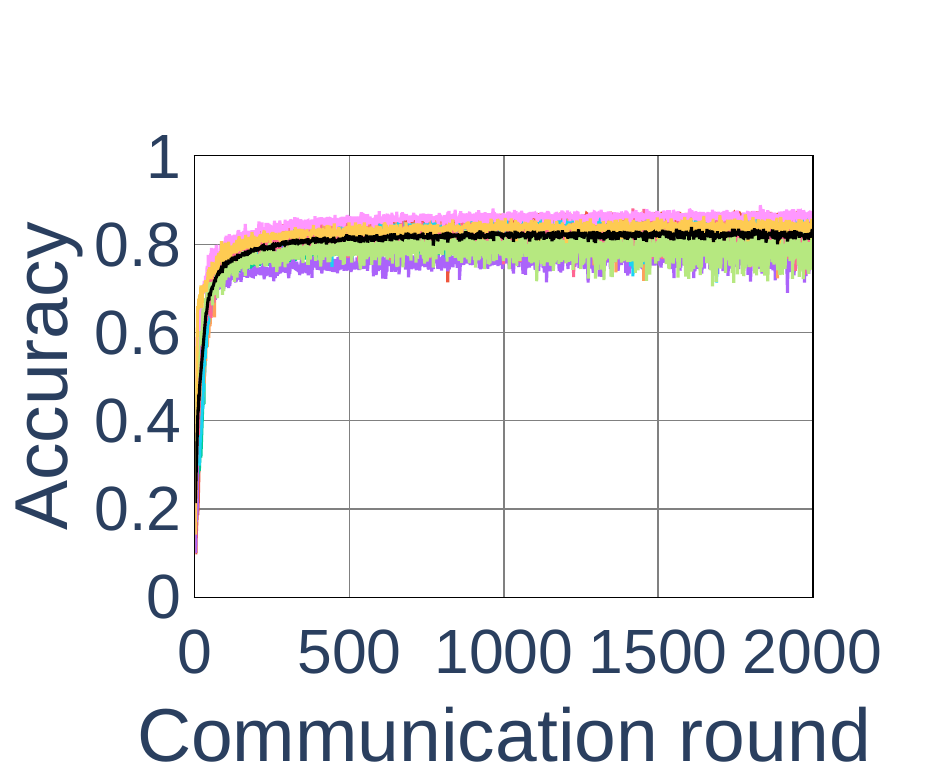}}
\subfloat[\revhlnon{DFedADMM}]{\includegraphics[width=\twidth,clip,trim=0 0 {\trimright} {\trimtop}]{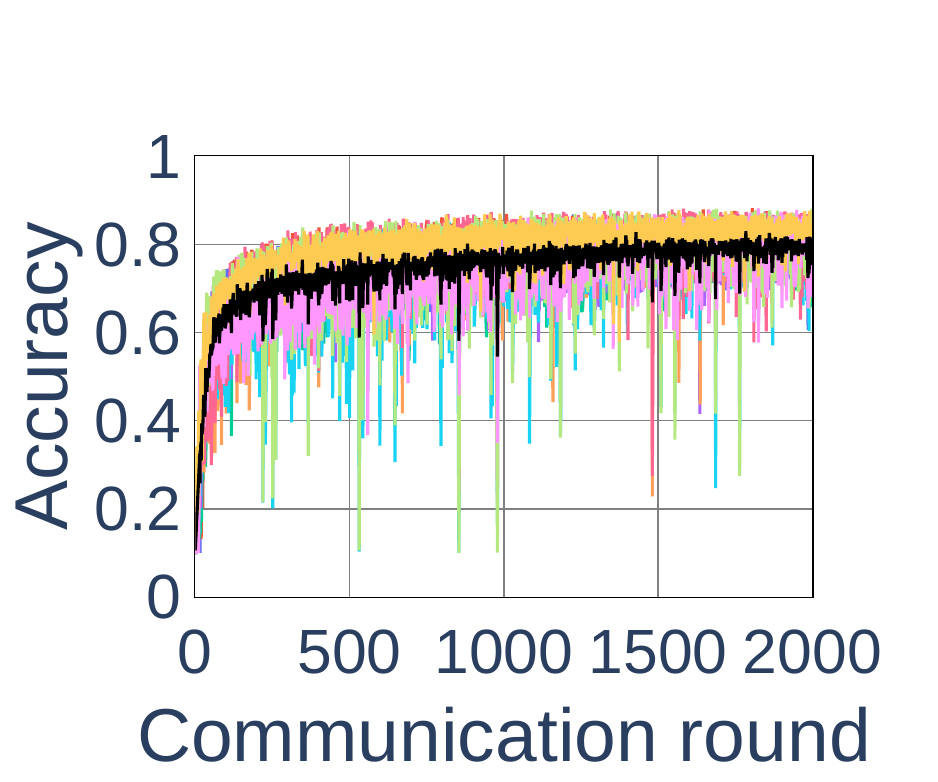}}\\
\subfloat[\gls{CMFD}]{\includegraphics[width=\twidth,clip,trim=0 0 {\trimright} {\trimtop}]{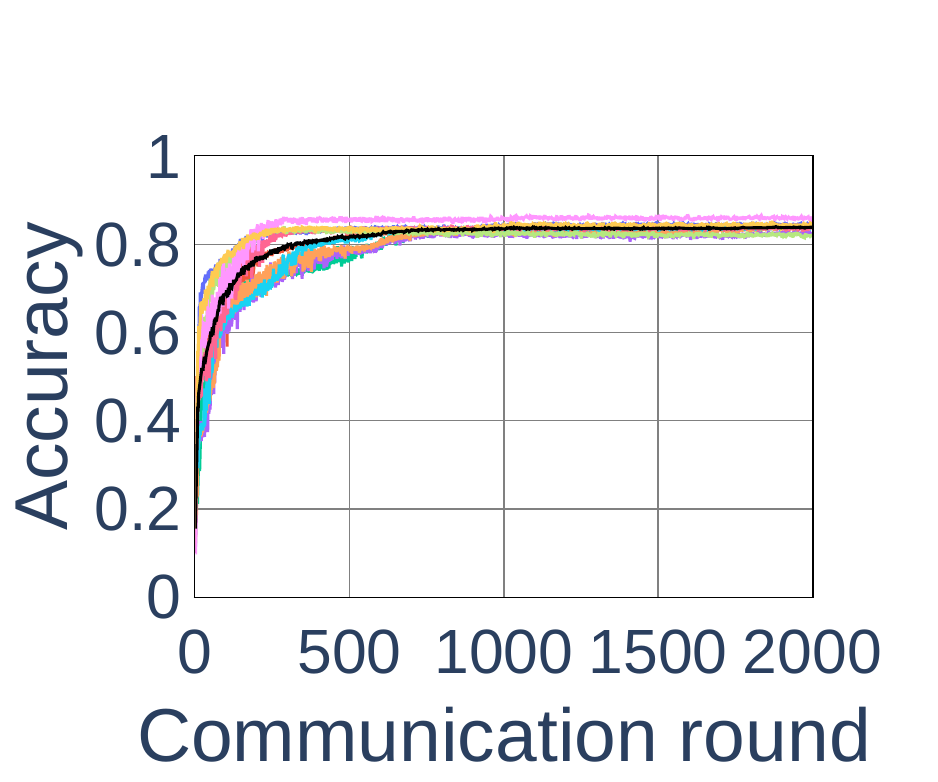}}
\subfloat[\gls{PropAlg} (Proposed)]{\includegraphics[width=\twidth,clip,trim=0 0 {\trimright} {\trimtop}]{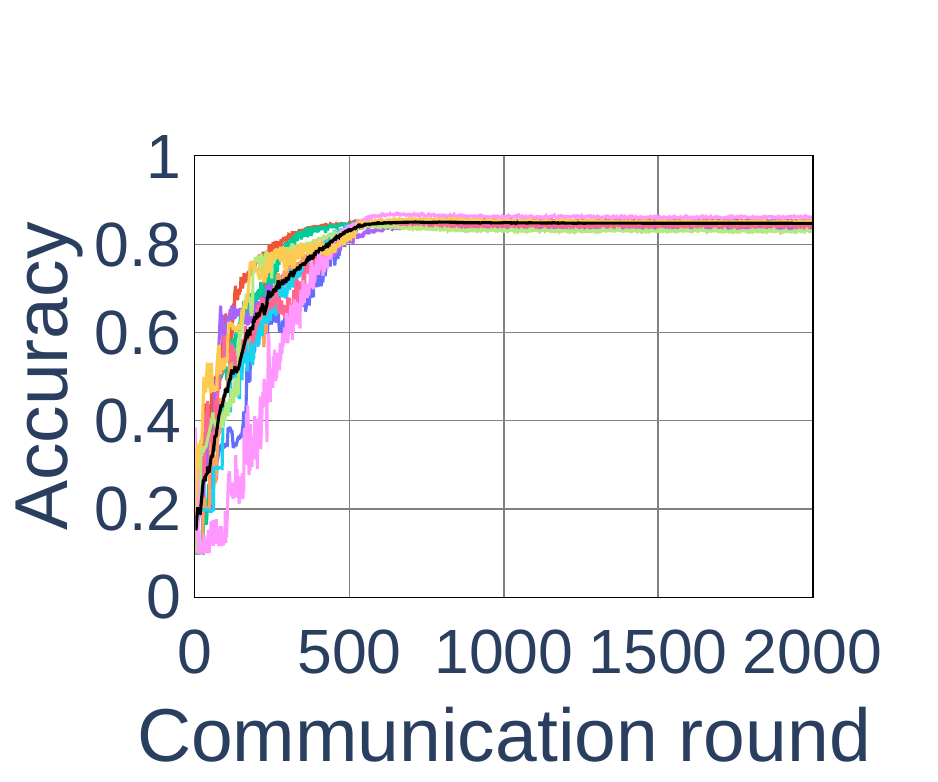}} 
\caption{Convergence performance on Fashion-MNIST with Dirichlet distribution ($\alpha=0.1$). Each device has 1000 unlabeled shared samples, and the total number of training samples is 10000. The black and colored lines represent average and individual device accuracies, respectively. Although DecFedAvg achieves the highest average accuracy, the proposed \gls{PropAlg} attains the smallest accuracy gap among devices.}
\label{fig:conv_fmnist_10000_dir}
\end{figure}

\begin{figure}[!t]
\centering
\subfloat[DecFedAvg]{\includegraphics[width=\twidth,clip,trim=0 0 {\trimright} {\trimtop}]{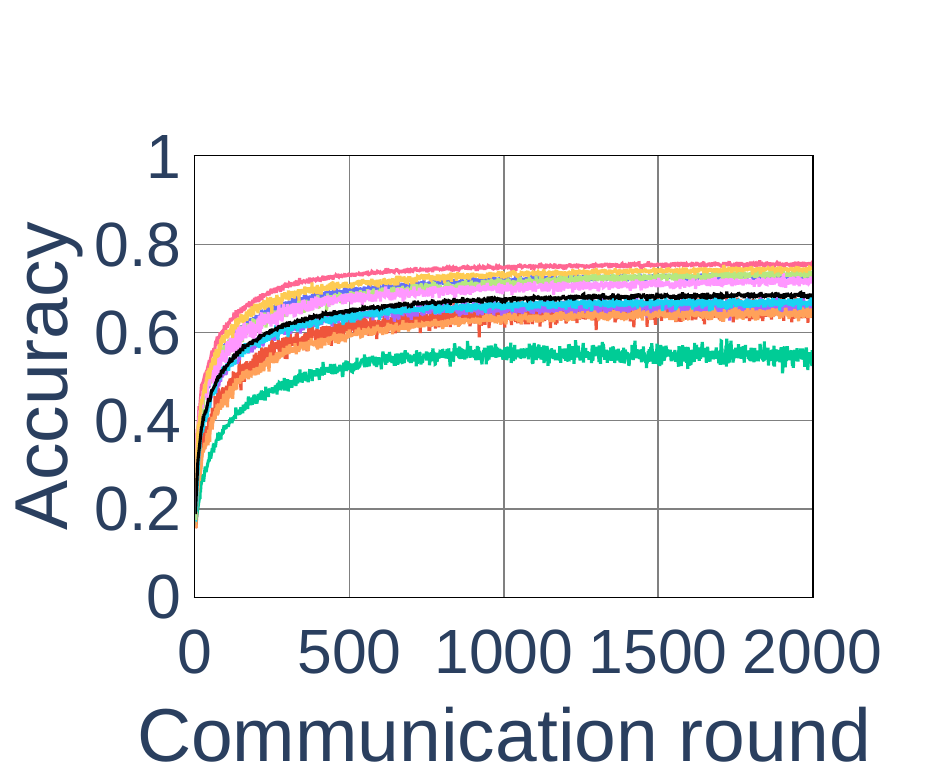}}
\subfloat[\revhlnon{DFedAvgM}]{\includegraphics[width=\twidth,clip,trim=0 0 {\trimright} {\trimtop}]{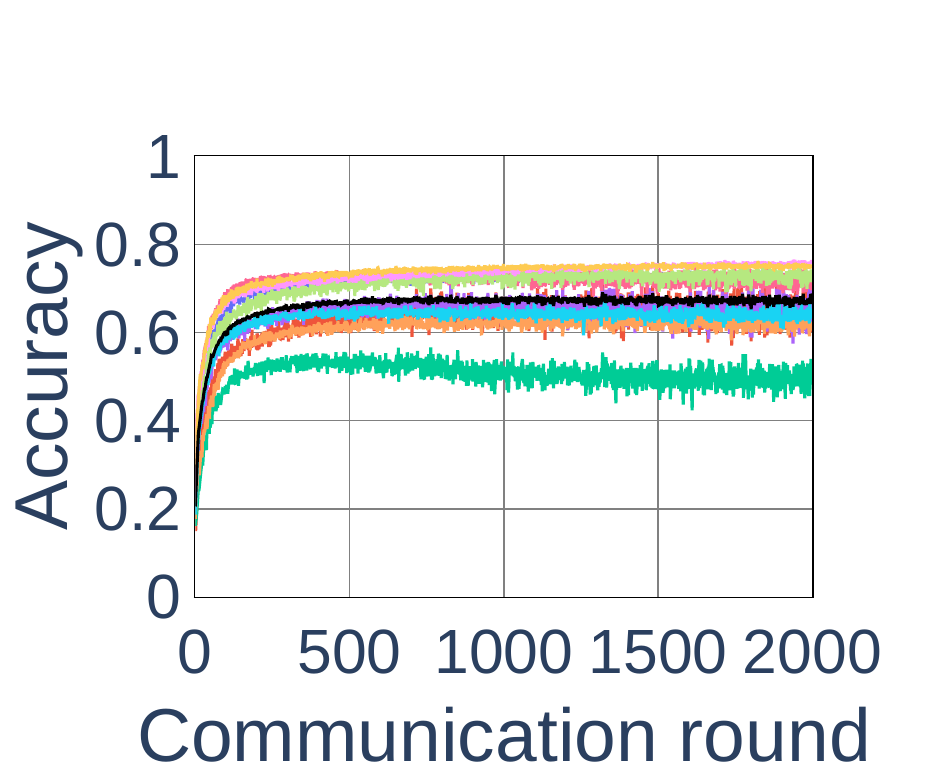}}
\subfloat[\revhlnon{DFedADMM}]{\includegraphics[width=\twidth,clip,trim=0 0 {\trimright} {\trimtop}]{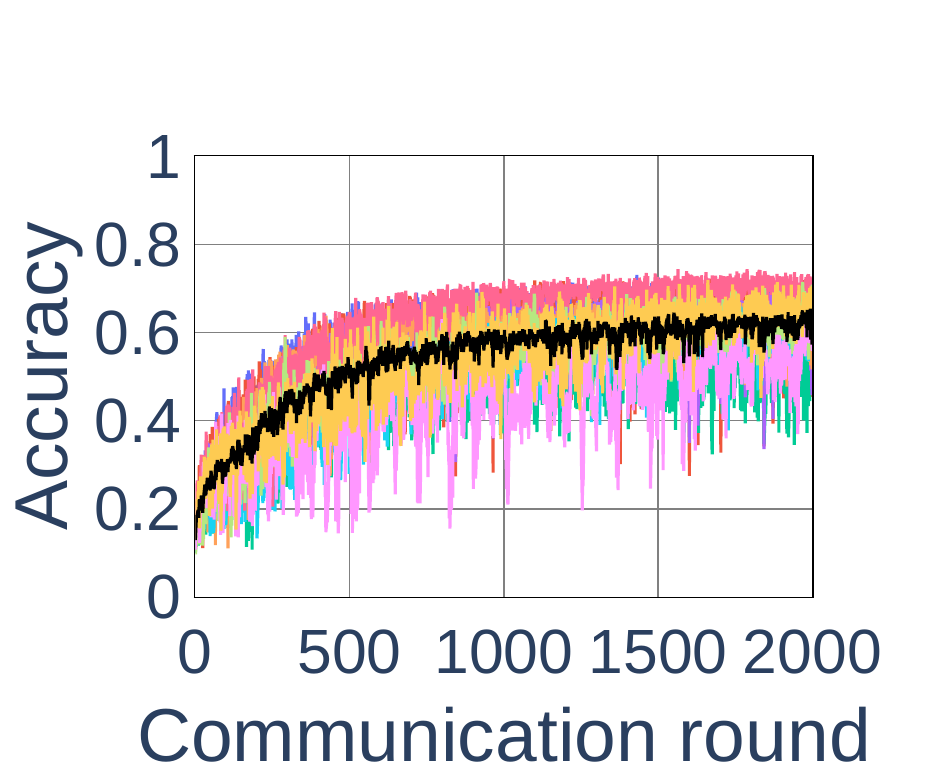}}\\
\subfloat[\gls{CMFD}]{\includegraphics[width=\twidth,clip,trim=0 0 {\trimright} {\trimtop}]{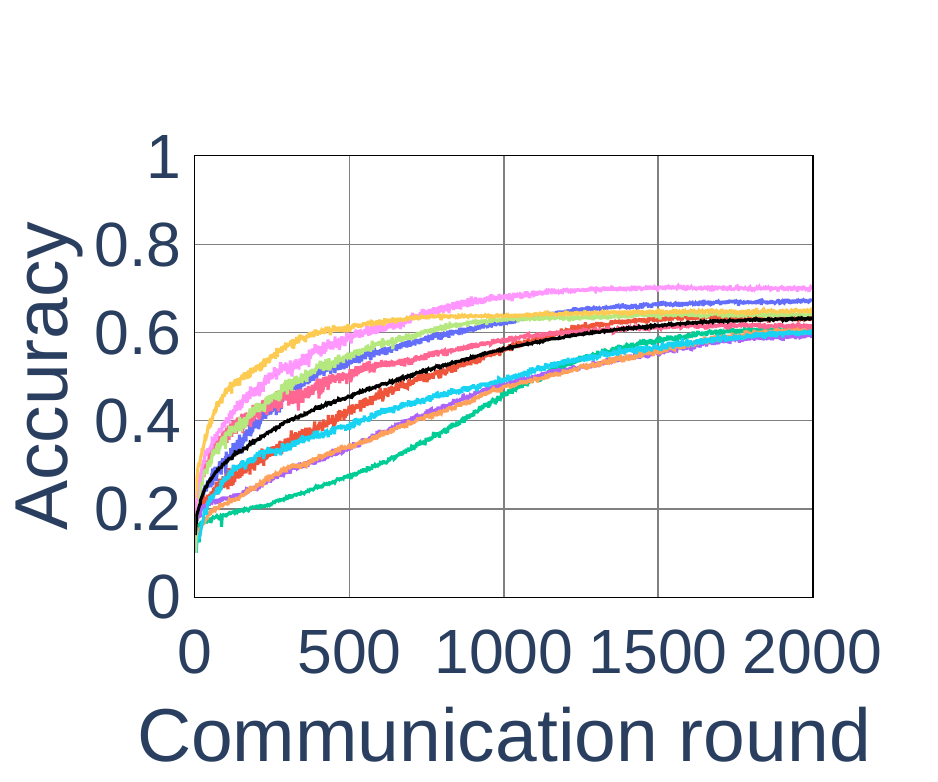}}
\subfloat[\gls{PropAlg} (Proposed)]{\includegraphics[width=\twidth,clip,trim=0 0 {\trimright} {\trimtop}]{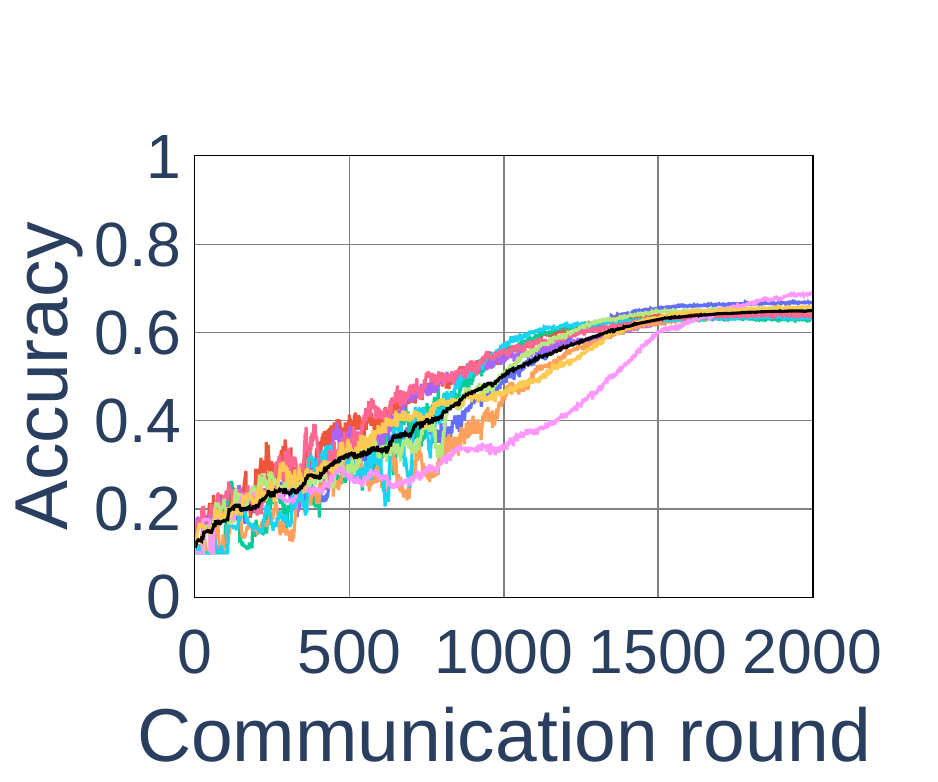}} 
\caption{Convergence performance on CIFAR-10 with Dirichlet distribution ($\alpha=0.1$). Each device has 5000 unlabeled shared samples, and the total number of training samples is 50000. The black and colored lines represent average and individual device accuracies, respectively. Similar to the Fashion-MNIST results, DecFedAvg achieves the highest average accuracy, while the proposed \gls{PropAlg} attains the smallest accuracy gap among devices.}
\label{fig:conv_cifar10_50000_dir}
\end{figure}

\begin{figure}[!t]
\centering
\subfloat[Dirichlet distribution ($\alpha=0.1$). Marker size indicates the number of samples.]{\includegraphics[width=0.3\textwidth]{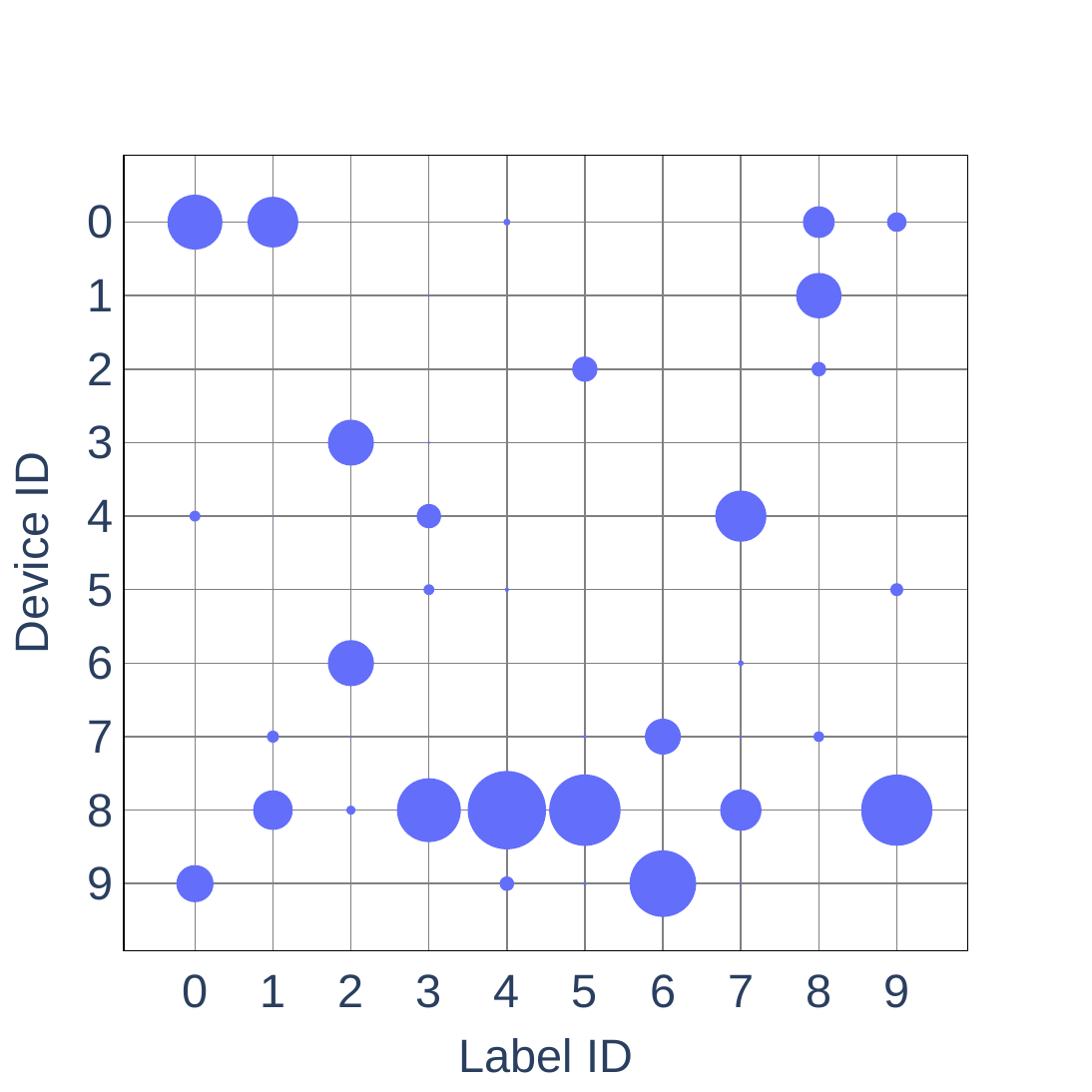}\label{fig:dir_dist_smpl}}
\hfill
\subfloat[KL divergence from the uniform distribution.]{\includegraphics[width=0.17\textwidth]{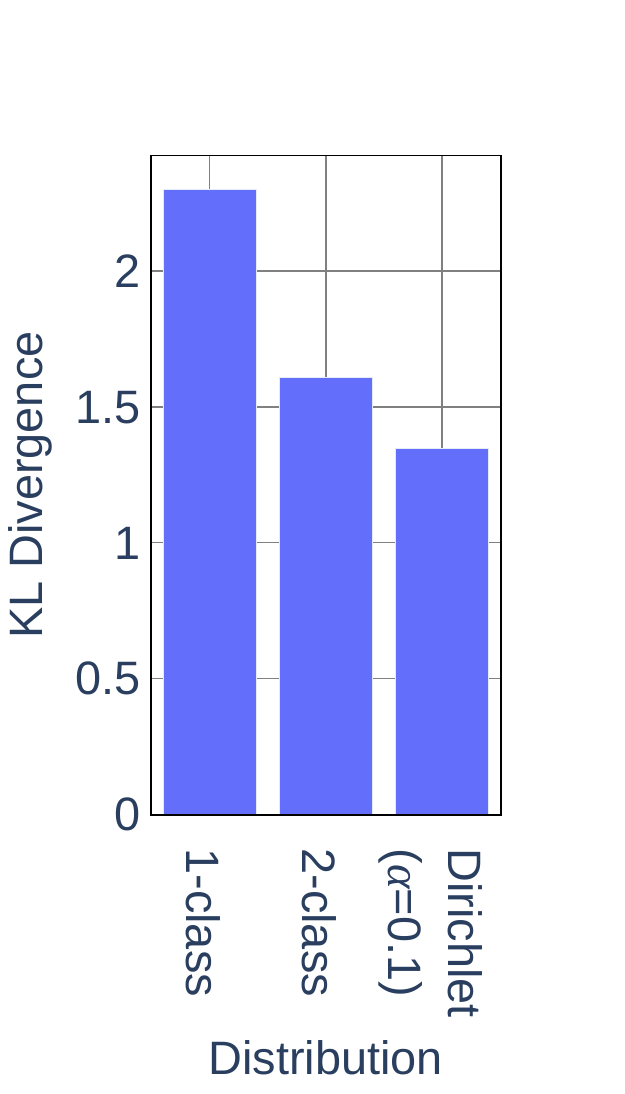}\label{fig:kl_div}}
\caption{Dirichlet distribution and comparison of KL divergence from the uniform distribution. The 1-class distribution has the largest KL divergence, indicating the most severe data heterogeneity.}
\label{fig:dir_dist}
\end{figure}

Figs.~\ref{fig:conv_fmnist_10000_dir} and \ref{fig:conv_cifar10_50000_dir} illustrate the convergence performance when the training samples are distributed according to a Dirichlet distribution with $\alpha=0.1$, using Fashion-MNIST and CIFAR-10, respectively.
The total number of samples used is 10,000 for Fashion-MNIST (Fig.~\ref{fig:conv_fmnist_10000_dir}) and 50,000 for CIFAR-10 (Fig.~\ref{fig:conv_cifar10_50000_dir}).
The same data distribution, illustrated in Fig.~\ref{fig:dir_dist_smpl}, is reproduced for each evaluation.
When the data are distributed according to a Dirichlet distribution, both conventional algorithms achieve comparable average accuracy to \gls{PropAlg}.
This is because the Dirichlet distribution, which is commonly used to model non-\gls{IID} data, is less imbalanced than the 1-class and 2-class distributions as shown in Fig.~\ref{fig:kl_div}; thus, even simple algorithms such as DecFedAvg and \gls{CMFD} can achieve good performance.
However, in terms of the accuracy gap among devices, the proposed \gls{PropAlg} exhibits significantly smaller accuracy gaps, indicating that it can better mitigate the effects of data heterogeneity.
\begin{revhl}{1-4}
Furthermore, individual device accuracies of the parameter-level \gls{ADMM} algorithm, DFedADMM, exhibit severe fluctuations in accuracy, even though the average accuracy is relatively high.
In contrast, the proposed \gls{PropAlg} shows stable convergence across all devices, which is a desirable property for real-world applications.
\end{revhl}
The final accuracy and accuracy gap are also listed in Table~\ref{tbl:final_acc}.

\subsection{Impact of stabilization coefficient}\label{sec:impact_decay}

Fig.~\ref{fig:decay_comparison} compares the learning curves of \gls{PropAlg} with different values of the stabilization coefficient $\decayPrm$ on Fashion-MNIST and CIFAR-10, respectively.
Both experiments are conducted under the 1-class distribution setting.
For Fashion-MNIST (Fig.~\ref{fig:decay_fmnist}), the convergence without $\decayPrm$ is highly sensitive to the choice of learning rates $\lr$ and $\kdCoef$, whereas introducing even a small $\decayPrm$ leads to stable convergence.
For CIFAR-10 (Fig.~\ref{fig:decay_cifar10}), \gls{PropAlg} achieves near-optimal performance when $\decayPrm=0.1$ or $0.01$, while the algorithm without $\decayPrm$ fails to converge.

\def\decaytwidth{0.48\textwidth}
\begin{figure}[!t]
\centering
\subfloat[Fashion-MNIST, 1000 samples from 1 label, 1000 unlabeled shared samples.]{\includegraphics[width=\decaytwidth,trim=0 0 0 1.5cm,clip]{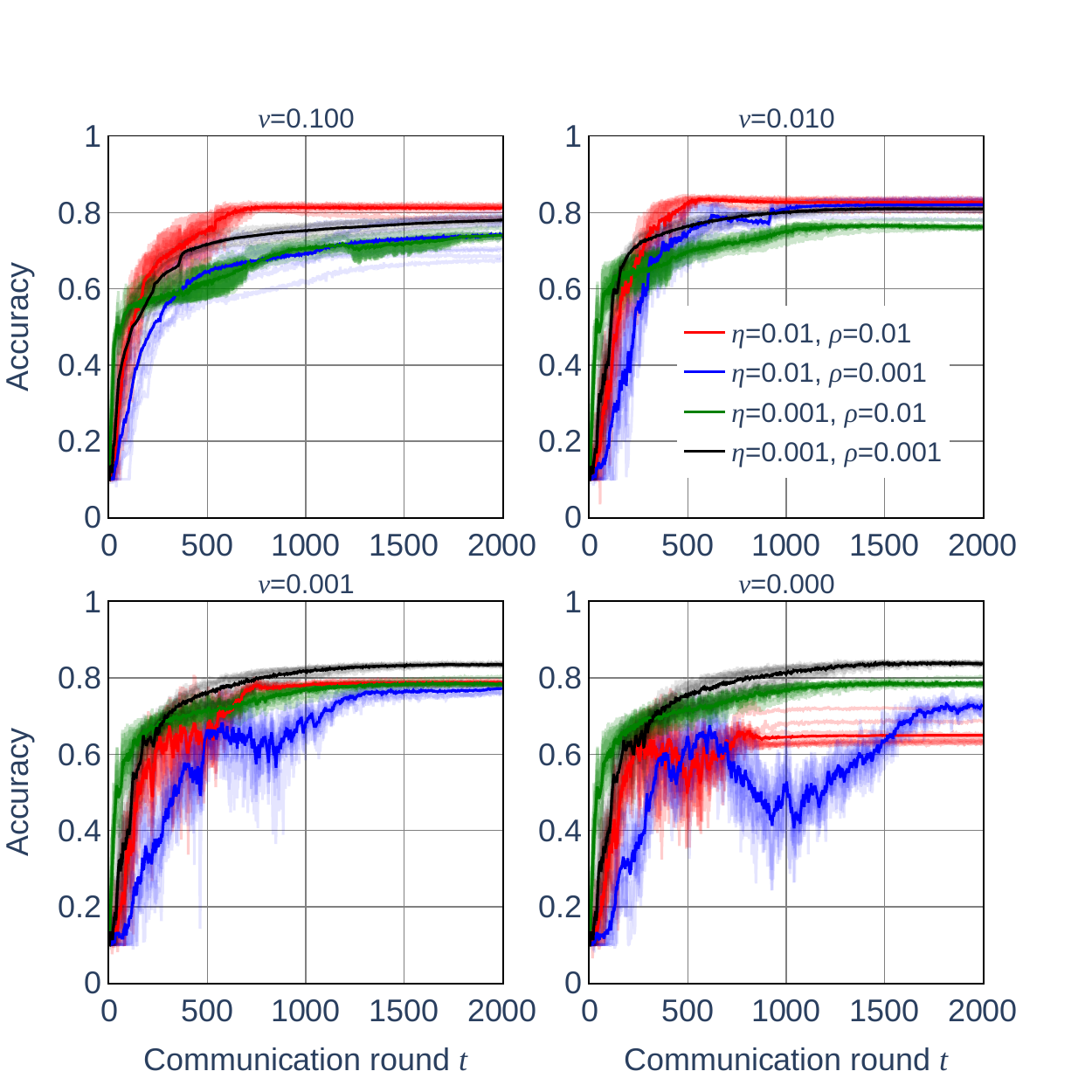}\label{fig:decay_fmnist}}\\
\subfloat[CIFAR-10, 5000 samples from 1 label, 5000 unlabeled shared samples.]{\includegraphics[width=\decaytwidth,trim=0 0 0 1.5cm,clip]{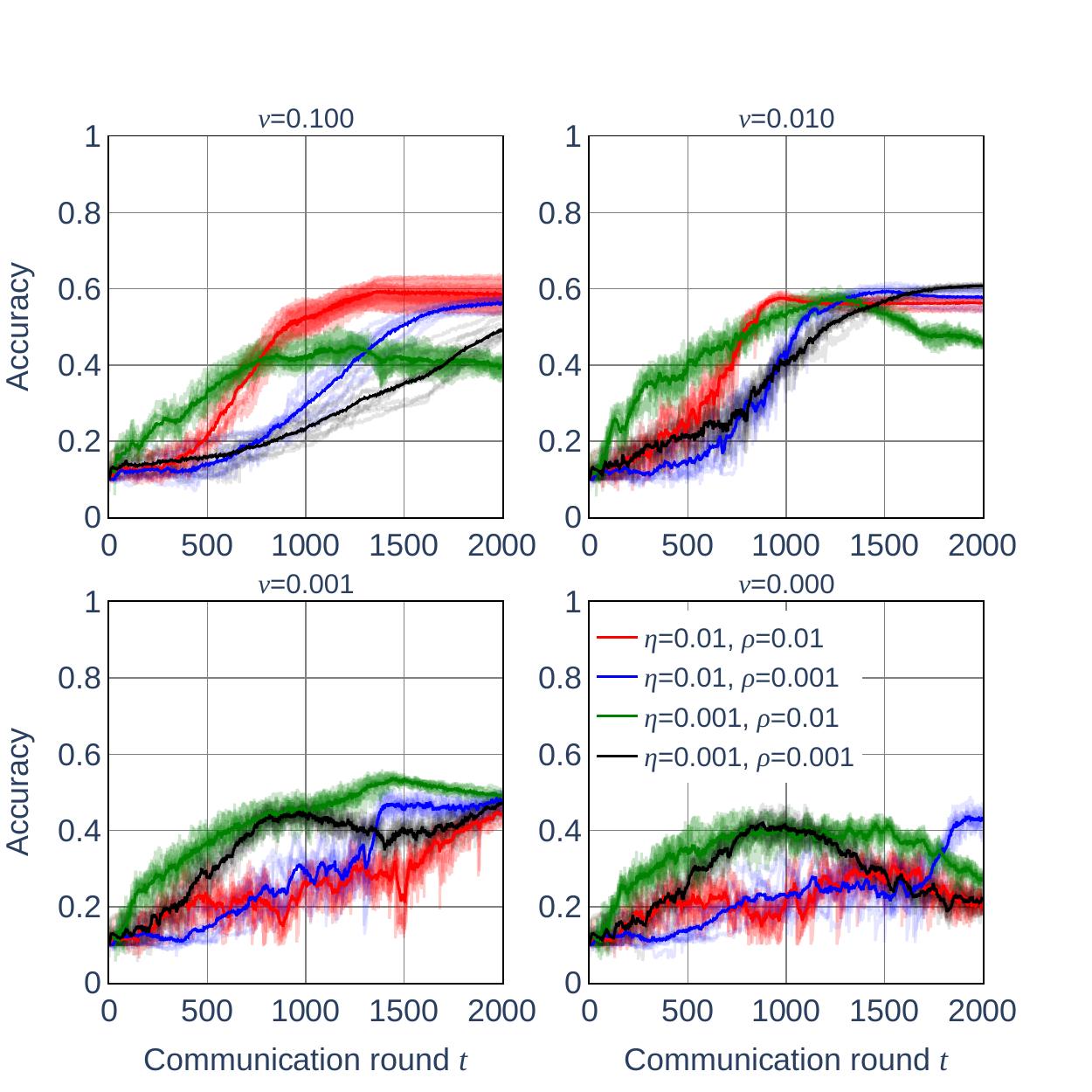}\label{fig:decay_cifar10}}
\caption{Impact of stabilization coefficient $\decayPrm$. Thick lines represent average accuracy across devices, while thin transparent lines represent individual device accuracies. Using a non-zero $\decayPrm$ stabilizes convergence against variations in the learning rate $\lr$ and \gls{KD} coefficient $\kdCoef$.}
\label{fig:decay_comparison}
\end{figure}

\revhlnon{
\subsection{Robustness against shared data distribution}\label{sec:robustness_shared_ds}
}
\begin{revhl}{1-1}
To evaluate the robustness of \gls{PropAlg} against the shared dataset distribution, we compared its convergence performance across three settings: (Std.) the standard \gls{IID} configuration defined in Table~\ref{tbl:sim_param}; (Half) a setting where the number of shared samples was reduced by half; and (Dirichlet) a setting where the same total number of samples as in the Std. setting was distributed according to the Dirichlet distribution shown in Fig.~\ref{fig:distri_shared}.
The Dirichlet parameter was set to $\alpha=10.0$ to ensure all classes were represented in the shared dataset, with the distribution kept consistent across evaluations.
As shown in Fig.~\ref{fig:robustness_shared_ds_results}, reducing the number of shared samples led to a decrease in final accuracy.
However, even under the Dirichlet distribution, \gls{PropAlg} maintained stable convergence and achieved an accuracy level comparable to the \gls{IID} settings.
These results demonstrate that \gls{PropAlg} is robust to variations in the distribution of the shared dataset.

\def\sharedsDirDist{0.30\textwidth}
\def\shareddsablation{0.48\textwidth}
\begin{figure}[!t]
\centering
\subfloat[Non-IID shared datasets]{\includegraphics[width=\sharedsDirDist,trim=0.0cm 0 0.5cm 2.5cm,clip]{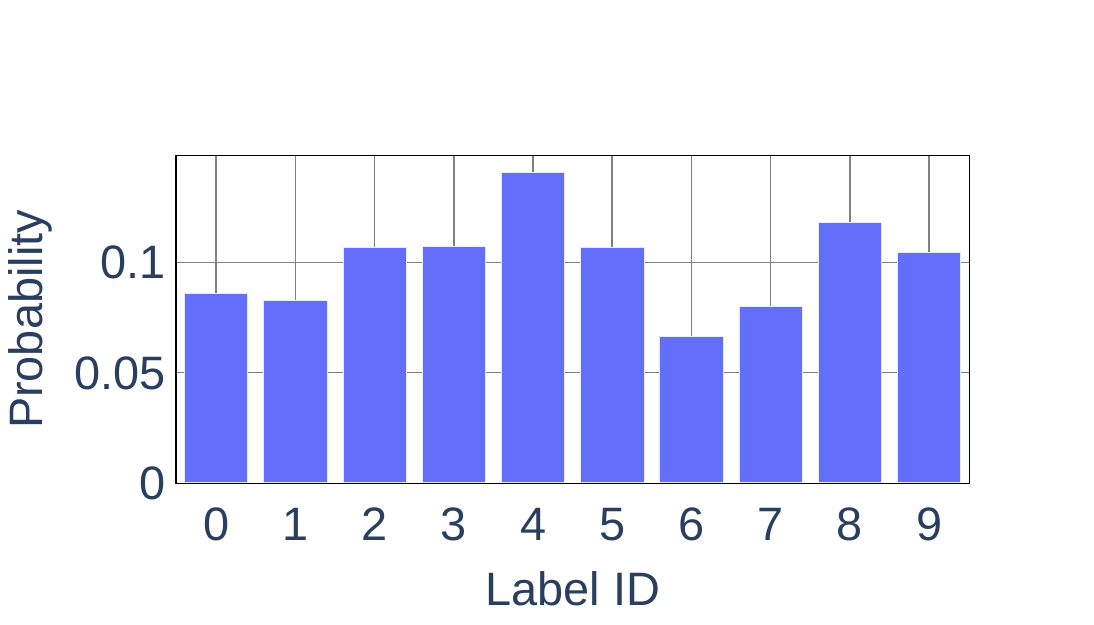}\label{fig:distri_shared}}\\
\subfloat[Convergence performance]{\includegraphics[width=\shareddsablation,trim=0 0 0 2.1cm,clip]{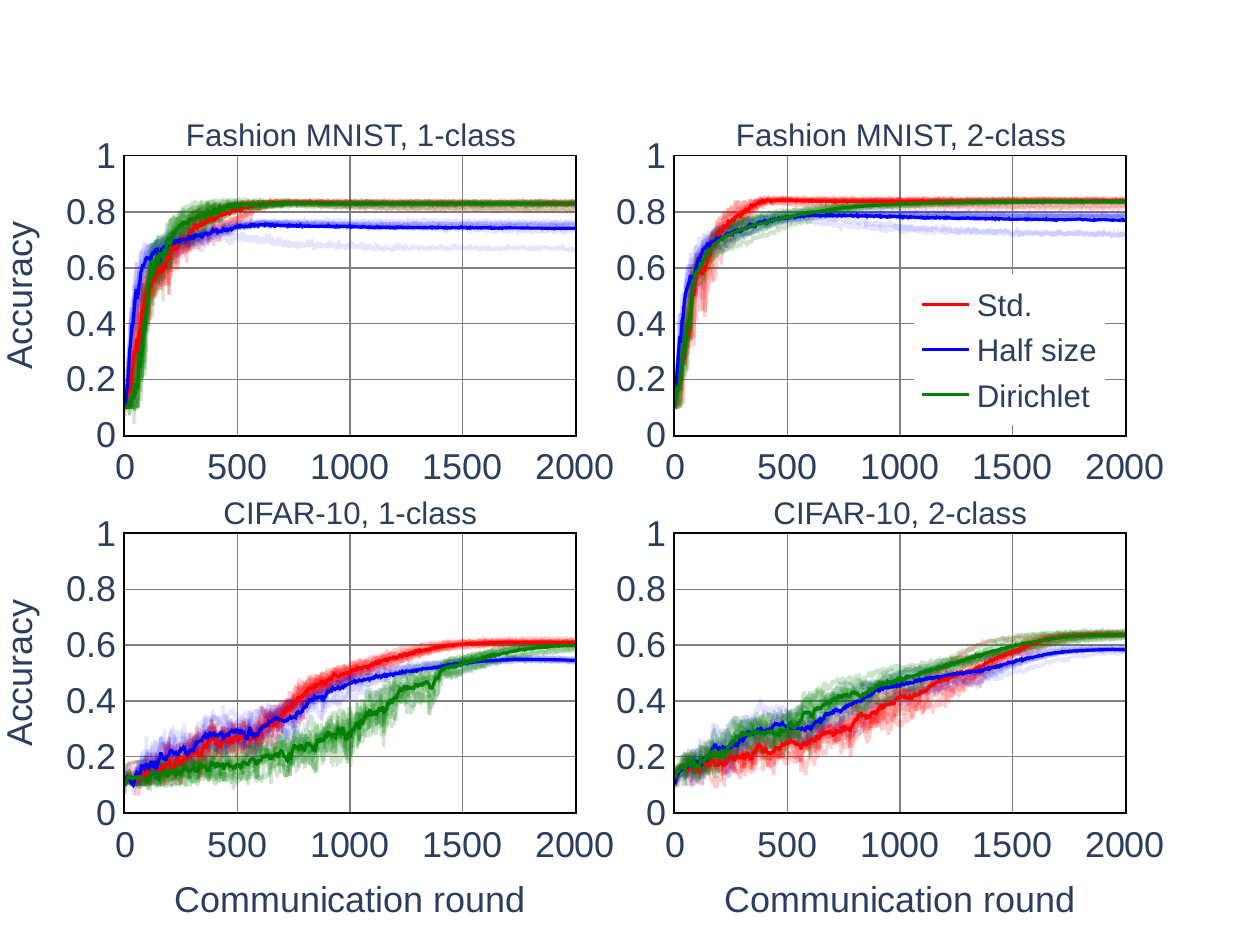}\label{fig:robustness_shared_ds_results}}
\caption{\begin{revhl}{1-1}Comparison of convergence performance with different shared dataset. Thick and thin lines represent average and individual device accuracies, respectively. When the number of shared samples is reduced, the accuracy decreases. However, even with a label-imbalanced shared dataset, \gls{PropAlg} achieves stable convergence and high accuracy.\end{revhl}}
\label{fig:robustness_shared_ds}
\end{figure}
\end{revhl}

\revhlnon{
\subsection{Scalability and robustness against network topologies}\label{sec:robustness_topology}
}
\begin{revhl}{2-8}
Fig.~\ref{fig:robustness_topology} compares the convergence performance of \gls{PropAlg} across different numbers of devices and network topologies: ring, star, and random networks.
The top-left panel reproduces the result in Fig.~\ref{fig:conv_fmnist_1000x1}, where $\NumDev=10$ devices are connected in a ring topology.
The evaluated random network topologies for $\NumDev=10$ and $\NumDev=20$ are illustrated in Fig.~\ref{fig:rnd_nw_topology}.
These random networks were generated such that the number of edges remains consistent with the ring topology. 
Although the star topology tends to suppress fluctuations in individual device accuracies, the convergence speed and final accuracy are comparable across all topologies and network sizes.
This indicates that \gls{PropAlg} is robust to variations in network topology and scales effectively with the number of devices, which is a desirable property for real-world applications.

\begin{figure}[!t]
\centering
\includegraphics[width=0.48\textwidth,trim=0 0 0 2.0cm,clip]{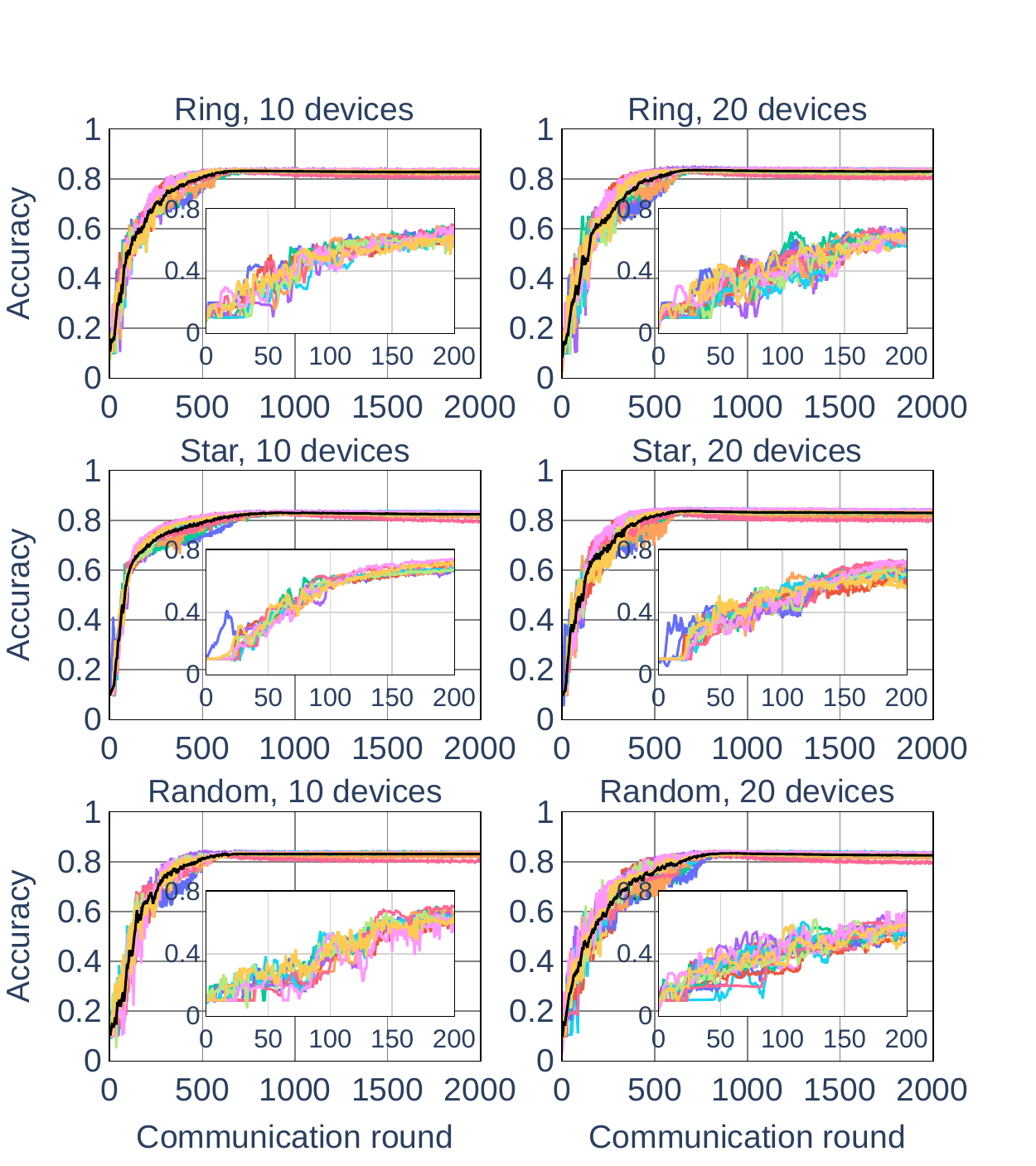}
\caption{\begin{revhl}{2-8}Comparison of convergence performance across different network topologies. Black and colored lines represent average and individual device accuracies, respectively. The convergence speed and final accuracy remain comparable, although the star topology suppresses fluctuations.\end{revhl}}
\label{fig:robustness_topology}
\end{figure}

\begin{figure}[!t]
\centering
\subfloat[$\NumDev=10$ devices]{

\begin{tikzpicture}
  \def\radius{1.3}

  \foreach \i in {0,...,9} {
    \pgfmathsetmacro{\angle}{36*\i}
    \coordinate (pos\i) at ({\radius*cos(\angle)}, {\radius*sin(\angle)});
    \node (node\i) at (pos\i) [draw, circle, minimum size=3, inner sep=0pt, label={[anchor=\angle+180, font=\scriptsize]\angle:\i}] {};
  }

  \foreach \i/\j in {0/5, 0/1, 2/7, 2/3, 3/4, 3/7, 5/9, 5/6, 5/8, 7/9} {
    \draw[thin] (node\i) to (node\j);
  }

\end{tikzpicture}}
\hspace{1em}
\subfloat[$\NumDev=20$ devices]{

\begin{tikzpicture}
  \def\radius{1.3}

  \foreach \i in {0,...,19} {
    \pgfmathsetmacro{\angle}{18*\i}
    \coordinate (pos\i) at ({\radius*cos(\angle)}, {\radius*sin(\angle)});
    \node (node\i) at (pos\i) [draw, circle, minimum size=3, inner sep=0pt, label={[anchor=\angle+180, font=\scriptsize]\angle:\i}] {};
  }

  \foreach \i/\j in {0/7, 0/3, 1/13, 1/17, 2/19, 2/10, 2/15, 3/19, 4/14, 5/10, 5/16, 6/11, 7/18, 7/12, 8/17, 9/15, 9/18, 11/17, 13/18, 14/17} {
    \draw[thin] (node\i) to (node\j);
  }

\end{tikzpicture}}
\caption{\begin{revhl}{2-8}Random network topologies for $\NumDev=10$ and $\NumDev=20$ devices.\end{revhl}}
\label{fig:rnd_nw_topology}
\end{figure}

\end{revhl}

\section{Conclusion} \label{sec:conclusion}
We proposed \gls{PropAlg}, a novel \gls{DFL} algorithm based on \gls{D-ADMM} in function space.
The key idea of \gls{PropAlg} is to optimize prediction models in function space using \gls{ADMM} so as to alleviate issues arising from the non-convexity of parameter-space optimization.
\Gls{KD} is applied to gradually update the local parameters to align the prediction functions across devices, as if they were directly optimized in function space.
The proposed algorithm demonstrates superior performance and stability under non-\gls{IID} data distributions compared to existing methods, especially when each device has data from only one label, which is the most challenging scenario in \gls{DFL}.

Future work includes extending our framework to incorporate a derivative term to realize \gls{PID} control within our framework.
The derivative term is expected to reduce overshoot; thus it can further improve the stability of the training process.
Another direction is to address the limitation of requiring a shared public dataset among devices.
Since the need for a shared dataset may hinder practical applications, it is important to explore methods that can eliminate this requirement while maintaining performance.
A possible approach is to leverage data-free \gls{KD} \cite{zhu2021data} that enables \gls{KD} with augmented data generated from a generative model, collaboratively trained by devices without sharing raw data.

\appendices
\revhlnon{\section{Derivation of the element-wise loss update} \label{sec:derivation_loss_update}}
\begin{revhl}{1-5}
In this appendix, we derive the expression for the element-wise loss update $\Delta \elemloss_i^\step$ in \eqref{eq:loss_update}.
Using the first-order Taylor expansion, $\Delta \elemloss_i^\step$ is approximated as follows:
\begin{align}
  \Delta \elemloss_i^\step
    &\defeq \elemloss\(\updOutval_i^{\step}, \fngt(\inval)\) - \elemloss\(\outval_i^{\step}, \fngt(\inval)\) \nonumber \\
    &= \elemloss\(\fn(\inval;\weight_i^{\step+1}), \fngt(\inval)\) - \elemloss\(\fn(\inval;\weight_i^{\step}), \fngt(\inval)\) \nonumber \\
    &\approx \frac{\partial \elemloss}{\partial \weight_i^{\step}} \cdot \Delta \weight_i^{\step} \nonumber \\
    &= - \lr \frac{\partial \elemloss}{\partial \weight_i^{\step}} \cdot \nabla_{\weight_i} \sum_{\inval^{\prime}\in\DsLocal{i}} \elemloss\(\fn(\inval^{\prime} ;\weight_i^\step), \fngt(\inval^{\prime})\) \nonumber \\
    &= \revhlnon{- \lr \left.\frac{\partial \elemloss}{\partial \weight_i^{\step}}\right|_{\inval} \cdot \sum_{\inval^{\prime}\in\DsLocal{i}} \left.\frac{\partial \elemloss}{\partial \weight_i^\step}\right|_{\inval^\prime}},
\end{align}
where $\Delta \weight_i^{\step}$ denotes the update of the model parameters corresponding to the second term of \eqref{eq:localsgd}.
By decomposing the sum into the term corresponding to $\inval$ and the rest of the terms, we obtain:
\begin{align}
  \Delta \elemloss_i^\step
    &\approx \begin{dcases*}
      - \lr \left\|\frac{\partial \elemloss}{\partial \weight_i^\step}\right\|^2
          - \lr \sum_{\revhlnon{\inval^{\prime}\in\DsLocal{i}{\setminus}\{\!\inval\!\}}}
                  \left.\frac{\partial \elemloss}{\partial \weight_i^\step}\right|_\inval
            \cdot \left.\frac{\partial \elemloss}{\partial \weight_i^\step}\right|_{\inval^\prime} \\
            \hspace{13em} \textrm{if } \inval \in \DsLocal{i} \\
              - \revhlnon{\lr} \sum_{\inval^{\prime}\in\DsLocal{i}}
                  \left.\frac{\partial \elemloss}{\partial \weight_i^\step}\right|_\inval
            \cdot \left.\frac{\partial \elemloss}{\partial \weight_i^\step}\right|_{\inval^\prime} \\
            \hspace{13em} \textrm{otherwise,}
    \end{dcases*}
\end{align}
where $\DsLocal{i}{\setminus}\{\inval\}$ denotes the local dataset excluding $\inval$.
\end{revhl}

\section{Derivation of the update term} \label{sec:derivation}
In this appendix, we derive the expression for the update term $\updateTerm^\step$ in \eqref{eq:dist_update}.
Let us define $\updateTermDiff^{\step}$ as $\updateTermDiff^{\step} \defeq \Dist^{\step+1} - \updDist^{\step}$.
Using \eqref{eq:multiplier_update} and \eqref{eq:dist_update_aggregation}, $\updateTermDiff^{\step}$ is derived as follows:
\begin{align}
  \updateTermDiff^{\step}
  =& - \kdCoef \(\frac{1}{2} \IncidenceT \InvDeg \Incidence \updDist^\step + \IncidenceT \multiplier^\step + \IncidenceT \noiseKd^\step\) \label{eq:def_updateTermDiff} \\
  =& - \frac{\kdCoef}{2} \MatCoef \updDist^{\step}
     - \kdCoef \IncidenceT \(\(1-\decayPrm\) \multiplier^{\step-1} + \frac{1}{2} \InvDeg \Incidence \updDist^\step\) \nonumber \\
   & \mbox{} - \kdCoef \IncidenceT \noiseKd^\step \nonumber \\
  =& - \kdCoef \MatCoef \updDist^{\step}
     - \kdCoef \(1-\decayPrm\) \IncidenceT  \multiplier^{\step-1} - \kdCoef \IncidenceT \noiseKd^\step. \label{eq:updateTermDiff_a}
\end{align}
From \eqref{eq:def_updateTermDiff}, we also obtain the following relationship:
\begin{align}
  - \kdCoef \IncidenceT \multiplier^{\step-1}
  = \updateTermDiff^{\step-1} + \frac{\kdCoef}{2} \MatCoef \updDist^{\step-1} + \kdCoef \IncidenceT \noiseKd^{\step-1}.
\end{align}
Substituting this equation into \eqref{eq:updateTermDiff_a} yields
\begin{align}
  \updateTermDiff^{\step}
   =& - \kdCoef \MatCoef \updDist^{\step} \nonumber \\
    &\mbox{} + (1-\decayPrm) \(\updateTermDiff^{\step-1} + \frac{\kdCoef}{2} \MatCoef \updDist^{\step-1} + \kdCoef \IncidenceT \noiseKd^{\step-1}\) \nonumber \\
    &\mbox{} - \kdCoef \IncidenceT \noiseKd^\step \nonumber \\
   =&   \updateTermDiff^{\step-1} - \decayPrm \updateTermDiff^{\step-1}
      + \kdCoef \MatCoef \(-\updDist^{\step} + \frac{1-\decayPrm}{2}\updDist^{\step-1}\) \nonumber \\
    &\mbox{} + \kdCoef \IncidenceT \(-\noiseKd^\step + (1-\decayPrm) \noiseKd^{\step-1}\). \label{eq:updateTermDiff_b}
\end{align}
By substituting $\updateTermDiff^{\step-1}=\Dist^{\step}-\updDist^{\step-1}$ into the above equation, we have:
\begin{align}
  \updateTermDiff^{\step}
   =& -\decayPrm \Dist^{\step} - \kdCoef \MatCoef \updDist^{\step} - \kdCoef \IncidenceT \noiseKd^\step \nonumber \\
    &\mbox{} + \updateTermDiff^{\step-1} + \decayPrm \updDist^{\step-1} + \frac{\kdCoef(1-\decayPrm)}{2} \MatCoef \updDist^{\step-1} \nonumber \\
    &\mbox{} + \kdCoef(1-\decayPrm) \IncidenceT \noiseKd^{\step-1} \nonumber \\
   =& -\decayPrm \Dist^{\step} - \kdCoef \MatCoef \updDist^{\step} - \kdCoef \IncidenceT \noiseKd^\step + \tmpVariable^{\step-1}, \label{eq:updateTermDiff_c}
\end{align}
where $\tmpVariable^{\step}$ is defined as
\begin{align}
  \tmpVariable^{\step} \defeq \updateTermDiff^{\step} + \decayPrm \updDist^{\step} + \frac{\kdCoef(1-\decayPrm)}{2} \MatCoef \updDist^{\step} + \kdCoef(1-\decayPrm) \IncidenceT \noiseKd^{\step}. \label{eq:def_tmpVariable}
\end{align}
From \eqref{eq:updateTermDiff_c} and \eqref{eq:def_tmpVariable}, we derive the update rule for $\tmpVariable^{\step}$:
\begin{align}
  \tmpVariable^{\step} - \tmpVariable^{\step-1}
    =& \decayPrm \(\updDist^{\step} - \Dist^{\step}\) - \frac{\kdCoef (1+\decayPrm)}{2} \MatCoef \updDist^{\step} - \kdCoef \decayPrm \IncidenceT \noiseKd^{\step} \nonumber \\
    =& \decayPrm \IncidenceT \StackedDiffY^{\step} - \frac{\kdCoef (1+\decayPrm)}{2} \MatCoef \updDist^{\step} - \kdCoef \decayPrm \IncidenceT \noiseKd^{\step}. \label{eq:tmpVariable_update}
\end{align}
Summing the above equation from $\tmpStep=1$ to $\tmpStep=\step-1$, we obtain
\begin{align}
  \tmpVariable^{\step-1}
          =&   \tmpVariable^{0}
             + \decayPrm \sum_{\tmpStep=1}^{\step-1} \IncidenceT \StackedDiffY^{\tmpStep}
             - \frac{\kdCoef (1+\decayPrm)}{2} \sum_{\tmpStep=1}^{\step-1} \MatCoef \updDist^{\tmpStep} \nonumber \\
    &\mbox{} - \kdCoef \decayPrm \IncidenceT \sum_{\tmpStep=1}^{\step-1} \noiseKd^{\tmpStep}. \label{eq:tmpVariable_expanded}
\end{align}
Comparing \eqref{eq:updateTermDiff_a} and \eqref{eq:updateTermDiff_c}, the initial value $\tmpVariable^{0}$ is given by
\begin{align}
  \tmpVariable^{0} = \decayPrm \Dist^1 - \kdCoef (1-\decayPrm) \IncidenceT \multiplier^0.
\end{align}
Assuming that the initial model parameters are synchronized among all devices and the multiplier functions are initialized to zero, we have $\tmpVariable^0=\ZeroVec$.

Finally, the update term $\updateTerm^{\step}$ is expressed as follows:
\begin{align}
  \updateTerm^{\step}
    \underset{(a)}{=}&   \updateTermDiff^{\step} + \updDist^{\step} - \Dist^{\step} \nonumber \\ 
    \underset{(b)}{=}& - \decayPrm \Dist^{\step} - \kdCoef \MatCoef \updDist^{\step} - \kdCoef \IncidenceT \noiseKd^\step
                       + \tmpVariable^{\step-1} + \IncidenceT \StackedDiffY^{\step} \nonumber \\
    \underset{(c)}{=}&   \IncidenceT \(\StackedExpDiffY^{\step} + \noiseSgd^{\step}\)
                       - \decayPrm \Dist^{\step} - \kdCoef \MatCoef \updDist^{\step} - \kdCoef \IncidenceT \noiseKd^\step \nonumber \\
              &\mbox{} + \decayPrm \sum_{\tmpStep=1}^{\step-1} \IncidenceT \(\StackedExpDiffY^{\tmpStep} + \noiseSgd^{\tmpStep}\)
                       - \frac{\kdCoef (1+\decayPrm)}{2} \sum_{\tmpStep=1}^{\step-1} \MatCoef \updDist^{\tmpStep} \nonumber \\
              &\mbox{} - \kdCoef \decayPrm \IncidenceT \sum_{\tmpStep=1}^{\step-1} \noiseKd^{\tmpStep} \nonumber \\
                    =&   \IncidenceT \StackedExpDiffY^{\step} + \decayPrm \(\ZeroVec - \Dist^{\step}\) + \kdCoef \(\ZeroVec - \MatCoef \updDist^{\step}\) \nonumber \\
              &\mbox{} + \decayPrm \sum_{\tmpStep=1}^{\step-1} \IncidenceT \StackedExpDiffY^{\tmpStep}
                       + \frac{\kdCoef (1+\decayPrm)}{2} \sum_{\tmpStep=1}^{\step-1} \(\ZeroVec - \MatCoef \updDist^{\tmpStep}\) \nonumber \\
              &\mbox{} + \IncidenceT \(\noiseSgd^{\step} - \kdCoef \noiseKd^{\step} + \decayPrm \sum_{\tmpStep=1}^{\step-1} \(\noiseSgd^{\tmpStep} - \kdCoef \noiseKd^{\tmpStep}\) \), \label{eq:final_updateTerm}
\end{align}
where (a) follows from the definitions of $\updateTerm^\step$ and $\updateTermDiff^\step$, (b) from \eqref{eq:updateTermDiff_c}, and (c) from \eqref{eq:def_diff_y} and \eqref{eq:tmpVariable_expanded}.

\section{Reason for indirect consensus} \label{sec:reason_indirect_consensus}
In this appendix, we explain why the third and fifth terms of \eqref{eq:dist_update} contribute less effectively to achieving consensus among devices compared to the direct consensus terms.
The term $\MatCoef \updDist^{\step}$ can be expanded as:
\begin{align}
\MatCoef \updDist^{\step}
  = \IncidenceT \InvDeg \Incidence \updDist^{\step}
  = 2 \IncidenceT \InvDeg \Laplacian \StackedUpdOutVal^{\step}.
\end{align}
\begin{revhl}{2-11}%
Since the incidence matrix calculates the difference between the values of the incident nodes for each edge, the element of $\MatCoef \updDist^{\step}$ corresponding to the edge $(s,e)$ is given by
\begin{align}
[\MatCoef \updDist^{\step}]_{(s,e)}
  =& 2 \(\[\InvDeg \Laplacian \StackedUpdOutVal^{\step}\]_{s} - \[\InvDeg \Laplacian \StackedUpdOutVal^{\step}\]_{e}\),
\end{align}
where $[\bm{x}]_{i}$ denotes the $i$-th element of vector $\bm{x}$.
Recalling that the Laplacian matrix $\Laplacian$ is equal to the difference between the degree matrix $\DegMat$ and the adjacency matrix $\AdjMat$, we have
\begin{align}
\InvDeg \Laplacian \StackedUpdOutVal
  =& \InvDeg (\DegMat - \AdjMat) \StackedUpdOutVal \nonumber \\
  =& \StackedUpdOutVal - \InvDeg \AdjMat \StackedUpdOutVal.
\end{align}
Since applying $\InvDeg\AdjMat$ to a vector results in a vector whose elements represent the means of neighbor values, we finally obtain the element of $\MatCoef \updDist^{\step}$ as
\begin{align}
\[\MatCoef \updDist^{\step}\]_{(s,e)}
  =& 2 \(
       \(\updOutval^{\step}_s - \frac{1}{\NumNeigh{s}}\sum_{i\in\Neigh{s}} \updOutval^{\step}_i\) \right. \nonumber \\
     &- \left. \(\updOutval^{\step}_e - \frac{1}{\NumNeigh{e}}\sum_{j\in\Neigh{e}} \updOutval^{\step}_j\) \label{eq:edgewise_diff}
     \).
\end{align}
\end{revhl}
As shown in the above equation, each element of $\MatCoef \updDist^{\step}$ can be zero even if the output values $\updOutval^{\step}_s$ and $\updOutval^{\step}_e$ are not equal (e.g., if each node value equals the average of its neighbors).
Therefore, the term driving $\MatCoef \updDist^{\step}$ to zero enforces consensus among devices more weakly than the term driving $\Dist^{\step}$ to zero directly.

\ifCLASSOPTIONcompsoc
  \section*{Acknowledgments}
\else
  \section*{Acknowledgment}
\fi

This work was supported by JSPS KAKENHI Grant Number JP24K20759.

\ifCLASSOPTIONcaptionsoff
  \newpage
\fi


\bibliographystyle{IEEEtran}
\bibliography{IEEEabrv,myabrv,main}

@STRING{IEEE_J_VT         = "{IEEE} Trans. Veh. Technol."}

@STRING{IEEE_J_SP         = "{IEEE} Trans. Signal Process."}

@STRING{IEEE_J_IFS        = "{IEEE} Trans. Inf. Forensics Security"}

@STRING{IEEE_J_MC         = "{IEEE} Trans. Mobile Comput."}

@STRING{IEEE_J_PAMI       = "{IEEE} Trans. Pattern Anal. Mach. Intell."}

@STRING{IEEE_O_CSTO        = "{IEEE} Commun. Surveys Tuts."}

@STRING{IEEE_J_VT         = "{IEEE} Transactions on Vehicular Technology"}

@STRING{IEEE_J_SP         = "{IEEE} Transactions on Signal Processing"}

@STRING{IEEE_J_IFS        = "{IEEE} Transactions on Information Forensics and Security"}

@STRING{IEEE_J_MC         = "{IEEE} Transactions on Mobile Computing"}

@STRING{IEEE_J_PAMI       = "{IEEE} Transactions on Pattern Analysis and Machine Intelligence"}

@STRING{IEEE_O_CSTO        = "{IEEE} Communications Surveys and Tutorials"}

@article{savazzi2020federated,
  title={Federated learning with cooperating devices: A consensus approach for massive {IoT} networks},
  author={Savazzi, Stefano and Nicoli, Monica and Rampa, Vittorio},
  journal=IEEE_J_IOT,
  volume={7},
  number={5},
  pages={4641--4654},
  month=may,
  year={2020},
  publisher={IEEE}
}

@article{anil2018large,
  author={Anil, Rohan and Pereyra, Gabriel and Passos, Alexandre and Ormandi, Robert and Dahl, George E and Hinton, Geoffrey E},
  title={Large scale distributed neural network training through online distillation},
  journal={arXiv preprint arXiv:1804.03235},
  month=apr,
  year={2018}
}

@article{jeong2018communication,
  title={Communication-efficient on-device machine learning: Federated distillation and augmentation under non-{IID} private data},
  author={Jeong, Eunjeong and Oh, Seungeun and Kim, Hyesung and Park, Jihong and Bennis, Mehdi and Kim, Seong-Lyun},
  journal={arXiv preprint arXiv:1811.11479},
  month=nov,
  year={2018}
}

@InProceedings{mcmahan2016communication,
  title={Communication-Efficient Learning of Deep Networks from Decentralized Data},
  author={Brendan McMahan and Eider Moore and Daniel Ramage and Seth Hampson and Blaise Aguera y Arcas},
  booktitle={Proc.\ 20th Int. Conf. Artificial Intelligence and Statistics (AISTATS)},
  address={Fort Lauderdale, FL, USA},
  pages={1273--1282},
  year={2017},
  month=apr,
}

@article{kairouz2019advances,
  title={Advances and Open Problems in Federated Learning},
  author={Kairouz, Peter and others},
  journal={Foundations and Trends{\textregistered} in Machine Learning},
  volume={14},
  number={1--2},
  pages={1--210},
  year={2021},
  publisher={Now Publishers, Inc.},
  month=jun,
}

@article{lalitha2019peer,
  title={Peer-to-peer federated learning on graphs},
  author={Lalitha, Anusha and Kilinc, Osman Cihan and Javidi, Tara and Koushanfar, Farinaz},
  journal={arXiv preprint arXiv:1901.11173},
  month=jan,
  year={2019}
}

@inproceedings{lian2017can,
  title={Can decentralized algorithms outperform centralized algorithms? a case study for decentralized parallel stochastic gradient descent},
  author={Lian, Xiangru and Zhang, Ce and Zhang, Huan and Hsieh, Cho-Jui and Zhang, Wei and Liu, Ji},
  booktitle={Proc.\ 31st Conf. Neural Information Processing Systems (NeurIPS)},
  address={Long Beach, CA, USA},
  pages={5330--5340},
  month=dec,
  year={2017}
}

@article{itahara2021distillation,
  title={Distillation-Based Semi-Supervised Federated Learning for Communication-Efficient Collaborative Training with Non-{IID} Private Data},
  author={Itahara, Sohei and Nishio, Takayuki and Koda, Yusuke and Morikura, Masahiro and Yamamoto, Koji},
  journal=IEEE_J_MC,
  pages={191--205},
  volume = {22},
  number = {1},
  month=mar,
  year={2021},
  publisher={IEEE Computer Society}
}

@inproceedings{zhang2018deep,
  title={Deep Mutual Learning},
  author={Zhang, Ying and Xiang, Tao and Hospedales, Timothy M and Lu, Huchuan},
  booktitle={Proc.\ 2018 IEEE/CVF Conf. Computer Vision and Pattern Recognition (CVPR)},
  pages={4320--4328},
  year={2018},
  organization={IEEE}
}

@inproceedings{sato2020network,
  title={Network-Density-Controlled Decentralized Parallel Stochastic Gradient Descent in Wireless Systems},
  author={Sato, Koya and Satoh, Yasuyuki and Sugimura, Daisuke},
  booktitle={Proc.\ IEEE Int. Conf. Commun. (ICC)},
  address={Virtual Conference},
  month=jun,
  year={2020},
}

@article{chang2019cronus,
  title={Cronus: Robust and heterogeneous collaborative learning with black-box knowledge transfer},
  author={Chang, Hongyan and Shejwalkar, Virat and Shokri, Reza and Houmansadr, Amir},
  journal={arXiv preprint arXiv:1912.11279},
  month=dec,
  year={2019}
}

@article{jeong2019multi,
  title={Multi-hop federated private data augmentation with sample compression},
  author={Jeong, Eunjeong and Oh, Seungeun and Park, Jihong and Kim, Hyesung and Bennis, Mehdi and Kim, Seong-Lyun},
  journal={arXiv preprint arXiv:1907.06426},
  month=jul,
  year={2019}
}

@article{li2019fedmd,
  title={{FedMD}: Heterogenous federated learning via model distillation},
  author={Li, Daliang and Wang, Junpu},
  journal={arXiv preprint arXiv:1910.03581},
  month=oct,
  year={2019}
}

@inproceedings{lin2020ensemble,
  title={Ensemble distillation for robust model fusion in federated learning},
  author={Lin, Tao and Kong, Lingjing and Stich, Sebastian U and Jaggi, Martin},
  booktitle={Proc.\ 33rd Conf. Neural Information Processing Systems (NeurIPS)},
  address={Virtual Conference},
  volume={33},
  pages={2351--2363},
  month=dec,
  year={2020}
}

@inproceedings{niwa2020edge,
  title     = {Edge-consensus Learning: Deep Learning on {P2P} Networks with Nonhomogeneous Data},
  booktitle = {Proc.\ 26th {ACM} {SIGKDD} Int. Conf.  Knowledge Discovery \& Data Mining},
  author={Niwa, Kenta and Harada, Noboru and Zhang, Guoqiang and Kleijn, W Bastiaan},
  pages={668--678},
  month=aug,
  address={Virtual Conference},
  year={2020}
}

@article{taya2022decentralized,
  title={Decentralized and model-free federated learning: Consensus-based distillation in function space},
  author={Taya, Akihito and Nishio, Takayuki and Morikura, Masahiro and Yamamoto, Koji},
  journal=IEEE_J_SIPN,
  volume={8},
  month=sep,
  pages={799--814},
  year={2022}
}

@article{beltran2022decentralized,
  title={Decentralized Federated Learning: Fundamentals, State-of-the-art, Frameworks, Trends, and Challenges},
  author={Beltr{\'a}n, Enrique Tom{\'a}s Mart{\'\i}nez and P{\'e}rez, Mario Quiles and S{\'a}nchez, Pedro Miguel S{\'a}nchez and Bernal, Sergio L{\'o}pez and Bovet, G{\'e}r{\^o}me and P{\'e}rez, Manuel Gil and P{\'e}rez, Gregorio Mart{\'\i}nez and Celdr{\'a}n, Alberto Huertas},
  journal=IEEE_O_CSTO,
  volume={25},
  number={4},
  pages={2983--3013},
  year={2023},
  publisher={IEEE}
}

@INPROCEEDINGS{Akiba2019-jz,
  title     = {Optuna: A Next-generation Hyperparameter Optimization Framework},
  booktitle = {Proc.\ the 25th {ACM} {SIGKDD} International Conference on Knowledge Discovery \& Data Mining},
  author    = {Akiba, Takuya and Sano, Shotaro and Yanase, Toshihiko and Ohta, Takeru and Koyama, Masanori},
  pages     = {2623--2631},
  month     =  jul,
  year      =  2019,
  series = {KDD '19},
  address  = {Anchorage, AK, USA}
}

@inproceedings{jiang2017collaborative,
  title={Collaborative deep learning in fixed topology networks},
  author={Jiang, Zhanhong and Balu, Aditya and Hegde, Chinmay and Sarkar, Soumik},
  booktitle={Proc.\ 31st Conf. Neural Information Processing Systems (NeurIPS)},
  address={Long Beach, CA, USA},
  pages={1--11},
  month=dec,
  year={2017}
}

@article{liu2019linearized,
  title={Linearized {ADMM} for nonconvex nonsmooth optimization with convergence analysis},
  author={Liu, Qinghua and Shen, Xinyue and Gu, Yuantao},
  journal={IEEE Access},
  volume={7},
  pages={76131--76144},
  year={2019},
  publisher={IEEE}
}

@article{zhou2023federated,
  title={Federated learning via inexact {ADMM}},
  author={Zhou, Shenglong and Li, Geoffrey Ye},
  journal=IEEE_J_PAMI,
  volume={45},
  number={8},
  pages={9699--9708},
  year={2023},
  publisher={IEEE}
}

@article{bai2022inexact,
  title={An inexact accelerated stochastic {ADMM} for separable convex optimization},
  author={Bai, Jianchao and Hager, William W and Zhang, Hongchao},
  journal={Computational Optimization and Applications},
  volume={81},
  number={2},
  pages={479--518},
  year={2022},
  publisher={Springer}
}

@article{bai2025inexact,
  title={An inexact {ADMM} for separable nonconvex and nonsmooth optimization},
  author={Bai, Jianchao and Zhang, Miao and Zhang, Hongchao},
  journal={Computational Optimization and Applications},
  pages={1--35},
  year={2025},
  publisher={Springer}
}

@inproceedings{zeng2021federated,
  title={Federated learning for collaborative controller design of connected and autonomous vehicles},
  author={Zeng, Tengchan and Semiari, Omid and Chen, Mingzhe and Saad, Walid and Bennis, Mehdi},
  booktitle={60th IEEE Conference on Decision and Control (CDC)},
  pages={5033--5038},
  month = dec,
  year={2021},
  address={Austin, TX, USA},
  organization={IEEE}
}

@article{gao2023fedadt,
  title={{FedADT}: An adaptive method based on derivative term for federated learning},
  author={Gao, Huimin and Wu, Qingtao and Zhao, Xuhui and Zhu, Junlong and Zhang, Mingchuan},
  journal={Sensors},
  volume={23},
  number={13},
  pages={6034},
  year={2023},
  publisher={MDPI}
}

@inproceedings{mashaal2024extending,
  title={Extending Control Theory into Federated Learning Data Heterogeneity Problem},
  author={Mashaal, Omar and Baadel, Said and AbouZied, Hatem},
  booktitle={2024 IEEE International Conference on Artificial Intelligence and Mechatronics Systems (AIMS)},
  pages={1--4},
  month = feb,
  address={Bandung, Indonesia},
  year={2024},
  organization={IEEE}
}

@ARTICLE{li2025fed,
  title     = {{Fed-PID}: An adaptive learning rate scheduler for federated learning with {PID} controllers},
  author    = {Li, Shaofan and Dai, Mingjun and Kianoush, Sanaz and Minora, Alberto and Savazzi, Stefano},
  journal   = IEEE_J_CCN,
  publisher = {IEEE},
  pages     = {1--13},
  year      = {2025},
}

@article{wang2023communication,
  title={Communication-efficient {ADMM}-based distributed algorithms for sparse training},
  author={Wang, Guozheng and Lei, Yongmei and Qiu, Yongwen and Lou, Lingfei and Li, Yixin},
  journal={Neurocomputing},
  volume={550},
  pages={126456},
  year={2023},
  doi={10.1016/j.neucom.2023.126456},
  publisher={Elsevier}
}

@inproceedings{li2020federated,
 author = {Li, Tian and Sahu, Anit Kumar and Zaheer, Manzil and Sanjabi, Maziar and Talwalkar, Ameet and Smith, Virginia},
 booktitle = {Proc.\ Machine Learning and Systems},
 editor = {I. Dhillon and D. Papailiopoulos and V. Sze},
 pages = {429--450},
 title = {Federated Optimization in Heterogeneous Networks},
 volume = {2},
 year = {2020}
}

@inproceedings{mora2024knowledge,
  title={Knowledge distillation in federated learning: a practical guide},
  author={Mora, Alessio and Tenison, Irene and Bellavista, Paolo and Rish, Irina},
  booktitle={Proc.\ 33rd International Joint Conference on Artificial Intelligence},
  pages={8188--8196},
  doi = {10.24963/ijcai.2024/905},
  address = {Jeju, Korea},
  month = aug,
  year={2024},
}

@INPROCEEDINGS{pappas2021ipls,
  title     = {{IPLS}: A Framework for Decentralized Federated Learning},
  author    = {Pappas, Christodoulos and Chatzopoulos, Dimitris and Lalis,
               Spyros and Vavalis, Manolis},
  booktitle = {Proc.\ 2021 IFIP Networking Conference (IFIP Networking)},
  pages     = {1--6},
  month     =  jun,
  year      =  2021,
}

@ARTICLE{mota2013dadmm,
  author={Mota, Jo{\~a}o F. C. and Xavier, Jo{\~a}o M. F. and Aguiar, Pedro M. Q. and P{\"u}schel, Markus},
  journal=IEEE_J_SP, 
  title={{D-ADMM}: A Communication-Efficient Distributed Algorithm for Separable Optimization}, 
  year={2013},
  volume={61},
  number={10},
  pages={2718--2723},
  doi={10.1109/TSP.2013.2254478}
}

@article{hinton2015distilling,
  title={Distilling the knowledge in a neural network},
  author={Hinton, Geoffrey and Vinyals, Oriol and Dean, Jeff},
  journal={arXiv:1503.02531 [stat.ML]},
  year={2015}
}

@inproceedings{karimireddy2020scaffold,
  title={{SCAFFOLD}: Stochastic controlled averaging for federated learning},
  author={Karimireddy, Sai Praneeth and Kale, Satyen and Mohri, Mehryar and Reddi, Sashank and Stich, Sebastian and Suresh, Ananda Theertha},
  booktitle={Proc.\ 37th International Conference on Machine Learning},
  pages={5132--5143},
  year={2020},
  address={Virtual Conference},
  month={July},
  organization={PMLR}
}

@inproceedings{machler2022fedpidavg,
  title={{FedPIDAvg}: A {PID} controller inspired aggregation method for federated learning},
  author={M{\"a}chler, Leon and Ezhov, Ivan and Shit, Suprosanna and Paetzold, Johannes C},
  booktitle={Proc.\ Brainlesion:  Glioma, Multiple Sclerosis, Stroke  and Traumatic Brain Injuries: 8th International Workshop, BrainLes 2022, Held in Conjunction with MICCAI 2022},
  month=sep,
  address={Singapore},
  pages={209--217},
  year={2022},
  organization={Springer}
}

@article{onoszko2021decentralized,
  title={Decentralized federated learning of deep neural networks on non-{IID} data},
  author={Onoszko, Noa and Karlsson, Gustav and Mogren, Olof and Zec, Edvin Listo},
  journal={arXiv preprint arXiv:2107.08517  [cs.LG]},
  year={2021}
}

@inproceedings{chen2023enhancing,
  author={Chen, Min and Xu, Yang and Xu, Hongli and Huang, Liusheng},
  booktitle={Proc.\ 2023 IEEE 39th International Conference on Data Engineering (ICDE)}, 
  title={Enhancing Decentralized Federated Learning for Non-{IID} Data on Heterogeneous Devices}, 
  year={2023},
  month=apr,
  pages={2289--2302},
  address={Anaheim, CA, USA},
  organization={IEEE},
  doi={10.1109/ICDE55515.2023.00177}
  }

@article{boyd2011distributed,
  title={Distributed optimization and statistical learning via the alternating direction method of multipliers},
  author={Boyd, Stephen and Parikh, Neal and Chu, Eric and Peleato, Borja and Eckstein, Jonathan and others},
  journal={Foundations and Trends{\textregistered} in Machine learning},
  volume={3},
  number={1},
  pages={1--122},
  year={2011},
  publisher={Now Publishers, Inc.},
  doi={10.1561/2200000016},
}

@InProceedings{zhu2021data,
  title = {Data-Free Knowledge Distillation for Heterogeneous Federated Learning},
  author = {Zhu, Zhuangdi and Hong, Junyuan and Zhou, Jiayu},
  booktitle = {Proc.\ 38th International Conference on Machine Learning},
  pages = {12878--12889},
  year = {2021},
  volume = {139},
  month = jul,
}

@article{gad2024communication,
  title={Communication-efficient and privacy-preserving federated learning via joint knowledge distillation and differential privacy in bandwidth-constrained networks},
  author={Gad, Gad and Gad, Eyad and Fadlullah, Zubair Md and Fouda, Mostafa M and Kato, Nei},
  journal=IEEE_J_VT,
  volume={73},
  number={11},
  pages={17586--17601},
  year={2024},
  publisher={IEEE},
  doi={10.1109/TVT.2024.3423718}
}

@article{sun2024fkd,
  title={{FKD-Med}: privacy-aware, communication-optimized medical image segmentation via federated learning and model lightweighting through knowledge distillation},
  author={Sun, Guanqun and Shu, Han and Shao, Feihe and Racharak, Teeradaj and Kong, Weikun and Pan, Yizhi and Dong, Jingjing and Wang, Shuang and Nguyen, Le-Minh and Xin, Junyi},
  journal={IEEE Access},
  volume={12},
  pages={33687--33704},
  year={2024},
  publisher={IEEE},
  doi={10.1109/ACCESS.2024.3372394}
}

@inproceedings{zhang2024fedgmkd,
 title={{FedGMKD}: An efficient prototype federated learning framework through knowledge distillation and discrepancy-aware aggregation},
 author={Zhang, Jianqiao and Shan, Caifeng and Han, Jungong},
 booktitle={Advances in Neural Information Processing Systems},
 doi={10.52202/079017-3757},
 pages={118326--118356},
 volume={37},
 year={2024}
}

@article{li2025pfedkd,
  title={{PFedKD}: Personalized federated learning via knowledge distillation using unlabeled pseudo data for {Internet} of {Things}},
  author={Li, Hanxi and Chen, Guorong and Wang, Bin and Chen, Zheng and Zhu, Yongsheng and Hu, Fuqiang and Dai, Jiao and Wang, Wei},
  journal={IEEE Internet of Things Journal},
  journal=IEEE_J_IOT,
  volume={12},
  number={11},
  pages={16314--16324},
  year={2025},
  publisher={IEEE},
  doi={10.1109/JIOT.2025.3533003}
}

@article{sun2022decentralized,
  title={Decentralized federated averaging},
  author={Sun, Tao and Li, Dongsheng and Wang, Bao},
  journal=IEEE_J_PAMI,
  volume={45},
  number={4},
  pages={4289--4301},
  year={2023},
  publisher={IEEE},
  doi={10.1109/TPAMI.2022.3196503}
}

@article{li2025dfedadmm,
  title={{DFedADMM}: Dual constraint controlled model inconsistency for decentralize federated learning},
  author={Li, Qinglun and Shen, Li and Li, Guanghao and Yin, Quanjun and Tao, Dacheng},
  journal=IEEE_J_PAMI,
  year={2025},
  publisher={IEEE},
  volume={47},
  number={6},
  pages={4803--4815},
  doi={10.1109/TPAMI.2025.3546659}
}

@inproceedings{zhang2022fine,
  title={Fine-tuning global model via data-free knowledge distillation for non-{IID} federated learning},
  author={Zhang, Lin and Shen, Li and Ding, Liang and Tao, Dacheng and Duan, Ling-Yu},
  booktitle={Proc.\ 2022 IEEE/CVF Conf. Computer Vision and Pattern Recognition (CVPR)},
  pages={10164--10173},
  year={2022},
  address={New Orleans, LA, USA},
  doi={10.1109/CVPR52688.2022.00993}
}

@article{adnan2025framework,
  title={A Framework for privacy-preserving in {IoV} using Federated Learning with Differential Privacy},
  author={Adnan, Muhammad and Syed, Madiha Haider and Anjum, Adeel and Rehman, Semeen},
  journal={IEEE Access},
  volume={13},
  pages={13507--13521},
  year={2025},
  doi={10.1109/ACCESS.2025.3526934},
  publisher={IEEE}
}

@ARTICLE{wei2020federated,
  author={Wei, Kang and Li, Jun and Ding, Ming and Ma, Chuan and Yang, Howard H. and Farokhi, Farhad and Jin, Shi and Quek, Tony Q. S. and Vincent Poor, H.},
  journal=IEEE_J_IFS,
  title={Federated Learning With Differential Privacy: Algorithms and Performance Analysis}, 
  year={2020},
  volume={15},
  pages={3454--3469},
  doi={10.1109/TIFS.2020.2988575},
  publisher={IEEE}
}

@ARTICLE{he2024clustered,
  author={He, Zaobo and Wang, Lintao and Cai, Zhipeng},
  journal=IEEE_J_IOT,
  title={Clustered Federated Learning With Adaptive Local Differential Privacy on Heterogeneous {IoT} Data}, 
  year={2024},
  volume={11},
  number={1},
  pages={137--146},
  doi={10.1109/JIOT.2023.3299947},
  publisher={IEEE}
}

@article{yang2025dfun,
  title={{DFUN}-{KDF}: Efficient and robust decentralized federated framework for {UAV} networks via knowledge distillation and filtering},
  author={Yang, Wenyuan and Liu, Yuhang and Leng, Xinlin and Gu, Hanlin and Jiang, Gege and Yu, Xiaochuan and Cao, Xiaochun},
  journal={Communications in Transportation Research},
  volume={5},
  pages={100173},
  year={2025},
  doi={10.1016/j.commtr.2025.100173},
  publisher={Elsevier}
}

@inproceedings{soltani2024dflstar,
  author = {Soltani, Behnaz and Haghighi, Venus and Zhou, Yipeng and Sheng, Quan Z. and Yao, Lina},
  title = {{DFLStar}: A Decentralized Federated Learning Framework with Self-Knowledge Distillation and Participant Selection},
  month = oct,
  year = {2024},
  doi = {10.1145/3627673.3679853},
  booktitle = {Proc. of the 33rd ACM International Conference on Information and Knowledge Management},
  pages = {2108--2117},
  address = {Boise, ID, USA},
  series = {CIKM '24},
}

@STRING{IEEE_J_IOT        = "{IEEE} Internet Things J."}

@STRING{IEEE_J_SIPN       = "{IEEE} Trans. Signal Inf. Process. Netw."}

@STRING{IEEE_J_CCN        = "{IEEE} Trans. Cogn. Commun. Netw."}

%

\begin{IEEEbiography}[{\includegraphics[width=1in,height=1.25in,clip,keepaspectratio]{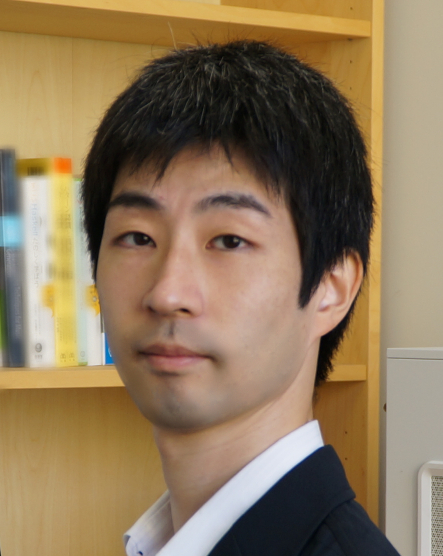}}]{Akihito~Taya}
received the B.E.\ degree in electrical and electronic engineering from Kyoto University, Kyoto, Japan in 2011,
and the master and Ph.D.\ degree in Informatics from Kyoto University in 2013 and 2019, respectively.
From 2013 to 2017, he joined Hitachi, Ltd., where he participated in the development of computer clusters.
From 2019 to 2022, he was an assistant professor at Aoyama Gakuin University.
He has been an assistant professor at Institute of Industrial Science, the University of Tokyo since 2022.
He received the IEEE VTS Japan Young Researcher's Encouragement Award in 2012, the IEICE Young Researcher's Award in 2018, and the IEEE CCNC Best Demo Award - Runner Up in 2026.
His current research interests include IoT, distributed machine learning, and Wi-Fi sensing.
He is a member of the IEEE and ACM.
\end{IEEEbiography}

\begin{IEEEbiography}[{\includegraphics[width=1in,height=1.25in,clip,keepaspectratio]{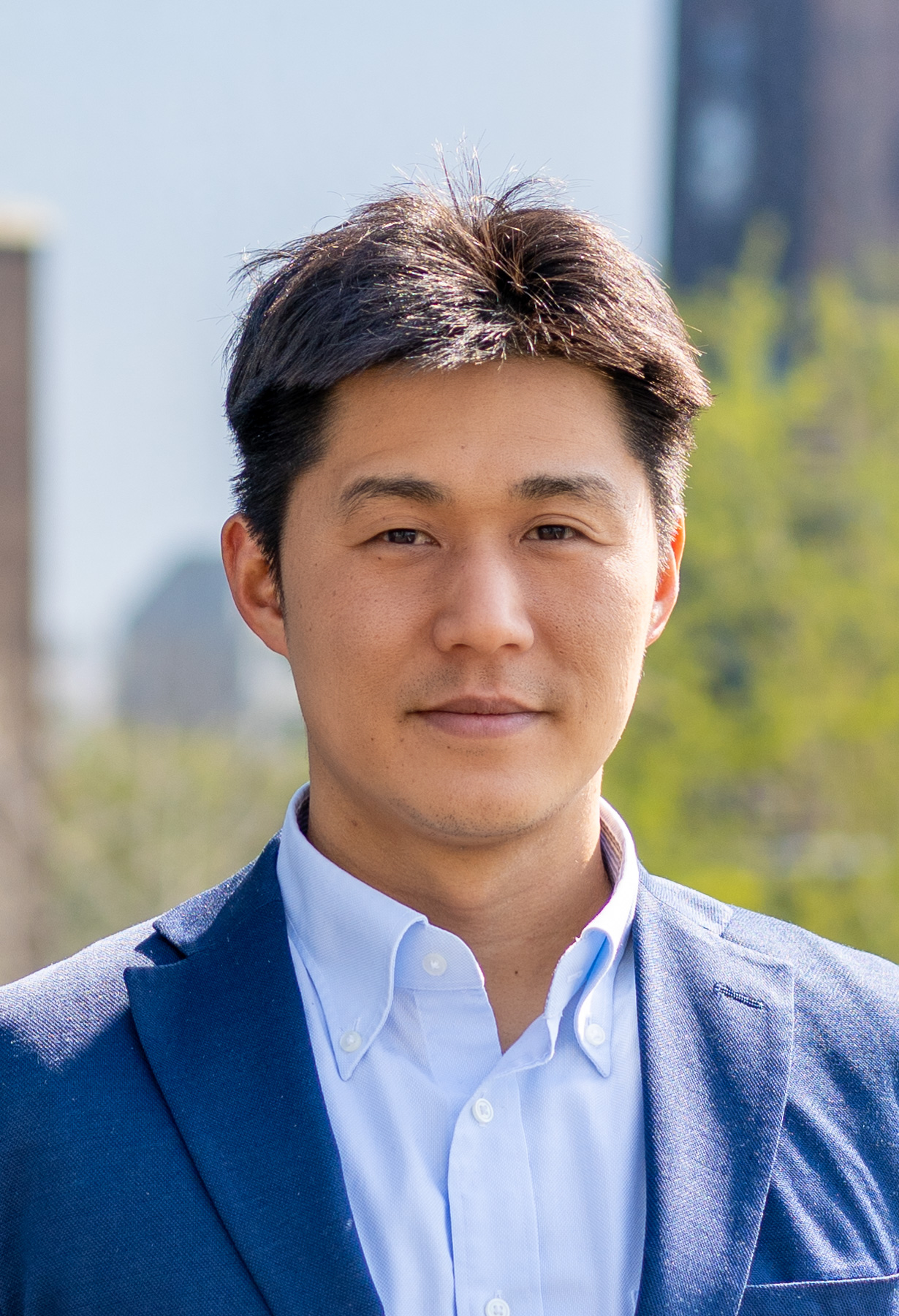}}]{Yuuki~Nishiyama}
is an Associate Professor at Center for Spatial Information Science (CSIS), the University of Tokyo. He obtained an M.S.\ (2014) in Media and Governance from Keio University and a Ph.D.\ in Media and Governance (2017) from the same institution. He had worked at Keio University in Japan and the University of Oulu in Finland as a post-doctoral researcher, respectively. He began his career at the Institute of Industrial Science, University of Tokyo, as a Research Associate in 2019 and has worked as an Assistant Professor (Lecturer) at CSIS since 2022. Since 2025, he has held his current position. His current research interests include ubiquitous computing systems, mobile and wearable sensing platforms, and human ability augmentation. He is a member of ACM, IEEE, and the Information Processing Society of Japan (IPSJ).
\end{IEEEbiography}

\begin{IEEEbiography}[{\includegraphics[width=1in,height=1.25in,clip,keepaspectratio]{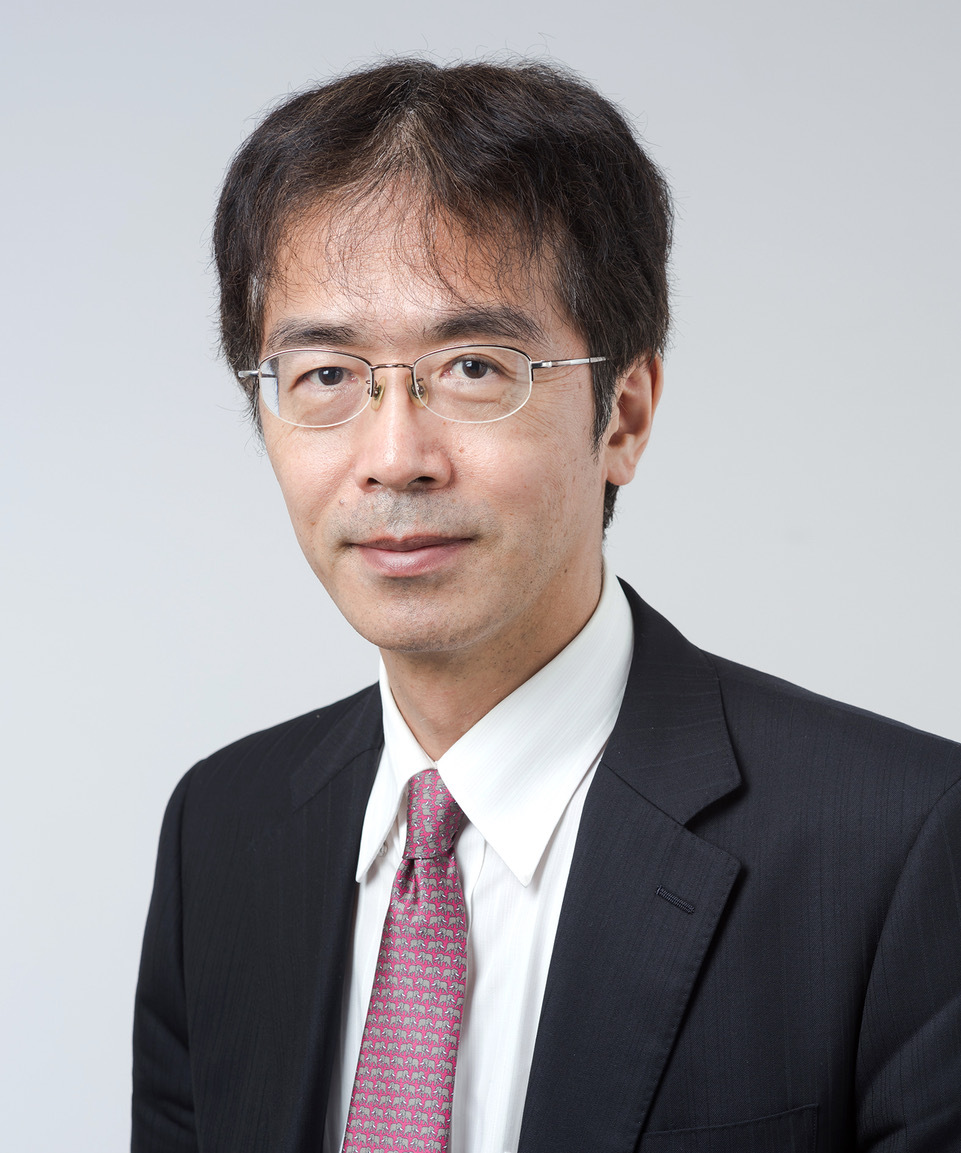}}]{Kaoru~Sezaki}
is a professor of Center for spatial Science at the University of Tokyo and co-appointed as professor of Institute of Industrial Science, the University of Tokyo. He served as director of Center for Spatial Information Science from 1998 to 2024. He is a steering member of e-Health Technical Committee COMSOC. He has been general chair and TPC Chair of many IEEE international conferences. He also served as Treasurer of IEEE Tokyo Section as well as that of Japan Council from 2003 to 2004. He received B.Eng., M.Eng., and Ph.D.\ degrees from the University of Tokyo, Tokyo, Japan, in 1984, 1986, and 1989, respectively, all in Electrical Engineering. Since 1989, he has been with the University of Tokyo. He was a Visiting Researcher at University of California at San Diego in 1996. His research interests include e-Health, sensor networks, IoT, and urban computing. 
\end{IEEEbiography}





\end{document}